\documentclass{article}

\usepackage{arxiv}

\usepackage[utf8]{inputenc} 
\usepackage[T1]{fontenc}    
\usepackage{hyperref}       
\usepackage{url}            
\usepackage{booktabs}       
\usepackage{amsfonts}       
\usepackage{nicefrac}       
\usepackage{lipsum}
\usepackage{graphicx}
\graphicspath{ {./images/} }

\usepackage{times}  
\usepackage{helvet}  
\usepackage{natbib}  
\usepackage{caption} 
\usepackage{algorithm}
\usepackage{amsmath}
\usepackage{amsthm}
\usepackage{booktabs}
\usepackage[switch]{lineno}
\usepackage{amssymb}
\usepackage{tikz}
\usepackage{mathtools}
\usepackage{sansmath}
\usepackage{multirow}
\usepackage{subcaption}
\usepackage{algpseudocode}
\usepackage{bm}

\usepackage{xcolor} 
\usepackage{float}

\title{Learning with Preserving for Continual Multitask Learning}

\author{
  Hanchen David Wang\thanks{Equal contribution.} \\
  Vanderbilt University\\
  Nashville, TN 37235 USA \\
  \texttt{hanchen.wang.1@vanderbilt.edu} \\
  \And
  Siwoo Bae\footnotemark[1] \\
  Vanderbilt University\\
  Nashville, TN 37235 USA \\
  \texttt{siwoo.bae@vanderbilt.edu} \\
  \And
  Zirong Chen \\
  Vanderbilt University\\
  Nashville, TN 37235 USA \\
  \texttt{zirong.chen@vanderbilt.edu} \\
  \And
  Meiyi Ma \\
  Vanderbilt University\\
  Nashville, TN 37235 USA \\
  \texttt{meiyi.ma@vanderbilt.edu} \\
}
\date{}
\begin{document}

\maketitle

\maketitle

\begin{abstract}
Artificial intelligence systems in critical fields like autonomous driving and medical imaging analysis often continually learn new tasks using a shared stream of input data. For instance, after learning to detect traffic signs, a model may later need to learn to classify traffic lights or different types of vehicles using the same camera feed. This scenario introduces a challenging setting we term Continual Multitask Learning (CMTL), where a model sequentially learns new tasks on an underlying data distribution without forgetting previously learned abilities. Existing continual learning methods often fail in this setting because they learn fragmented, task-specific features that interfere with one another. To address this, we introduce Learning with Preserving (LwP), a novel framework that shifts the focus from preserving task outputs to maintaining the geometric structure of the shared representation space. The core of LwP is a Dynamically Weighted Distance Preservation (DWDP) loss that prevents representation drift by regularizing the pairwise distances between latent data representations. This mechanism of preserving the underlying geometric structure allows the model to retain implicit knowledge and support diverse tasks without requiring a replay buffer, making it suitable for privacy-conscious applications. Extensive evaluations on time-series and image benchmarks show that LwP not only mitigates catastrophic forgetting but also consistently outperforms state-of-the-art baselines in CMTL tasks. Notably, our method shows superior robustness to distribution shifts and is the only approach to surpass the strong single-task learning baseline, underscoring its effectiveness for real-world dynamic environments.\footnote{Work accepted at AAAI-2026. The code is available at \href{https://github.com/AICPS-Lab/lwp}{https://github.com/AICPS-Lab/lwp}}.
\end{abstract}


\section{Introduction}

In critical applications such as intelligent driving, a system must continually adapt by learning new tasks from a consistent stream of sensory data. This paradigm is driven by practicality: when a new task is introduced, the cost of retrospectively annotating the entire existing dataset with the new labels is often unsustainable \cite{golatkar2020eternal}. It is far more efficient to instead leverage the existing data stream by acquiring labels only for the new task as needed. For instance, after a model learns to detect traffic signs, it can later be taught to classify other attributes like scene types using the same camera feed \cite{autonomous_CL_review, kang2024continual}. Similarly, in medical imaging, a model trained for tumor classification can be updated to recognize secondary characteristics such as tissue density or shape, all while using the same underlying patient scans \cite{an2025isl, freeman_iterative}. The central challenge, therefore, is to learn new tasks by acquiring new labels for a shared and potentially evolving input distribution. 


We term this challenging real-world setting \textbf{Continual Multitask Learning (CMTL)}. CMTL combines challenges from both Multitask Learning (MTL) and Continual Learning (CL). In a typical CMTL scenario, a model is presented with a sequence of tasks, $T_1, T_2, ..., T_n$. Each task introduces a new label set applied to inputs from the same sensor/input space (though their underlying distribution may differ across tasks). This setting is distinct from standard MTL, where all tasks are known and trained on concurrently, and it presents unique challenges not fully addressed by conventional Task-Incremental Learning (Task-IL) methods \cite{van_de_ven_three_2022}. The key distinctions are summarized in Table \ref{tab:learning_scenarios}.

\begin{table}[h!]
\centering
\caption{Comparison of Scenarios. CMTL uniquely requires learning new tasks sequentially on a shared data distribution without full access to past task labels.}
\label{tab:learning_scenarios}
\begin{tabular}{lccc}
\toprule
\textbf{Characteristic} & \textbf{MTL} & \textbf{\shortstack{Task-IL \\ (Standard CL)}} &  \textbf{\shortstack{CMTL \\ (Our Setting)}} \\
\midrule
Tasks Arrive Sequentially & \text{\sffamily X} & \checkmark & \checkmark \\
All Task Data Available Concurrently & \checkmark & \text{\sffamily X} & \text{\sffamily X} \\
Tasks Share an Input Domain & \checkmark & (Often Not) & \checkmark \\
Goal: Learn Shared Representation & \checkmark & \text{\sffamily X} & \checkmark \\
Goal: Mitigate Forgetting & N/A & \checkmark & \checkmark \\
\bottomrule
\end{tabular}%

\end{table}

The primary challenge in CMTL is twofold: the model must 1) retain knowledge from previous tasks to prevent catastrophic forgetting (a core CL goal), and 2) develop robust, shared representations that benefit multiple distinct tasks (a core MTL goal), all without having simultaneous access to the complete labeled data for all tasks. This is especially difficult when the underlying data distribution shifts over time (a non-stationary setting), which can exacerbate task interference.

Although CMTL can be formally categorized as a case of Task-IL, its strong emphasis on building a unified representation from a shared input domain exposes a key weakness in conventional CL methods. These approaches are primarily designed to prevent catastrophic forgetting, often by isolating task-specific knowledge \cite{kirkpatrick_overcoming_2017, zenke_continual_2017, ma2020stlnet}. As our experiments confirm (Table \ref{tab:model_accuracy_combined}), this strategy frequently struggles in the CMTL setting, leading to performance below that of even single-task baselines. Handling heterogeneous tasks on shared inputs requires unified representations \cite{jiao2025unitoken,huang2023video}, yet conventional CL methods fail this requirement by design, relying on parameter freezing and replay buffers that isolate rather than integrate task-specific knowledge.

To address these challenges, we introduce \textbf{Learning with Preserving (LwP)}, a framework designed specifically for the CMTL setting. Instead of focusing only on task outputs, LwP directly preserves the integrity of the shared representation space throughout sequential training. Its core principles are: \textbf{(i)}, a novel regularization term (the DWDP loss) to explicitly maintain the geometric structure of the model's latent space and prevent representation drift, \textbf{(ii)}, a stabilized representation space to preserve the implicit knowledge encoded in the geometric relationships between data points, and \textbf{(iii)}, a framework operates without a replay buffer, making it efficient for privacy-constrained applications.

The main contributions of this paper are: \textbf{(1)} We formally define and analyze CMTL, and we demonstrate that conventional CL methods are often ill-suited for this context. \textbf{(2)} We propose Learning with Preserving (LwP), a novel, replay-free framework whose key innovation is a Dynamically Weighted Distance Preservation (DWDP) loss function that maintains the geometric integrity of the latent representation space, mitigating catastrophic forgetting while promoting knowledge sharing. \textbf{(3)} We conduct extensive evaluations on image and time-series benchmarks, showing that LwP consistently and significantly outperforms state-of-the-art baselines and, unlike other methods, surpasses the performance of independently trained single-task models, especially in scenarios with distribution shifts.

\newcommand{\inputx}{\bm{x}}

\section{Problem Formulation: Continual Multitask Learning}


CMTL is a sequential learning scenario involving T tasks $\{\mathcal{T}_t\}_{t=1}^{T}$. Each task $\mathcal{T}_t$ is associated with a label space $\mathcal{Y}_t$ and involves learning a mapping $f_{t}:\mathcal{X}\rightarrow\mathcal{Y}_{t}$. 
At each time step t, we receive a dataset $D_{t}=\{(x_{i},y_{i}^{t})\}_{i=1}^{n_{t}}$ where the input $x_i$ is drawn from a task-specific distribution $x_{i}\sim P_{X}^{(t)}$, and $y_{i}^{t}\in\mathcal{Y}_{t}$ is the corresponding label for task $\mathcal{T}_{t}$. Note that for time t, only label $y_{i}^{t}$ is available.

In simpler CMTL settings, the distribution is stationary ($P_{X}^{(t)} = P_{X}^{(j)}$ for all $t, j$), while in more challenging settings, it can be non-stationary ($P_{X}^{(t)} \neq P_{X}^{(j)}$ for $t \neq j$). Our goal is to find a predictor $\varphi(x;\theta_{s},\theta_{t}):\mathcal{X}\rightarrow\mathcal{Y}_{1}\times \dots \times \mathcal{Y}_{T}$ parameterized by a set of shared parameters \( \theta_s \) and task-specific parameters \( \theta_t \), such that

\begin{equation}
\mathcal{L}(\theta_s, \{\theta_t\}_{t=1}^{T}) \coloneqq \sum_{t=1}^{T} \mathbb{E}_{(\bm{x}, y^t) \leftarrow \mathcal{D}_t} \left[\ell\left(y^t, \varphi(\bm{x}, t; \theta_s, \theta_t)\right)\right],
\end{equation}

is minimized for some loss function \( \ell(\cdot, \cdot) \).

\section{Learning with Preserving}\label{section:lwp}
This section details our proposed framework, Learning with Preserving (LwP). We begin with a high-level overview of the architecture and training process in Section \ref{ssec:lwp_overview}. Then we explore the theoretical motivation for our core contribution, a preservation loss designed to maintain the geometric structure of the latent space, in Section \ref{pres_impl_knowledge}. Finally, in Section \ref{ssec:dynamic_weighting}, we introduce the dynamic weighting mechanism that makes this loss effective for discriminative tasks.

\subsection{Overview}\label{ssec:lwp_overview}
\begin{figure*}[!t]
    \centering
    \includegraphics[width=0.75\linewidth]{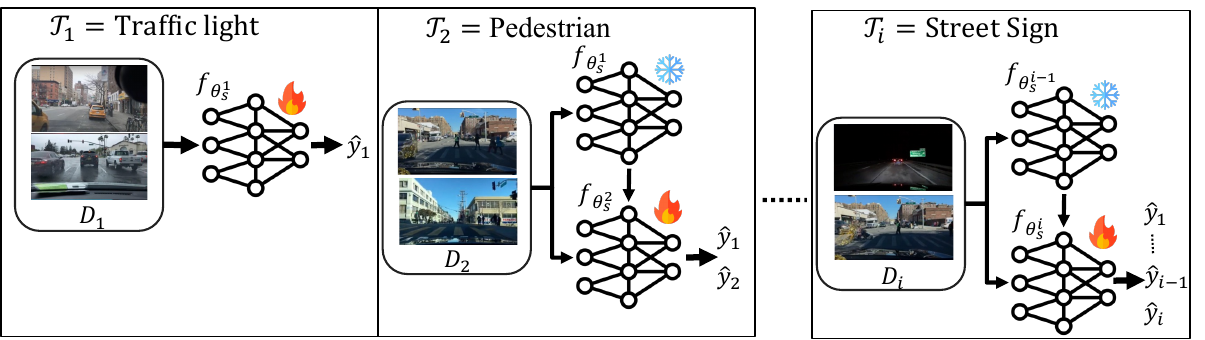}
    \caption{Overview of the LwP framework. For the first task, $\mathcal{T}_1$ (e.g., Traffic Light), the model is trained on data $D_1$. When learning a subsequent task like $\mathcal{T}_2$ (Pedestrian), the model from $\mathcal{T}_1$ is frozen as a teacher. This process generalizes for any current task $\mathcal{T}_i$: the model from the previous step, $f_{\theta_{s}^{[i-1]}}$, acts as a teacher for the student model, $f_{\theta_{s}^{[i]}}$, which learns on new data $D_i$ using a supervised loss ($\mathcal{L}_{\text{cur}}$), a distillation loss ($\mathcal{L}_{\text{old}}$), and our geometric preservation loss ($\mathcal{L}_{\text{DWDP}}$).}
    
    \label{fig:framework_overview_tumor}
\end{figure*}

We introduce \textbf{Learning with Preserving (LwP)}, a framework designed to manage CMTL scenarios by maintaining the structural integrity of the model's shared representation space across a sequence of tasks. As depicted in Figure \ref{fig:framework_overview_tumor}, LwP uses a shared feature extractor $f_{\theta_s}(\bm{x})$ to produce a representation $\bm{z}$. For each task $t$, a separate, task-specific \textit{head} $g_{\theta_t}(\bm{z})$ (e.g., a linear layer) generates the final prediction.

The training process at a given task $t$ proceeds as follows. First, we duplicate the model from task $t-1$ and add a new, randomly initialized head, $g_{\theta_t}$, for the new task. The parameters of the previous model (both the feature extractor $f_{\theta_s}^{[t-1]}$ and all previous heads $g_{\theta_o}^{[t-1]}$ for $o < t$) are then \textbf{frozen} to serve as a stable teacher. 

The current model is trained using a composite loss function, $\mathcal{L}_{\text{lwp}}$, which consists of three key components:
\begin{enumerate}
    \item A standard supervised loss ($\mathcal{L}_{\text{cur}}$) for the current task $t$, which trains the new head $g_{\theta_t}$ and fine-tunes the shared extractor $f_{\theta_s}$ on new data $(x_i, y_i^t)$.
    \item A distillation loss ($\mathcal{L}_{\text{old}}$) that preserves performance on previous tasks. The frozen teacher model generates ``pseudolabels'' $\tilde{y}_o$ for old tasks $o < t$, and the current model is trained to match them.
    \item Our novel preservation loss ($\mathcal{L}_{\text{DWDP}}$), which is the core of LwP. This loss prevents representation drift by ensuring the geometric structure of the current latent space ($z^{[t]}$) remains consistent with the structure of the frozen latent space ($z^{[t-1]}$).
\end{enumerate}
Thus, the total objective is a weighted sum of these three components: 
\begin{align}
    \mathcal{L}_{\text{lwp}} = &\; \lambda_{\text{c}} \mathcal{L}_{\text{cur}}+ \lambda_{\text{o}} \mathcal{L}_{\text{old}}+ \lambda_{\text{d}} \mathcal{L}_{\text{DWDP}}
\end{align}

\subsection{Preserving Implicit Knowledge}
\label{pres_impl_knowledge}

In CMTL, a model must preserve not only explicit knowledge from past tasks but also the \textit{implicitly learned knowledge} encoded in its shared representation $\bm{z}$. We define this implicit knowledge as the geometric structure of the latent space. To prevent this structure from degrading as new tasks are learned, we introduce a loss function designed to explicitly preserve it.

A direct way to maintain this geometric structure is to ensure that the pairwise distances between data points in the current model's latent space, $Z' = f_{\theta_s^{[t]}}(X)$, remain close to those from the frozen previous model's space, $Z = f_{\theta_s^{[t-1]}}(X)$. 
This leads to a family of preservation losses:
\begin{equation}
    \mathcal{L}_{\text{pres}}(Z, Z') = \frac{1}{N^2} \sum_{i=1}^{N} \sum_{j=1}^{N} \left( d(\bm{z}_i, \bm{z}_j) - d(\bm{z}'_i, \bm{z}'_j) \right)^2,
    \label{eq:general_pres_loss}
\end{equation}
where $d(\cdot, \cdot)$ is a distance or similarity function, which in our primary implementation is the squared Euclidean distance.

This formulation is closely related to kernel methods. Preserving pairwise distances implicitly preserves the structure defined by certain kernels. As we show in Appendix \ref{appendix:sqdist}, the change in the Gaussian kernel value is bounded by the change in the squared Euclidean distance. Therefore, minimizing the difference in distances effectively minimizes the difference between the Gram matrices $K(Z)$ and $K(Z')$, where $K_{ij} = k(\bm{z}_i, \bm{z}_j)$.

Preserving the Gram matrix ($K(Z') \approx K(Z)$) ensures that the new representation $Z'$ is functionally equivalent to the old representation $Z$ within the Reproducing Kernel Hilbert Space (RKHS) induced by the kernel \cite{yamada2013inequalities, scholkopf2001generalized}. This means that for any function $f$ in this universal function space, its evaluation on a new data point $f(\bm{z}'_i)$ can be mapped to an equivalent function $f'$ evaluated on the old data point, $f'(\bm{z}_i)$. Formally, this is because preserving the kernel matrix implies an isometry $T$ in the RKHS such that $\phi(\bm{z}'_i) = T(\phi(\bm{z}_i))$, where $\phi$ is the feature map. Consequently, any learning problem defined on $Z$ has an equivalent solution on $Z'$, as shown in Equations \ref{eq:learning_problem_Z}-\ref{eq:optimal_solutions_equivalent}.

\usetikzlibrary{arrows.meta, intersections}
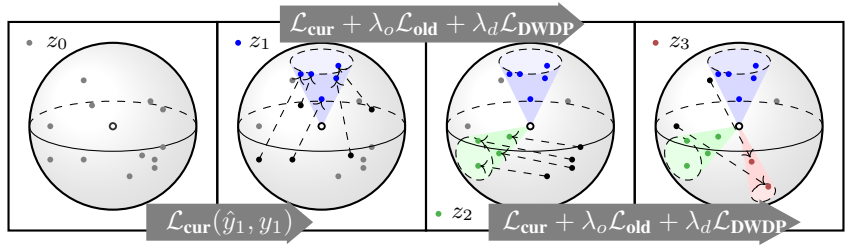
\begin{figure*}
    \centering
    \begin{tikzpicture}[font = \sansmath, scale=0.555] 

  \begin{scope}[xshift=-8cm]
    \coordinate (O) at (0,0);
    \fill[black] (O) circle (2.8pt); 
    \fill[white] (O) circle (1.5pt); 
    \shade[ball color = white, opacity = 0.2] (0,0) circle [radius = 2cm];
    \draw[thick] (0,0) circle [radius=2cm];
    \draw[thin] (-2,0) arc (180:360:2 and 0.6);
    \draw[thin, dashed] (2,0) arc (0:180:2 and 0.6);
    \fill[gray] (-0.5, 0.5) circle (2pt);
    \fill[gray] (0.7, -0.7) circle (2pt);
    
    \fill[gray] (1, -0.8) circle (2pt);
    \fill[gray] (1.2, 0.4) circle (2pt);
    \fill[gray] (-1.5, 0) circle (2pt);
    \fill[gray] (0.9, 0.6) circle (2pt);
    \fill[gray] (1.2, -0.5) circle (2pt);
    \fill[gray] (0.4, -1.2) circle (2pt);
    \fill[gray] (-0.7, 1.1) circle (2pt);
    \fill[gray] (1.0, -1.0) circle (2pt);
    \fill[gray] (-1.5, -0.8) circle (2pt);
    \fill[gray] (-0.7, -0.8) circle (2pt);
    
    \draw[thick] (-2.5,-2.5) rectangle (2.5,2.5);
    \fill[gray] (-2, 2) circle (2pt);
    \node[right=1mm, font=\footnotesize] at (-2, 2) {$z_0$}; 

  \end{scope}

  \begin{scope}[xshift=-3cm]
    \coordinate (O) at (0,0);
    \shade[ball color = white, opacity = 0.2] (0,0) circle [radius = 2cm];

    \def\rx{0.75}
    \def\ry{0.3}
    \def\z{1.55}

    \path [name path = ellipse]    (0,\z) ellipse ({\rx} and {\ry});
    \path [name path = horizontal] (-\rx,\z-\ry*\ry/\z)
                                  -- (\rx,\z-\ry*\ry/\z);
    \path [name intersections = {of = ellipse and horizontal}];

    \draw[fill = blue!15, blue!15] (intersection-1) -- (0,0)
      -- (intersection-2) -- cycle;
    \draw[fill = blue!5, densely dashed] (0,\z) ellipse ({\rx} and {\ry});

    \fill[gray] (1, -0.8) circle (2pt);
    \fill[gray] (-1.5, 0) circle (2pt);
    \fill[gray] (0.9, 0.6) circle (2pt);
    \fill[gray] (1.2, -0.5) circle (2pt);
    \fill[gray] (0.4, -1.2) circle (2pt);
    \fill[gray] (-0.7, 1.1) circle (2pt);
    \fill[gray] (1.0, -1.0) circle (2pt);
    \fill[black] (-0.5, 0.5) circle (2pt);
    \fill[blue] (-0.25, \z-0.3) circle (2pt);
    
    \draw[->, dashed] (-0.5, 0.5) -- (-0.25, \z-0.35);

    \fill[black] (0.7, -0.7) circle (2pt);
    \fill[blue] (0.35, \z-0.4) circle (2pt);
    \draw[->, dashed] (0.7, -0.7) -- (0.35, \z-0.45);

    \fill[black] (-0.7, -0.8) circle (2pt);
    \fill[blue] (0, \z-0.9) circle (2pt);
    \draw[->, dashed] (-0.7, -0.8) -- (0, \z-0.95);

    \fill[black] (-1.5, -0.8) circle (2pt);
    \fill[blue] (-0.5, \z-0.3) circle (2pt);
    \draw[->, dashed] (-1.5, -0.8) -- (-0.5, \z-0.35);

    \fill[black] (1.2, 0.4) circle (2pt);
    \fill[blue] (0.4, \z-0.1) circle (2pt);
    
    \draw[->, dashed] (1.2, 0.4) -- (0.4, \z-0.15);

    \draw[thick] (0,0) circle [radius=2cm];
    \draw[thin] (-2,0) arc (180:360:2 and 0.6);
    \draw[thin, dashed] (2,0) arc (0:180:2 and 0.6);

    \draw[thick] (-2.5,-2.5) rectangle (2.5,2.5);
    \fill[black] (O) circle (2.8pt); 
    \fill[white] (O) circle (1.5pt); 
    
    \fill[blue] (-2, 2) circle (2pt);
    \node[right=.2mm, font=\footnotesize] at (-2, 2) {$z_1$}; 

    \end{scope}

  \begin{scope}[xshift=2cm]
    \coordinate (O) at (0,0);
    \shade[ball color = white, opacity = 0.2] (0,0) circle [radius = 2cm];

    \def\rx{0.75}
    \def\ry{0.3}
    \def\z{1.55}

    \path [name path = ellipse]    (0,\z) ellipse ({\rx} and {\ry});
    \path [name path = horizontal] (-\rx,\z-\ry*\ry/\z)
                                  -- (\rx,\z-\ry*\ry/\z);
    \path [name intersections = {of = ellipse and horizontal}];

    \draw[fill = blue!15, blue!15] (intersection-1) -- (0,0)
      -- (intersection-2) -- cycle;
    \draw[fill = blue!5, densely dashed] (0,\z) ellipse ({\rx} and {\ry});

    \fill[gray] (-1.5, 0) circle (2pt);
    \fill[gray] (0.9, 0.6) circle (2pt);
    \fill[gray] (-0.7, 1.1) circle (2pt);
    
    \fill[blue] (-0.25, \z-0.3) circle (2pt);
    \fill[blue] (0.4, \z-0.1) circle (2pt);
    \fill[blue] (-0.5, \z-0.3) circle (2pt);
    \fill[blue] (0, \z-0.9) circle (2pt);
    \fill[blue] (0.35, \z-0.4) circle (2pt);

    \begin{scope}[rotate around={120:(O)}]
      \def\rx{0.55}
      \def\ry{0.4}
      \def\z{1.55}

      \path [name path = ellipse]    (0,\z) ellipse ({\rx} and {\ry});
      \path [name path = horizontal] (-\rx,\z-\ry*\ry/\z)
                                  -- (\rx,\z-\ry*\ry/\z);
      \path [name intersections = {of = ellipse and horizontal}];

      \draw[fill = green!15, green!15] (intersection-1) -- (0,0)
        -- (intersection-2) -- cycle;
      \draw[fill = green!10, densely dashed] (0,\z) ellipse ({\rx} and {\ry});

    \end{scope}
    
    \fill[black] (1, -0.8) circle (2pt);
    \fill[green!40!gray] (-1.25, \z-1.9) circle (2pt);
    \draw[->, dashed] (1, -0.8) -- (-1.2, \z-1.9);
    
    
    \fill[black] (1.0, -1.0) circle (2pt);
     \fill[green!40!gray] (-.75, \z-2.2) circle (2pt);
    \draw[->, dashed] (1.0, -1.0) -- (-.7, \z-2.2);

    \fill[black] (0.4, -1.2) circle (2pt);
    \fill[green!40!gray] (-1.25, \z-2.5) circle (2pt);
    \draw[->, dashed] (0.4, -1.2) -- (-1.2, \z-2.5);

    \fill[black] (1.2, -0.5) circle (2pt);
    \fill[green!40!gray] (-.5, \z-1.8) circle (2pt);
    \draw[->, dashed] (1.2, -0.5) -- (-.45, \z-1.8);

    \draw[thick] (0,0) circle [radius=2cm];
    \draw[thin] (-2,0) arc (180:360:2 and 0.6);
    \draw[thin, dashed] (2,0) arc (0:180:2 and 0.6);
    \fill[black] (O) circle (2.8pt); 
    \fill[white] (O) circle (1.5pt); 
    \draw[thick] (-2.5,-2.5) rectangle (2.5,2.5);
    \fill[green!40!gray] (-2.2, -2.1) circle (2pt);
    \node[right=.5mm, font=\footnotesize] at (-2.2, -2.1) {$z_2$}; 
  \end{scope}
    
  \begin{scope}[xshift=7cm]
    \coordinate (O) at (0,0);
    \shade[ball color = white, opacity = 0.2] (0,0) circle [radius = 2cm];

    \def\rx{0.75}
    \def\ry{0.3}
    \def\z{1.55}

    \path [name path = ellipse]    (0,\z) ellipse ({\rx} and {\ry});
    \path [name path = horizontal] (-\rx,\z-\ry*\ry/\z)
                                  -- (\rx,\z-\ry*\ry/\z);
    \path [name intersections = {of = ellipse and horizontal}];

    \draw[fill = blue!15, blue!15] (intersection-1) -- (0,0)
      -- (intersection-2) -- cycle;
    \draw[fill = blue!10, densely dashed] (0,\z) ellipse ({\rx} and {\ry});
    
    \fill[gray] (0.9, 0.6) circle (2pt);
    \fill[blue] (-0.25, \z-0.3) circle (2pt);
    \fill[blue] (0.4, \z-0.1) circle (2pt);
    \fill[blue] (-0.5, \z-0.3) circle (2pt);
    \fill[blue] (0, \z-0.9) circle (2pt);
    \fill[blue] (0.35, \z-0.4) circle (2pt);

    \begin{scope}[rotate around={120:(O)}]
      \def\rx{0.55}
      \def\ry{0.4}
      \def\z{1.55}

      \path [name path = ellipse]    (0,\z) ellipse ({\rx} and {\ry});
      \path [name path = horizontal] (-\rx,\z-\ry*\ry/\z)
                                  -- (\rx,\z-\ry*\ry/\z);
      \path [name intersections = {of = ellipse and horizontal}];

      \draw[fill = green!15, green!15] (intersection-1) -- (0,0)
        -- (intersection-2) -- cycle;
      \draw[fill = green!10, densely dashed] (0,\z) ellipse ({\rx} and {\ry});
    \end{scope}
    \fill[green!40!gray] (-.5, \z-1.8) circle (2pt);
    \fill[green!40!gray] (-1.25, \z-2.5) circle (2pt);
    \fill[green!40!gray] (-.75, \z-2.2) circle (2pt);
     \fill[green!40!gray] (-1.25, \z-1.9) circle (2pt);
    \begin{scope}[rotate around={200:(O)}]
      \def\rx{0.3}
      \def\ry{0.2}
      \def\z{1.7}

      \path [name path = ellipse]    (0,\z) ellipse ({\rx} and {\ry});
      \path [name path = horizontal] (-\rx,\z-\ry*\ry/\z)
                                  -- (\rx,\z-\ry*\ry/\z);
      \path [name intersections = {of = ellipse and horizontal}];

      \draw[fill = red!15, red!15] (intersection-1) -- (0,0)
        -- (intersection-2) -- cycle;
      \draw[fill = red!10, densely dashed] (0,\z) ellipse ({\rx} and {\ry});
    \end{scope}

    \fill[black] (-1.5, 0) circle (2pt);
    \fill[red!40!gray] (.7, \z-3) circle (2pt);
    \draw[->, dashed] (-1.5, 0) -- (.625, \z-2.95);
    
    
    \fill[black] (-0.7, 1.1) circle (2pt);
     \fill[red!40!gray] (.3, \z-2.4) circle (2pt);
    \draw[->, dashed] (-0.7, 1.1) -- (.27, \z-2.3);
    \draw[thick] (0,0) circle [radius=2cm];
    \draw[thin] (-2,0) arc (180:360:2 and 0.6);
    \draw[thin, dashed] (2,0) arc (0:180:2 and 0.6);
    \fill[black] (O) circle (2.8pt); 
    \fill[white] (O) circle (1.5pt); 
    \draw[thick] (-2.5,-2.5) rectangle (2.5,2.5);
    \fill[red!40!gray] (-2, 2) circle (2pt);
    \node[right=.5mm, font=\footnotesize] at (-2, 2) {$z_3$}; 
  \end{scope}
  
\begin{scope}[transparency group, opacity=1, gray]
\draw[-{Triangle[width=16pt,length=10pt]}, line width=14pt]
    (-7.2, -2.3) -- (-3.2, -2.3) node[midway, text=white, font=\bfseries] {$\mathcal{L}_{\text{cur}}(\hat{y}_{1}, y_{1})$};
\end{scope}

\begin{scope}[transparency group, opacity=1, gray]
\draw[-{Triangle[width=16pt,length=10pt]}, line width=14pt]
    (-4, 2.45) -- (3.3, 2.45) node[midway, text=white, font=\bfseries] {$\mathcal{L}_{\text{cur}} + \lambda_{o}\mathcal{L}_{\text{old}} + \lambda_{d}\mathcal{L}_{\text{DWDP}}$};
\end{scope}

\begin{scope}[transparency group, opacity=1, gray]
\draw[-{Triangle[width=16pt,length=16pt]}, line width=14pt]
    (1., -2.3) -- (8.5, -2.3) node[midway, text=white, font=\bfseries]{$\mathcal{L}_{\text{cur}} + \lambda_{o}\mathcal{L}_{\text{old}} + \lambda_{d}\mathcal{L}_{\text{DWDP}}$}; 
\end{scope}

\end{tikzpicture}
\caption{A visualization of the latent representation space (depicted as a sphere) as new tasks are learned sequentially. The points represent data embeddings. The LwP framework organizes representations for new tasks into distinct clusters (colored points) while its preservation loss maintains the geometric structure of prior representations.}
\end{figure*}


In essence, by using a simple and efficient distance preservation loss, we ensure that the representation space remains stable in a high-dimensional feature space, preserving its capability to solve not only previously learned tasks but also potential future ones.

To formalize the alignment objective, we define the loss function \( \mathcal{L}_{pres} \) as the squared Frobenius norm of the difference between the two kernel matrices:

\begin{equation}
\begin{split}
\mathcal{L}_{\text{pres}}(Z, Z') &= \left\| K(Z) - K(Z') \right\|_F^2 \\
\end{split}
\label{eq:loss_pres}
\end{equation}


By minimizing \( \mathcal{L}_{pres} \) while keeping $Z$ fixed, we align the images of \( Z \) and \( Z' \) under the feature map \( \phi \):
\begin{equation}
\langle \phi(\bm{z}_i), \phi(\bm{z}_j) \rangle_{\mathcal{H}} \approx \langle \phi(\bm{z}'_i), \phi(\bm{z}'_j) \rangle_{\mathcal{H}}, \quad \forall i, j.
\end{equation}
This alignment implies that there exists an isometry \( T: \mathcal{H} \to \mathcal{H} \) such that:
\begin{equation}
\phi(\bm{z}'_i) = T(\phi(\bm{z}_i)), \quad \forall i.
\end{equation}


For any function \( f \in \mathcal{H} \), the Riesz representation theorem states that there exists a unique element \( w_f \in \mathcal{H} \) such that \( f(\bm{z}) = \langle w_f, \phi(\bm{z}) \rangle_{\mathcal{H}} \). 
The evaluation of \( f \) at \( \bm{z}'_i \) becomes:

\begin{equation}
f(\bm{z}'_i) = \langle w, \phi(\bm{z}'_i) \rangle_{\mathcal{H}} = \langle w, T(\phi(\bm{z}_i)) \rangle_{\mathcal{H}}.
\label{eq:function_evaluation_z_prime}
\end{equation}

Because \( T \) is an isometry, its adjoint \( T^* \) is also an isometry, and we can write:

\begin{equation}
f(\bm{z}'_i) = \langle T^* w, \phi(\bm{z}_i) \rangle_{\mathcal{H}}.
\label{eq:function_evaluation_adjoint}
\end{equation}

Define \( w' = T^* w \) and \( f'(\bm{z}) = \langle w', \phi(\bm{z}) \rangle_{\mathcal{H}} \). Then:

\begin{equation}
f(\bm{z}'_i) = f'(\bm{z}_i), \quad \forall i.
\label{eq:functions_equivalent}
\end{equation}

Thus, \( Z' \) becomes an alternative representation that is functionally equivalent to \( Z \) in terms of any operations performed within the RKHS induced by the Gaussian kernel. Now, consider a learning problem defined on \( Z \):

\begin{equation}
\min_{f \in \mathcal{H}} \frac{1}{n} \sum_{i=1}^n \ell(f(\bm{z}_i), y_i) + \Omega(f),
\label{eq:learning_problem_Z}
\end{equation}

and the corresponding problem on \( Z' \):

\begin{equation}
\min_{f \in \mathcal{H}} \frac{1}{n} \sum_{i=1}^n \ell(f(\bm{z}'_i), y_i) + \Omega(f).
\label{eq:learning_problem_Z_prime}
\end{equation}

Using the relationship \( f(\bm{z}'_i) = f'(\bm{z}_i) \), the loss terms satisfy $\ell(f(\bm{z}'_i), y_i) = \ell(f'(\bm{z}_i), y_i)$. Since \( \| f \|_{\mathcal{H}} = \| f' \|_{\mathcal{H}} \), the regularization terms are equal: $\Omega(f) = \Omega(f')$. Thus, the risk functionals for the problems on \( Z \) and \( Z' \) are equivalent when considering \( f \) and \( f' \):

\begin{equation}
\frac{1}{n} \sum_{i=1}^n \ell(f(\bm{z}'_i), y_i) + \Omega(f) = \frac{1}{n} \sum_{i=1}^n \ell(f'(\bm{z}_i), y_i) + \Omega(f').
\label{eq:risk_functionals_equivalent}
\end{equation}

Because the risk functionals are equivalent, the optimal solutions \( f^* \) obtained on \( Z' \) correspond to the optimal solutions \( f'^{*} \) on \( Z \) via the isometry \( T^* \):

\begin{equation}
f^*(\bm{z}'_i) = f'^{*}(\bm{z}_i).
\label{eq:optimal_solutions_equivalent}
\end{equation}

This means any model trained on \( Z \) can be transformed to a model on \( Z' \) with identical performance, and vice versa. 

Through empirical observation, we have determined that maintaining the squared Euclidean distance leads to enhanced performance. This is likely because the non-exponentiated distance metric more effectively retains the global structure of the representation. Refer to Section \ref{subsec:ablation_results} for the ablation result. Additionally, in Appendix \ref{appendix:sqdist}, we show that the difference in RBF kernel values is bounded by the difference in the squared $L^2$ norm.




\subsection{Dynamic Weighting} \label{ssec:dynamic_weighting}

\( \mathcal{L}_{\text{pres}} \) is designed to maintain the implicitly learned knowledge of the input data in the representation space. However, in scenarios where there are distinct classes or labels, this loss can conflict with other objectives, such as separating distinct classes.

To address this issue, we introduce the Dynamically Weighted Distance Preservation (DWDP) Loss, \( \mathcal{L}_{\text{DWDP}} \), which applies a dynamic mask \( m_{ij} \) to deactivate preservation for pairs with different labels, preventing conflicts with separation objectives. Unlike PODNet \cite{douillard2020podnet}, which preserves spatial features uniformly across all pairs, RKD \cite{park2019relational}, which maintains all pairwise distances regardless of class, or Asadi et al. \cite{asadi2023prototype}, which preserves only prototype-sample distances, LwP uses dynamic per-batch masking to preserve complete intra-class pairwise structure while avoiding inter-class conflicts. The dynamic mask \( m_{ij} \) is defined as follows:

\begin{equation}
m_{ij} =
\begin{cases}
1, & \text{if } y^{[t]}_i = y^{[t]}_j, \\
0, & \text{otherwise},
\end{cases}
\end{equation}

where $y^{[t]}$ represents the labels of the current task. Thus, the DWDP loss is given by:



\begin{subequations}
\begin{align}
\mathcal{L}_{\text{DWDP}} &=
\frac{1}{N^{2}}\sum_{i,j=1}^{N}
m_{ij}\,(\Delta d_{ij})^{2},\\
\Delta d_{ij} &=
d(\bm z_i,\bm z_j)
- d(\bm z_i',\bm z_j').
\end{align}
\end{subequations}

Consequently, this modification alleviates the objective conflict issue at the cost of reducing the scope for preservation to intraclass sets of the current task. Our detailed pseudo-code algorithm is presented in Appendix \ref{appendix:algo_overview}.

\section{Evaluation}
\label{sec:eval}

\begin{table*}[!htbp]
\centering
\caption{Accuracy Comparison Across Models, Datasets, and Distribution Shifts}
\label{tab:model_accuracy_combined}
\resizebox{\textwidth}{!}{%
\begin{tabular}{@{}l l c c c c c c c c@{}}
\toprule
& & \multicolumn{4}{c}{\textbf{No Shift}} & \multicolumn{4}{c}{\textbf{Shift Scenarios (BDD100k)}} \\
\cmidrule(lr){3-6} \cmidrule(lr){7-10}
\shortstack{\textbf{Method} \\ \textbf{Type}} & \textbf{Model} & \shortstack{\textbf{BDD100k} \\ (3 Tasks)} & \shortstack{\textbf{CelebA} \\ (10 Tasks)} & \shortstack{\textbf{PhysiQ} \\ (3 Tasks)} & \shortstack{\textbf{FairFace} \\ (3 Tasks)} & \shortstack{\textbf{Weather} \\ \textbf{Shift}} & \shortstack{\textbf{Scene} \\ \textbf{Shift}} & \shortstack{\textbf{Time-of-Day} \\ \textbf{Shift}} & \shortstack{\textbf{Combined} \\ \textbf{Shift}} \\
\midrule
STL & - & 75.123 $\pm$ 6.543 & 72.230 $\pm$ 7.297 & 87.167 $\pm$ 10.102 & 64.435 $\pm$ 3.660 & 76.760 $\pm$ 5.210 & 76.787 $\pm$ 5.183 & 76.418 $\pm$ 5.567 & 76.751 $\pm$ 5.180 \\
Naive FT & - & 75.572 $\pm$ 6.382 & 70.068 $\pm$ 9.941 & 81.588 $\pm$ 15.429 & 56.291 $\pm$ 2.221 & 75.281 $\pm$ 5.893 & 73.797 $\pm$ 8.804 & 76.417 $\pm$ 5.609 & 76.029 $\pm$ 5.895 \\
\midrule
\multirow{10}{*}{CL} & LwF & 76.645 $\pm$ 6.577 & 64.626 $\pm$ 10.806 & 69.952 $\pm$ 21.090 & 61.034 $\pm$ 6.162 & 76.794 $\pm$ 5.552& 77.499 $\pm$ 5.215 & 76.031 $\pm$ 6.062 & \textbf{76.938 $\pm$ 5.006} \\
& oEWC & 74.873 $\pm$ 8.375 & 69.666 $\pm$ 9.019 & 82.640 $\pm$ 12.166 & 63.604 $\pm$ 3.122 & 73.529 $\pm$ 9.222& 77.224 $\pm$ 5.057 & 75.885 $\pm$ 5.782 & 74.999 $\pm$ 6.449 \\
& ER & 69.933 $\pm$ 9.112 & 67.598 $\pm$ 7.452 & 76.798 $\pm$ 16.347 & 63.220 $\pm$ 4.730 & 72.287 $\pm$ 7.359& 68.533 $\pm$ 9.071 & 68.331 $\pm$ 8.110 & 67.372 $\pm$ 11.000 \\
& SI & 76.601 $\pm$ 5.277 & 68.735 $\pm$ 10.545 & 83.727 $\pm$ 11.828 & 63.359 $\pm$ 3.451 & 75.848 $\pm$ 6.193& 77.893 $\pm$ 4.239 & 74.818 $\pm$ 5.745 & 74.567 $\pm$ 7.218 \\
& GSS & 75.434 $\pm$ 4.066 & 71.680 $\pm$ 8.468 & 85.741 $\pm$ 10.950 & 64.230 $\pm$ 3.918 & 74.049 $\pm$ 6.793& 74.582 $\pm$ 5.554 & 74.332 $\pm$ 5.541 & 73.520 $\pm$ 7.924 \\
& FDR & 76.779 $\pm$ 6.024 & 69.514 $\pm$ 8.917 & 71.859 $\pm$ 18.687 & 63.709 $\pm$ 3.151 & 76.098 $\pm$ 6.564& 73.623 $\pm$ 11.168 & 75.588 $\pm$ 6.654 & 76.200 $\pm$ 5.847 \\
& DER & 77.183 $\pm$ 5.055 & 70.703 $\pm$ 8.388 & 84.796 $\pm$ 11.168 & 64.114 $\pm$ 3.484 & 76.748 $\pm$ 5.519& 76.166 $\pm$ 6.507 & 76.727 $\pm$ 5.252 & 75.943 $\pm$ 6.326 \\
& DERPP & 76.677 $\pm$ 5.751 & 67.693 $\pm$ 9.425 & 82.838 $\pm$ 13.775 & 63.806 $\pm$ 3.694 & 76.582 $\pm$ 6.071& 77.409 $\pm$ 5.078 & 75.846 $\pm$ 6.411 & 68.581 $\pm$ 14.210 \\
& DVC & 72.683 $\pm$ 4.982 & 71.441 $\pm$ 7.640 & 85.100 $\pm$ 10.381 & 63.848 $\pm$ 3.193 & 72.011 $\pm$ 7.061& 69.830 $\pm$ 7.905 & 70.019 $\pm$ 6.902 & 70.661 $\pm$ 7.100 \\
& OBC & 76.993 $\pm$ 5.118 & 70.829 $\pm$ 8.267 & 83.999 $\pm$ 11.377 & 63.872 $\pm$ 3.449 & 72.270 $\pm$ 14.364& 76.661 $\pm$ 5.988 & 74.835 $\pm$ 6.853 & 73.732 $\pm$ 8.306 \\
\midrule
\multirow{1}{*}{CMTL} & \textbf{LwP} & \textbf{78.299 $\pm$ 3.828} & \textbf{73.484 $\pm$ 8.019} & \textbf{88.242 $\pm$ 12.010} & \textbf{66.482 $\pm$ 3.138} & \textbf{77.937 $\pm$ 4.041}& \textbf{78.198 $\pm$ 3.842} & \textbf{76.820 $\pm$ 5.331} & 74.004 $\pm$ 11.268 \\
\bottomrule
\end{tabular}}
\end{table*}

\begin{figure*}[!htbp]
\centering
\begin{subfigure}[b]{0.18\textwidth}
    \centering
    \includegraphics[width=\textwidth]{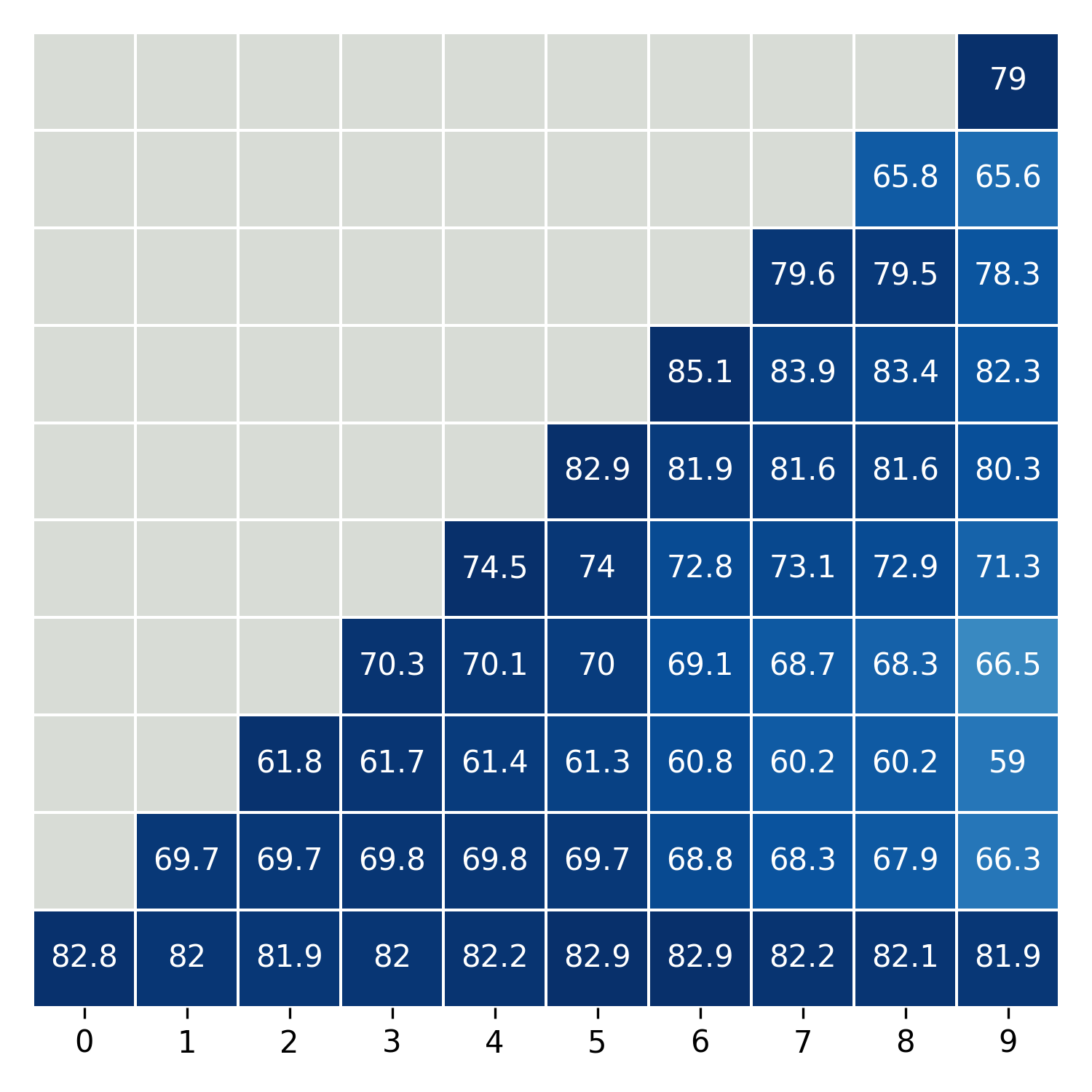}
    \caption{\textbf{LwP}}
    \label{fig:selected_lwp}
\end{subfigure}
\begin{subfigure}[b]{0.18\textwidth}
    \centering
    \includegraphics[width=\textwidth]{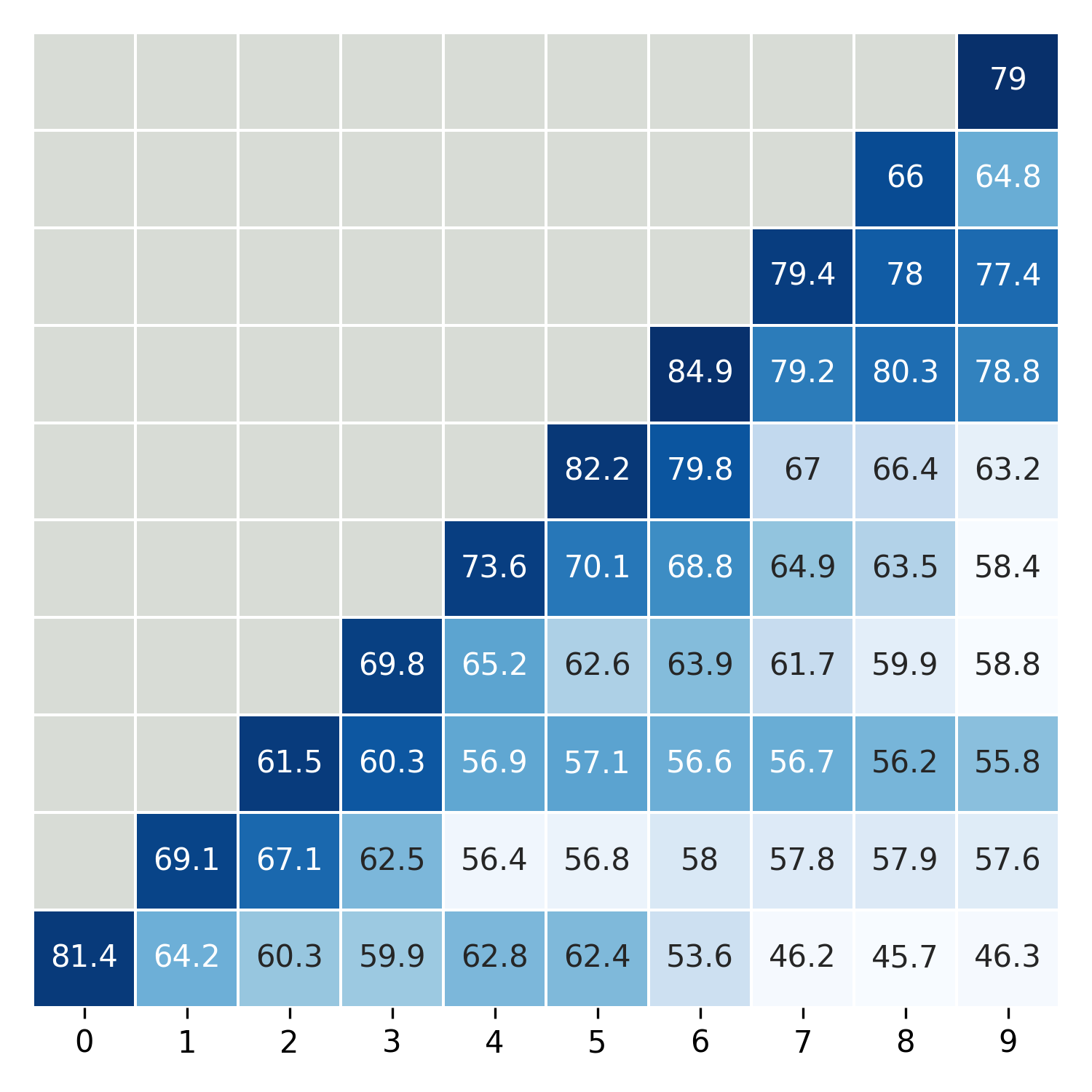}
    \caption{LwF}
    \label{fig:selected_lwf}
\end{subfigure}
\begin{subfigure}[b]{0.18\textwidth}
    \centering
    \includegraphics[width=\textwidth]{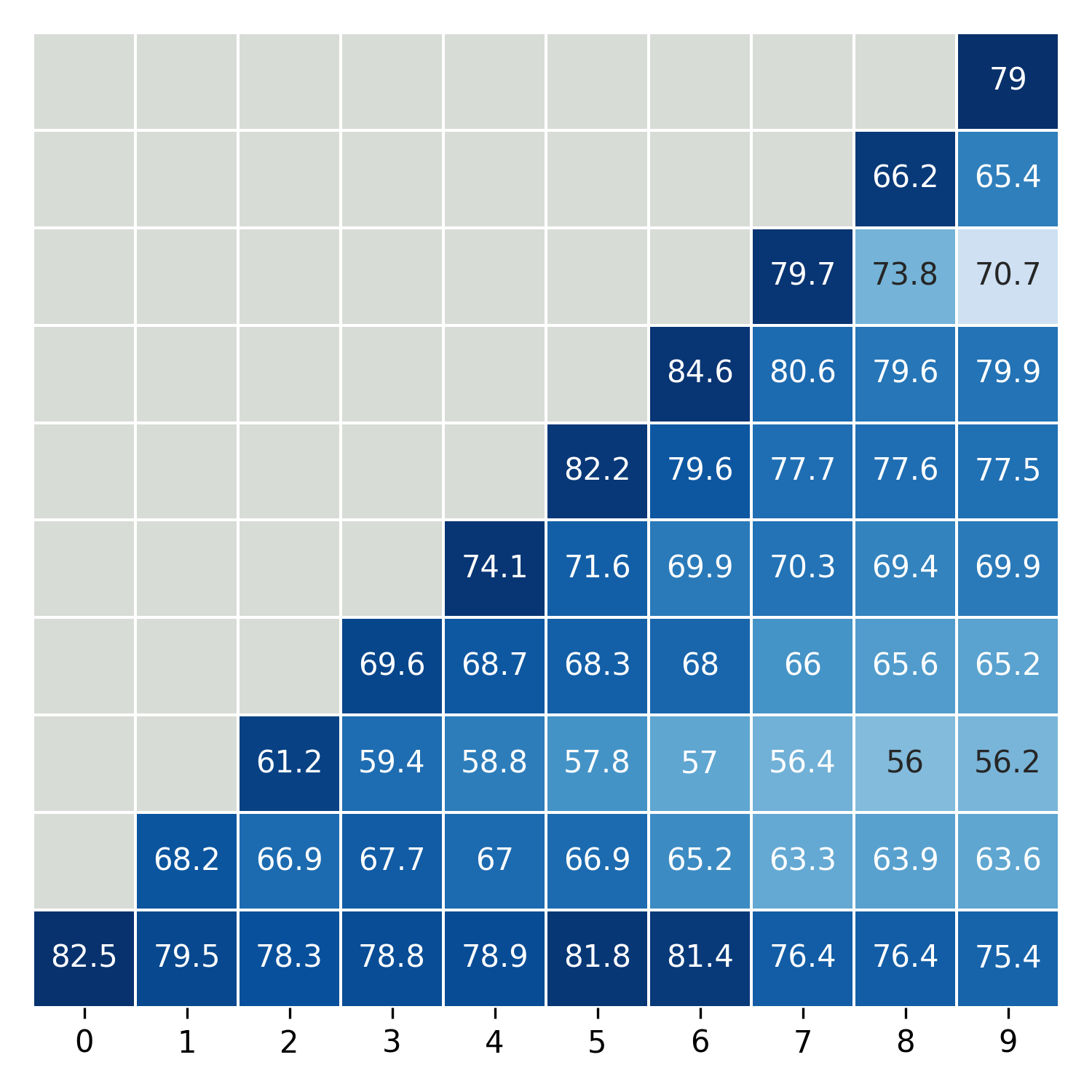}
    \caption{OBC}
    \label{fig:selected_obc}
\end{subfigure}
\begin{subfigure}[b]{0.18\textwidth}
    \centering
    \includegraphics[width=\textwidth]{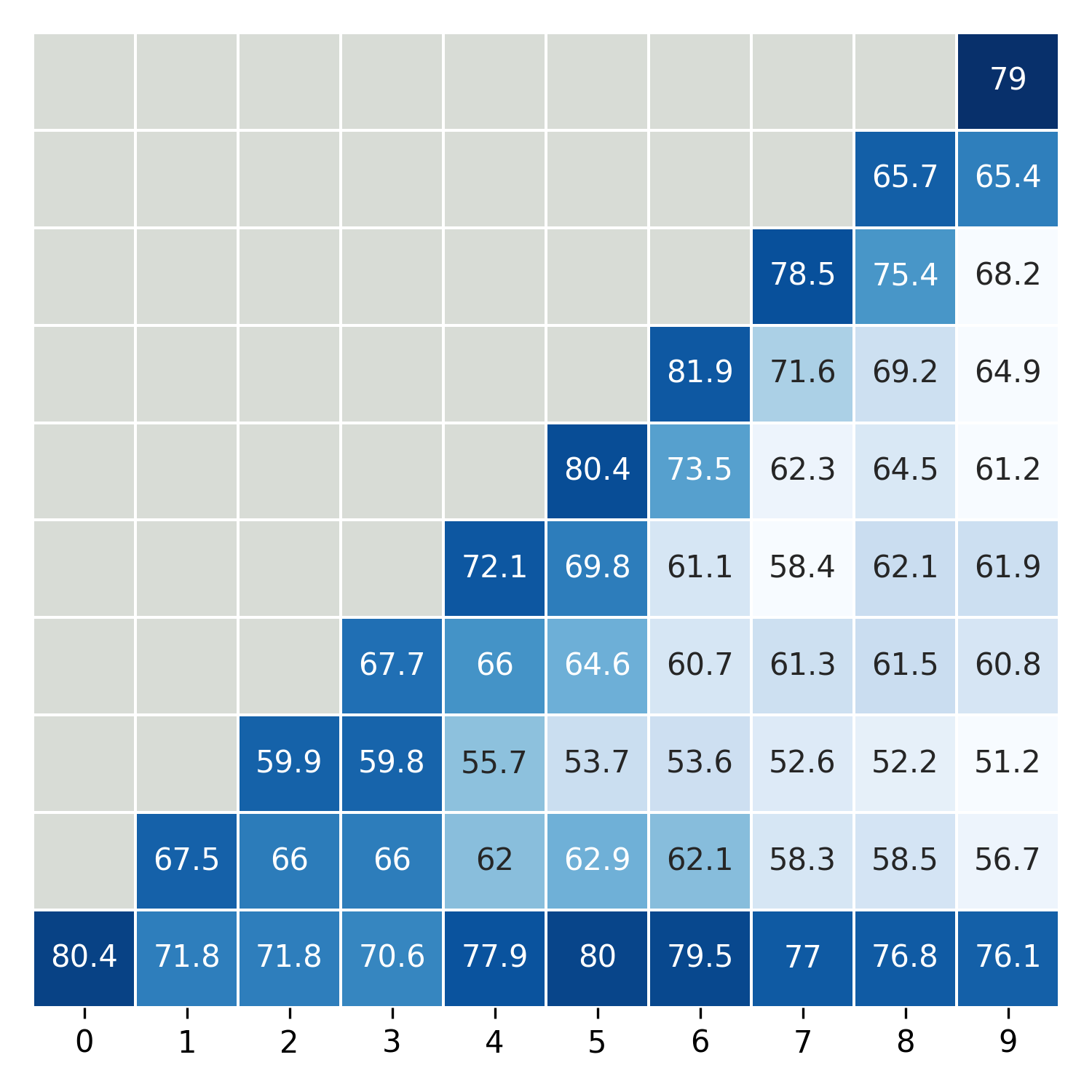}
    \caption{DER}
    \label{fig:selected_der}
\end{subfigure}
\begin{subfigure}[b]{0.18\textwidth}
    \centering
    \includegraphics[width=\textwidth]{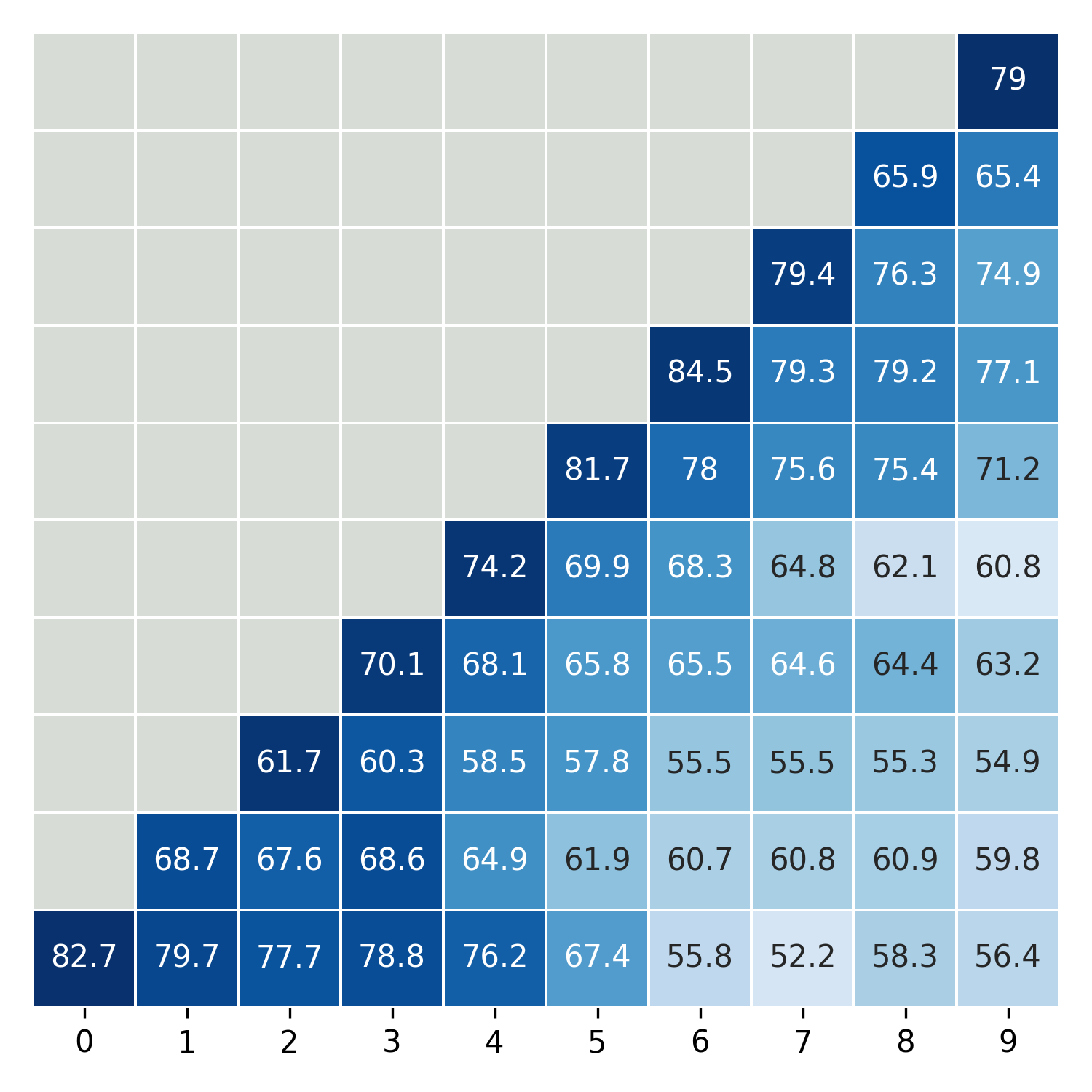}
    \caption{DERPP}
    \label{fig:selected_derpp}
\end{subfigure}

\caption{Selected matrices showcasing the accuracy progression for the dataset CelebA. More details in Appendix \ref{appendix:additional_result}.}
\label{fig:selected_celeba_confusion_matrices}
\end{figure*}

\subsection{Experiment Setup}
We validate our approach through a series of experiments designed to demonstrate LwP's effectiveness. First, we establish LwP's superior performance and its enhanced robustness to distribution shifts against state-of-the-art CL baselines (Sec.~\ref{sec:eval:comprehensive}, \ref{sec:eval:robustness}). We then investigate the underlying reason for this success, showing how LwP mitigates catastrophic forgetting by preserving the geometric structure of the representation space (Sec.~\ref{sec:eval:mitigating}). Finally, we validate our specific design choices and scalability with ablation studies and tests on larger models and higher-resolution inputs (Sec.~\ref{subsec:effect_param_image_size}, \ref{subsec:ablation_results}). The Appendix provides further details on our experimental setup, model architectures, and baselines, along with additional results such as MTL comparisons, accuracy progression over time, and evaluation on training tasks from the MTL to the CMTL setting (Appendix~\ref{appendix:additional_result}).



\paragraph{Dataset} \paragraph{Dataset} We utilize four datasets from two distinct modalities for our task-incremental learning experiments: BDD100K \cite{yu2020bdd100k} (object detection in driving scenes), CelebA \cite{liu2018large} (facial attribute recognition), PhysiQ \cite{wang_physiq_2023, wang2025exact, wang2024microxercise} (IMU-based exercise quality HAR), and FairFace \cite{karkkainenfairface} (facial attribute recognition). Each dataset is structured into a series of tasks. Further details on each dataset, including the attributes, data splits, and preprocessing steps, are provided in the Appendix \ref{appendix:setup_cont}.

\paragraph{Baselines} Our primary emphasis is on CL baselines since integrating many MTL methods into CMTL often requires substantial modifications to accommodate the incremental characteristics of CMTL. For CL, we compare against Online Bias Correction (OBC) \cite{chrysakis2023online}, Dual View Consistency (DVC) \cite{Gu_2022_CVPR},  Dark Experience Replay (DER) \cite{buzzega_dark_2020}, DERPP \cite{boschini2022class}, Function Distance Regularization (FDR) \cite{benjamin_measuring_2019}, Experience Replay (ER) \cite{robins1995catastrophic,ratcliff1990connectionist}, Gradient-based Sample Selection (GSS) \cite{aljundi2019gradient}, online Elastic Weight Consolidation (oEWC) \cite{kirkpatrick_overcoming_2017}, Synaptic Intelligence (SI) \cite{zenke2017continual}, and Learning without Forgetting (LwF) \cite{li_learning_2017}. In addition, we compare our approach with MTL methods, which are detailed in  Appendix \ref{appendix:MTL_comparison}. These include the basic MTL approach of training all tasks simultaneously with different predictors \cite{caruana1997multitask}, as well as more advanced techniques like PCGrad \cite{yu2020gradient}, Impartial MTL (IMTL) \cite{liu2021towards}, and NashMTL \cite{navon2022multi}. We also include a single-task learning (STL) baseline, where each task is learned separately, and a naive fine-tune (FT), where a previously trained model is fine-tuned on the current task. For the choice of distance metric $d$, we test common options such as Euclidean distance and cosine similarity, as well as loss functions designed to preserve relational knowledge, such as those proposed in RKD \cite{park_relational_2019} and Co2L \cite{cha_co2l_2021}. 

\paragraph{Model Architectures} For the image-based datasets (BDD100K, CelebA, and FairFace), we use a ResNet structure as the shared feature extractor $f_{\theta_s}$. For PhysiQ, which consists of time-series data from IMU sensors, we use a 3-layer 1D-CNN model more suited to that modality. This demonstrates the versatility of our LwP framework across different data types and architectures. For all models, each task is handled by a separate linear projection layer (head) applied to the shared representation $\bm{z}$. We evaluate additional architectures and image sizes in Section \ref{subsec:effect_param_image_size}.

\begin{figure*}[!t]
    \centering
    \begin{minipage}[t]{0.010\textwidth}
        \centering
        \vspace*{-7mm}
        \tiny
        \rotatebox{90}{\fontsize{5pt}{6pt}\selectfont BWT}
    \end{minipage}%
    \begin{minipage}[c]{0.9\textwidth}
        \begin{subfigure}[b]{0.33\textwidth}
            \centering
            \includegraphics[width=\textwidth, clip, trim={1cm 0 0 0}]{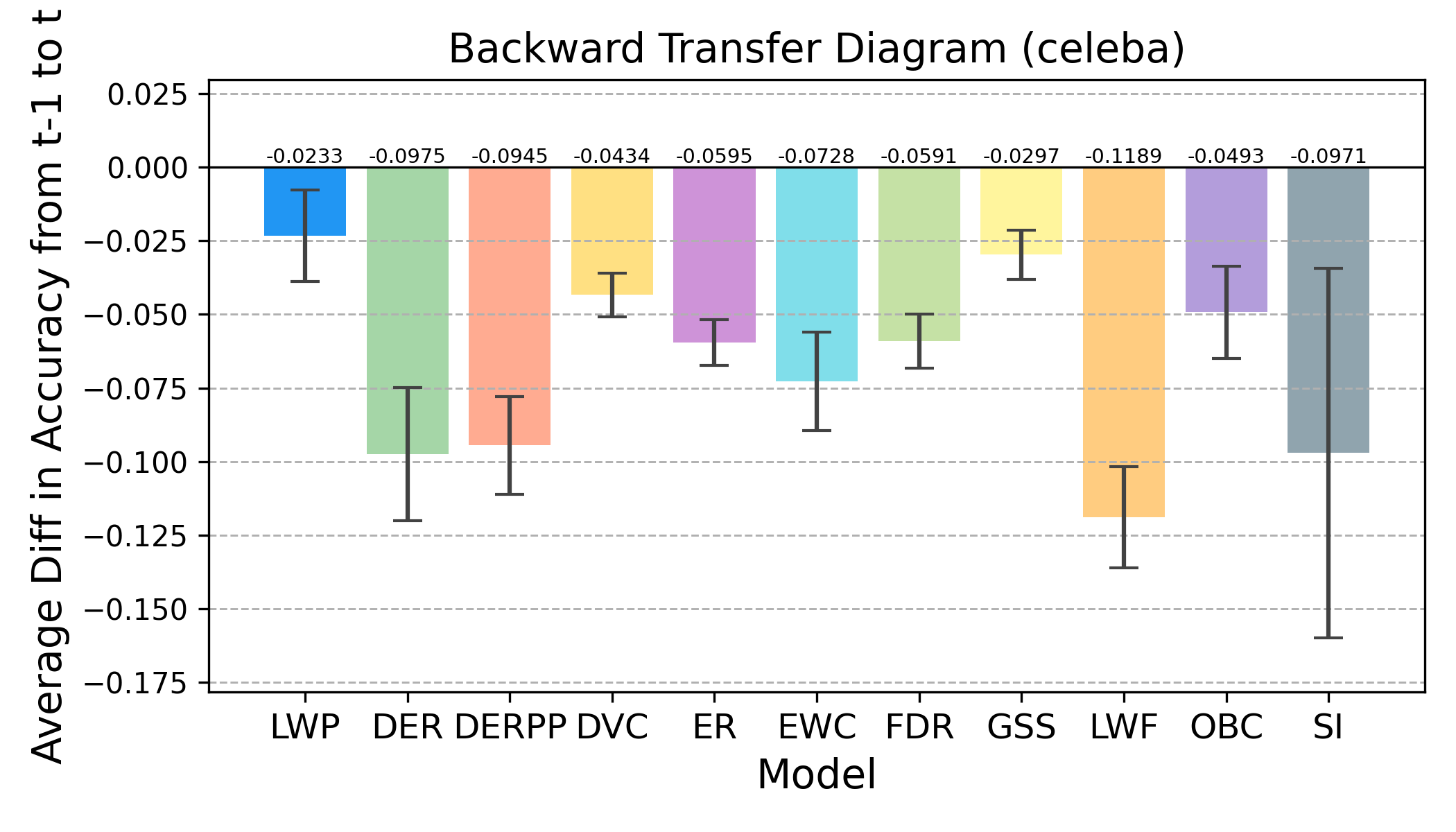}
            \caption{CelebA}
            \label{Fig:celeba}
        \end{subfigure}
        \begin{subfigure}[b]{0.33\textwidth}
            \centering
            \includegraphics[width=\textwidth, clip, trim={1cm 0 0 0}]{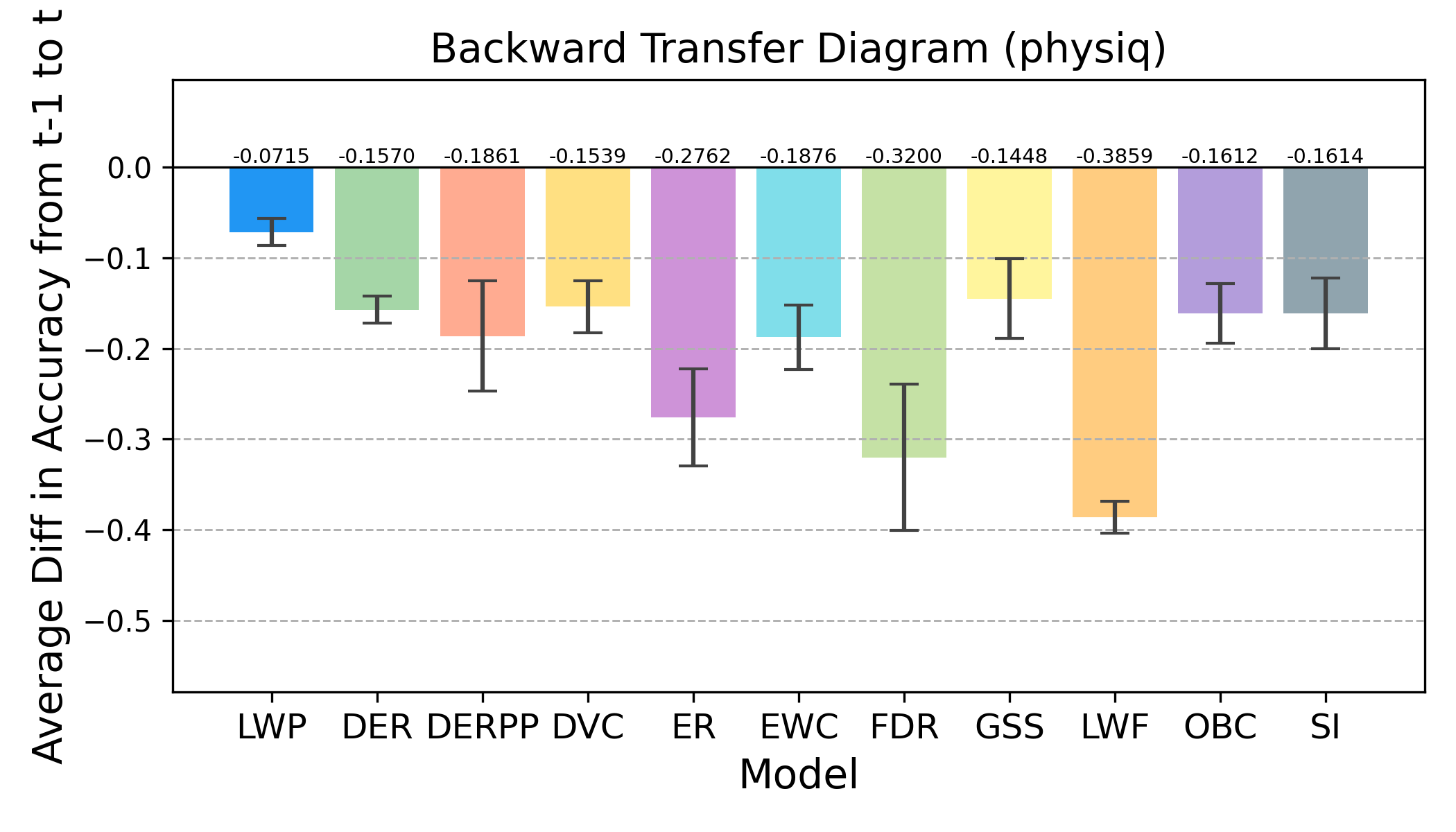}
            \caption{PhysiQ}
            \label{fig:physiq}
        \end{subfigure} 
        \begin{subfigure}[b]{0.33\textwidth}
            \centering
            \includegraphics[width=\textwidth, clip, trim={1cm 0 0 0}]{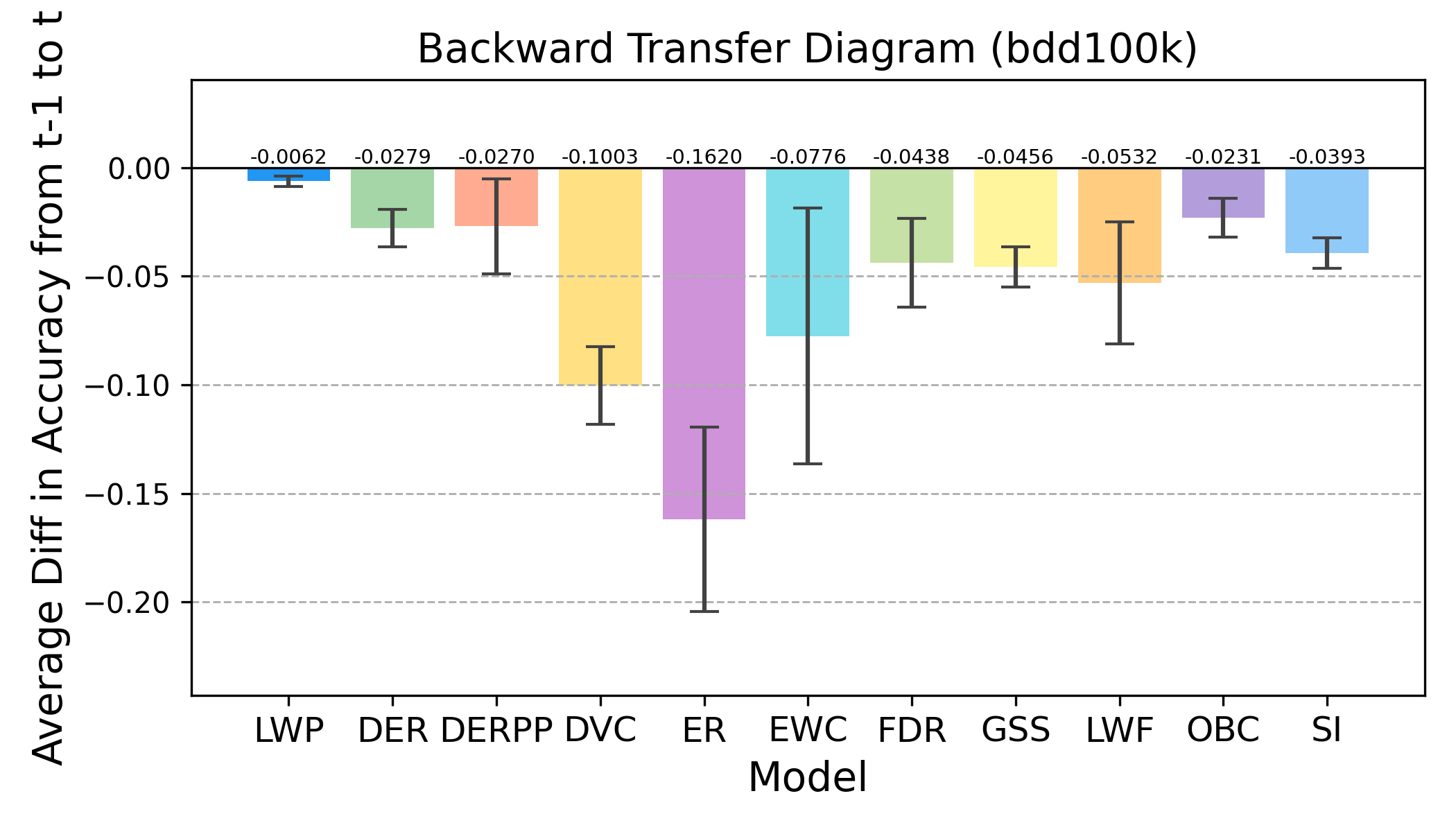}
            \caption{BDD100k}
            \label{fig:bdd100k}
        \end{subfigure}
    \end{minipage}
    
    \caption{Selected Backward Transfer Diagrams for the Benchmark Datasets. More details in Appendix \ref{appendix:acc_per_iter}}
    \label{fig:forward_info_loss_diagrams}
\end{figure*}

\subsection{Comprehensive Performance in CMTL} \label{sec:eval:comprehensive}
In this experiment, we evaluate the performance of our LwP against several state-of-the-art CL methods. All methods, except for LwF and LwP, are provided with a buffer size of 512 for the CelebA and FairFace datasets, and 46 for the PhysiQ dataset, corresponding to approximately 2-3\% of the training set for each dataset. Each model is trained five times using different random seeds. The standard training protocol consists of 20 epochs, with a batch size of 256 for image-based datasets and 32 for PhysiQ, coupled with early stopping. For PhysiQ, we only compare the average accuracy across the final task iteration due to the training instability caused by a smaller dataset size. Table \ref{tab:model_accuracy_combined} reports the average test accuracy, along with the standard deviation over five runs for each method and dataset. Fig. \ref{fig:selected_celeba_confusion_matrices} visualizes the progression of task accuracy in task iterations (left to right). Additional results are provided in Appendix \ref{appendix:accuracy_progression}

Table \ref{tab:model_accuracy_combined} highlights that LwP consistently achieves superior performance across all three benchmarks and is the only method to exceed the Single Task Learning (STL) baseline. This suggests that other continual learning methods likely experience significant task interference. Additional results with MTL are provided in Appendix \ref{appendix:additional_result}. Furthermore, our approach is modality-agnostic, as evidenced by LwP's ability to generalize across different domains. This is demonstrated by the results on the PhysiQ dataset from the IMU sensor domain, which underscores LwP's robustness against challenges unique to non-image-based tasks.
While high final accuracy is a primary goal, the strength of a continual learning method is also measured by its ability to retain knowledge from past tasks. LwP's superior accuracy, particularly its success in surpassing the strong STL baseline, suggests it is more effective at mitigating the task interference that degrades the performance of other methods. This ability to minimize forgetting is analyzed more directly using the Backward Transfer metric in Section \ref{sec:eval:mitigating}.

\subsection{Robustness to Non-Stationary Task Distributions} \label{sec:eval:robustness}

For real-world applications, such as autonomous driving, a model must be robust not only to test-time shifts but also to scenarios where the training data itself is non-stationary. We leveraged the BDD100k dataset's rich annotations to design a CMTL protocol that explicitly simulates this challenge. 

Instead of training on the whole, mixed dataset, we created sequential tasks defined by shifting environmental conditions. For instance, in the Weather Shift scenario, the model was trained sequentially on distinct tasks, where the data for each task came from a specific weather condition (e.g., Task 0: clear, Task 1: rainy, etc.). This protocol directly evaluates the model's ability to handle non-stationary input distributions ($P_{X}^{(t)} \neq P_{X}^{(t+1)}$).

The final three columns of Table \ref{tab:model_accuracy_combined} show the average accuracy of all models trained under these non-stationary protocols. While most models experience a significant performance drop when faced with compounding distribution shifts between tasks, LwP maintains a more significant advantage over the baselines. 
This demonstrates that LwP learns more generalizable and fundamentally robust representations. By preserving the core latent structure, our method is less susceptible to catastrophic forgetting induced by shifting data distributions, a critical advantage for deploying intelligent systems in dynamic, real-world environments.

\subsection{Mitigating Forgetting by Preserving Representation Structure}\label{sec:eval:mitigating}

\begin{figure}[!htbp]
    \centering
    \includegraphics[width=0.53\textwidth]{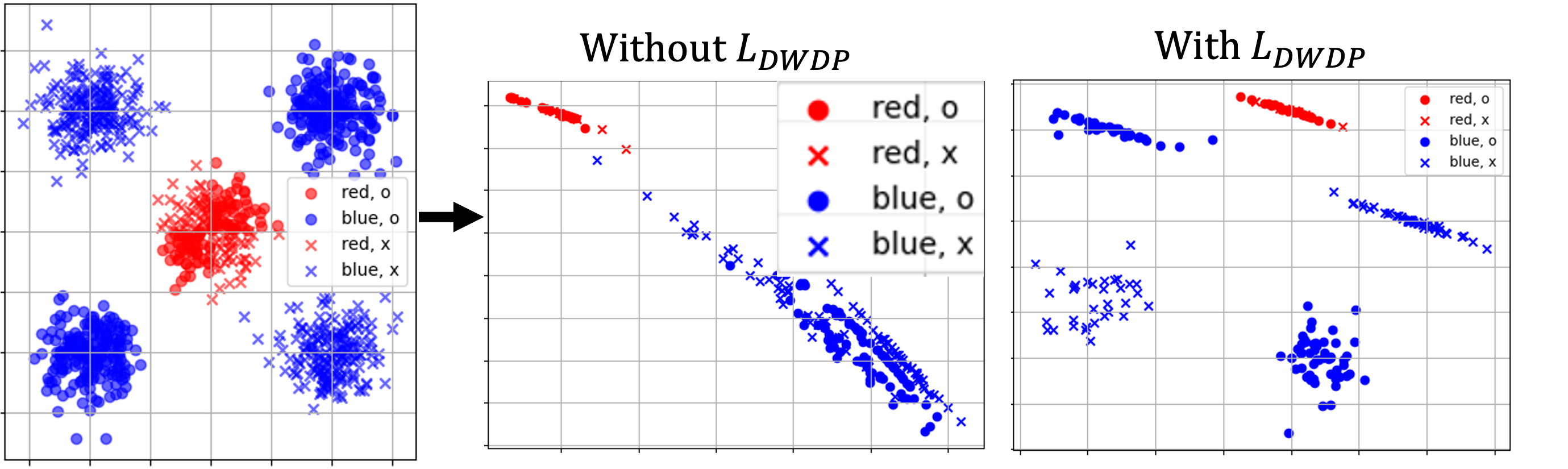}
    \caption{Impact of $\mathcal{L}_{\text{DWDP}}$ on a constructed example. After training on a concentric circle task, the representation space without our loss (left) degrades. With $\mathcal{L}_{\text{DWDP}}$ (right), the structure required for a subsequent XOR task is preserved.}
    \label{fig:toy_example}
\end{figure}

We demonstrate that LwP effectively mitigates catastrophic forgetting in CMTL settings through both empirical evaluation and controlled demonstrations on a constructed dataset. By preserving the structure of the learned representation space, LwP enables the model to retain performance on previously learned tasks, even as it acquires new ones. This structured preservation also enhances the model’s future adaptability: as it accumulates experience across tasks, previously learned features are retained in a generalized form, facilitating more efficient learning on subsequent tasks that share latent structure. This is supported by improvements in the Backward Transfer (BWT) metric across all benchmarks, as well as by the toy example in Fig. \ref{fig:toy_example}, where maintaining latent geometry is critical for success on sequential tasks.

The Backward Transfer \cite{BackwardTransfer} is a metric to evaluate the influence of learning the current task on the performance of previous tasks. A positive backward transfer value indicates that, on average, accuracies on the previous tasks have increased during the current task iteration and vice versa. It is defined as:

\begin{equation}
    BWT = \frac{1}{T-1} \sum_{i=1}^{T-1} R_{T,i} - R_{i, i},
\end{equation}

where $T$ is the index of the current task, $i$ is an index of previous tasks ranging from $1$ to $T-1$, $R_{T,i}$ is the accuracy on task $i$ after training up to task T, and $R_{i, i}$ is the accuracy on task $i$  after learning. As illustrated in Fig. \ref{fig:forward_info_loss_diagrams}, we observe that LwP outperforms all baselines in terms of BWT across all benchmarks. This result is consistent with the visualization, where LwP can maintain the accuracy of each task since its initial training.


\subsection{Effect of Model Parameters and Image Sizes on Performance} \label{subsec:effect_param_image_size}

\begin{table}[!ht]
\centering
\small
\caption{Accuracy Percentage Comparison Across Models on CelebA Dataset (10 Task)}
\label{tab:celeba_model_accuracy}
\begin{tabular}{@{}l l c c c@{}}
\toprule
\shortstack{\textbf{Method} \\ \textbf{Type}} & \textbf{Model} & \shortstack{\textbf{ResNet50 } \\ ($32\times32$)} & \shortstack{\textbf{ResNet101} \\ ($32 \times 32$)} & \shortstack{\textbf{ResNet50} \\ ($224 \times 224$)} \\
\midrule
\multirow{10}{*}{CL}
& LwF & 59.277 $\pm$ 11.920 & 58.279 $\pm$ 11.202 & 60.012 $\pm$ 14.448 \\ 
& oEWC & 66.975 $\pm$ 10.110 & 67.159 $\pm$ 10.506 & 68.511 $\pm$ 13.352 \\
& ER & 65.335 $\pm$ 9.298 & 65.646 $\pm$ 8.784 & 65.973 $\pm$ 14.729 \\
& SI & 66.698 $\pm$ 10.030 & 67.456 $\pm$ 9.880 & 67.747 $\pm$ 13.754 \\
& GSS & 65.926 $\pm$ 13.120 & 65.587 $\pm$ 13.142 & 69.817 $\pm$ 18.771 \\
& FDR & 61.753 $\pm$ 11.943 & 61.720 $\pm$ 12.017 & 65.225 $\pm$ 15.545 \\
& DER & 62.105 $\pm$ 12.114 & 63.797 $\pm$ 10.774 & 69.859 $\pm$ 12.690 \\
& DERPP & 62.814 $\pm$ 11.071 & 62.957 $\pm$ 11.577 & 68.102 $\pm$ 13.557 \\
& DVC & 67.084 $\pm$ 10.380 & 65.340 $\pm$ 11.427 & 70.921 $\pm$ 13.823 \\
& OBC & 64.220 $\pm$ 11.237 & 66.058 $\pm$ 10.370 & 69.319 $\pm$ 13.607 \\
\midrule
\multirow{1}{*}{CMTL}
& \textbf{LwP} & \textbf{67.388 $\pm$ 11.125} & \textbf{69.432 $\pm$ 10.416} & \textbf{85.064 $\pm$ 5.388} \\
\bottomrule
\end{tabular}
\end{table}


Table \ref{tab:celeba_model_accuracy} illustrates that the LwP method scales effectively with increased input resolution and model size. We find that preserving the Gaussian kernel, as shown in eq. \ref{eq:loss_pres}, results in improved performance on larger scales, especially with respect to input resolution. In the ResNet50 benchmark utilizing a 224x224 image size, LwP notably surpasses other baselines by achieving an 85\% accuracy, which is about 15\% percentage points greater than the runner-up. This suggests that, as the input allows the model to create more insightful representations, LwP becomes increasingly advantageous because it can maintain these representations. We also note that the bigger models with the same input size are not performing as well as the one with ResNet18. This is because the inputs lack sufficient information to capture generalized patterns, leading to overfitting.

\subsection{Ablation Study}
\label{subsec:ablation_results}


\begin{table}[ht]
\centering
\caption{Ablation study on the components of the LwP framework on the PhysiQ dataset.}
\renewcommand{\arraystretch}{1.15}
\tabcolsep=0.11cm 
\begin{tabular}{l|cc}
\hline
\textbf{Method on PhysiQ} & \textbf{Dynamic Weighting} & \textbf{w/o Dynamic Weighting} \\
\hline
\textbf{LwP (Full Model)} & \textbf{88.2 $\pm$ 12.0} & 86.0 $\pm$ 12.3 \\
\textbf{LwP w/o $\mathcal{L}_{\text{old}}$} & 87.1 $\pm$ 9.44 & 85.4 $\pm$ 12.1 \\ 
\hline
LwP (Cosine) & 85.4 $\pm$ 13.1 & 84.1 $\pm$ 14.4 \\
LwP (RBF) & 84.5 $\pm$ 13.7 & 84.8 $\pm$ 14.5 \\
IRD (Co2L) & 86.4 $\pm$ 11.5 & 79.9 $\pm$ 17.1 \\
RKD & 85.1 $\pm$ 13.3 & 85.9 $\pm$ 11.9 \\
\hline
\end{tabular}
\label{tab:ablation_main}
\end{table}

To validate our framework, we perform two types of ablation studies. First, we evaluate the design of the proposed DWDP loss function by selectively disabling the dynamic weighting feature and comparing different distance metrics. We include Co2L \cite{cha_co2l_2021}, RKD \cite{park_relational_2019} (which preserves distances across all pairs), cosine similarity, and the RBF kernel (eq. \ref{eq:loss_pres}) as alternatives. The results in Table \ref{tab:ablation_main} confirm our design choices: the squared Euclidean distance combined with Dynamic Weighting yields the best performance, underscoring the importance of preserving global structure while avoiding conflicts with class-separation objectives. 

\begin{figure}[ht]
    \centering
    \includegraphics[width=0.4\textwidth]{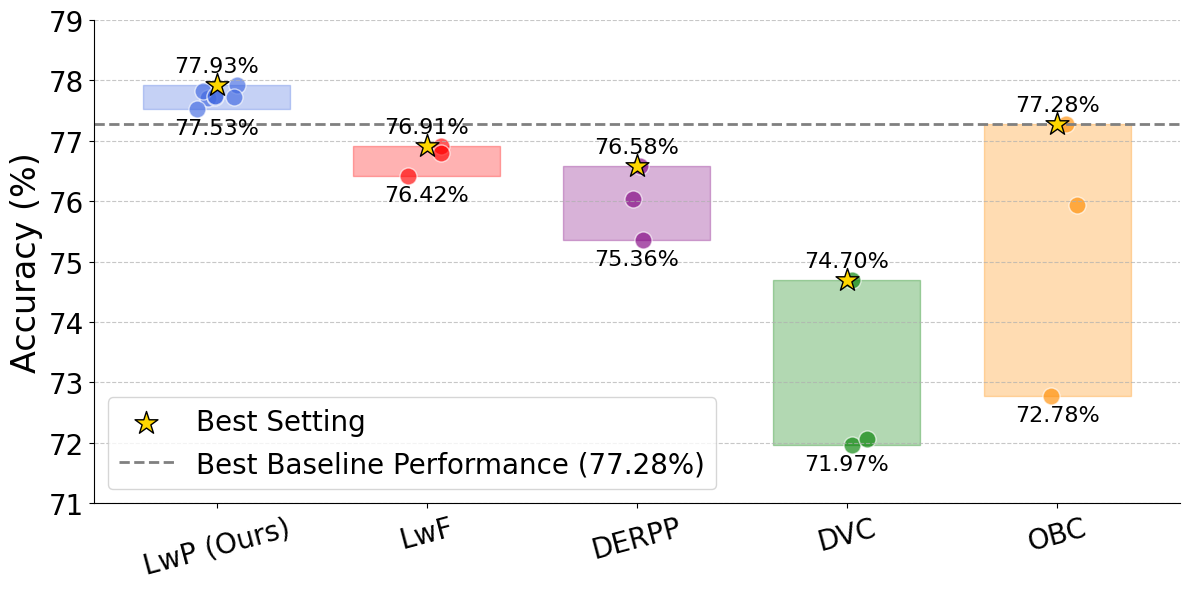}
    \caption{Accuracy on hyperparameter sensitivity analysis on the BDD100k weather shift scenario. } 
    \label{fig:hyperparam_robustness}
\end{figure}

Second, we analyze the model's sensitivity to its key hyperparameters. As shown in Figure~\ref{fig:hyperparam_robustness}, LwP exhibits remarkable robustness. Its performance remains high and stable across a range of hyperparameter values. Crucially, its entire performance range surpasses that of the best-performing baseline, demonstrating a stable advantage that is not dependent on precise tuning. Detailed results of this hyperparameter sweep for both LwP and baseline models are provided in Appendix~\ref{appendix:hyperparameter}.

\section{Related Work}

MTL enhances generalization and computational efficiency by leveraging shared representations across related tasks \cite{caruana1997multitask,sener2018multi,cityspec1,cityspec2}. However, optimizing multiple objectives often presents conflicting gradients. Approaches like the \textbf{Multiple Gradient Descent Algorithm (MGDA)} \cite{sener2018multi} seek Pareto optimal solutions through convex combinations of task-specific gradients, while Gradient Surgery (PCGrad) \cite{yu2020gradient} projects conflicting gradients onto the normal plane of each other to reduce interference. Navon et al. \cite{navon2022multi} modeled gradient combination as a cooperative bargaining game to ensure fairness among tasks. Loss balancing is also crucial, with methods like IMTL \cite{liu2021towards} incorporating both gradient and loss balancing mechanisms. CL enables sequential task learning without catastrophic forgetting \cite{ratcliff1990connectionist,robins1995catastrophic}. Techniques like \textbf{MER \cite{riemer2018learning}} focus on maximizing knowledge transfer while minimizing interference, while HAL \cite{chaudhry2021using} anchors past knowledge to prevent representation drift. Duncker et al. \cite{duncker2020organizing} demonstrate that organizing recurrent network dynamics by task computation enables continual learning. Bridging MTL and CL, CMTL aims to manage performance across sequential and concurrent tasks \cite{wu2023multi}, using methods like MC-SGD \cite{mirzadeh2020linear} to enhance CL by leveraging linear mode connectivity. \textbf{Task-free CL \cite{aljundi2019task}} eliminates task boundaries. More detailed discussions are available in the Appendix \ref{appendix:related_work}.

\section{Conclusion}

We identified and analyzed Continual Multitask Learning (CMTL), a challenging and practical setting within continual learning where conventional methods often fail to perform effectively. 
Our findings show that conventional approaches underperform because they focus on preserving task-specific information while neglecting the broadly applicable, implicit features required to build a unified representation.
To address this, we introduced \textit{Learning with Preserving} (LwP) with a dynamically weighted distance preservation function. This approach maintains the structure of the representation space, preserving implicit knowledge without needing replay buffers, making it especially valuable in privacy-sensitive domains like healthcare. Our experiments across various datasets demonstrated that LwP surpasses state-of-the-art baselines and outperforms single-task models, consistently retaining accuracy and mitigating catastrophic forgetting. The results emphasize the importance of preserving implicit knowledge and the effectiveness of our loss function.

\newpage

\section*{Acknowledgments}
This work was supported in part by the U.S. National Science Foundation under Grants 2427711 and 2443803, the Institute of Education Sciences under Award Numbers R305C240010. The opinions, findings, conclusions, or recommendations expressed in this material are those of the author(s) and do not necessarily reflect the views of the sponsoring agencies.

\bibliographystyle{aaai2026}
\bibliography{references,references2}


\clearpage
\newpage
\appendix

\section{Related Works}\label{appendix:related_work}
\subsection{Multitask Learning}

Multitask learning (MTL) has been extensively explored for its ability to leverage shared representations across multiple related tasks, thereby enhancing generalization and computational efficiency.
In MTL, models are trained on multiple tasks simultaneously, with the assumption that learning tasks together allows the model to capture commonalities and differences among tasks, leading to better performance than training each task separately. 

A central challenge in MTL is the optimization of multiple objectives, which often present conflicting gradients that can impede the convergence and performance of the model. To address this, various gradient balancing approaches have been proposed. \cite{sener2018multi} introduced the Multiple Gradient Descent Algorithm (MGDA), which seeks Pareto optimal solutions by finding a convex combination of task-specific gradients. Building on this, \cite{yu2020gradient} proposed Gradient Surgery (PCGrad), which directly modifies conflicting gradients by projecting them onto the normal plane of each other to reduce negative interference. More recently, \cite{navon2022multi} approached MTL from a game-theoretic perspective, modeling the gradient combination step as a cooperative bargaining game and employing the Nash Bargaining Solution to ensure proportional fairness among tasks. 

In addition to gradient balancing, loss balancing is crucial for achieving stable and unbiased learning in MTL. \cite{liu2021towards} introduced IMTL, which incorporates both gradient and loss balancing mechanisms. Their method, IMTL-G, ensures unbiased updates to task-shared parameters by finding the geometric angle bisector of task gradients. At the same time, IMTL-L automatically learns loss weighting parameters to harmonize the scales of different task losses.

While these MTL methods have advanced the ability to learn multiple tasks simultaneously, they typically assume that all task data is available at training time and can be processed jointly. This assumption does not hold in scenarios where tasks and their associated data arrive sequentially, as is the case in our defined problem, Continual Multitask Learning (CMTL). In such cases, models must learn new tasks without access to all previous data, and ideally, they should utilize new tasks to enhance their performance on prior tasks.

Our work differs from traditional MTL approaches in that it addresses the sequential arrival of tasks and data, where tasks are learned iteratively rather than simultaneously. Unlike MTL methods that focus on balancing gradients and losses across tasks trained together, our approach must handle the challenge of incorporating new tasks without retraining on the data of previous tasks. Furthermore, we introduce mechanisms to utilize new task data to enhance the model's generalizability on earlier tasks, which is not typically considered in standard multi-task learning (MTL) frameworks.


Rather than creating entirely new frameworks, some research focuses on improving existing models. For instance, \cite{park_relational_2019} proposes a method for transferring knowledge via the loss function, known as relational knowledge distillation, which enhances the latent space. Similarly, \cite{guo_improved_2017} and \cite{avidan_continual_2022} concentrate on structurally maintaining the embedding space, ensuring its robustness over time.


\subsection{Continual Learning (CL)}
CL aims to enable models to learn sequentially from a stream of tasks without forgetting previously acquired knowledge, addressing the challenge of catastrophic forgetting~\cite{ratcliff1990connectionist,robins1995catastrophic}. Various methods have been developed to tackle this problem, broadly categorized into \textit{rehearsal-based methods}, \textit{knowledge distillation}, and \textit{regularization-based techniques}.


\paragraph{Rehearsal-based methods} Early works such as \cite{ratcliff1990connectionist,robins1995catastrophic} introduced Experience Replay (ER), where old data samples are mixed with current ones during training. Building upon this concept, Robins \cite{robins1995catastrophic} explored pseudorehearsal techniques. More recent methods like Meta-Experience Replay (MER) \cite{riemer2018learning} reformulate ER within a meta-learning framework, aiming to enhance knowledge transfer between past and present tasks while reducing interference. 
Gradient-based Sample Selection (GSS) \cite{aljundi2019gradient} modifies ER by selecting optimal examples for storage in the memory buffer, improving retention of past knowledge.
Lastly, Aljundi et al. \cite{aljundi2019task} introduce task-free continual learning, eliminating the need for task boundaries and enabling more flexible adaptation to new tasks without explicit task identifiers.

These methods, while effective in certain scenarios, rely heavily on storing and replaying data from previous tasks, which may not be feasible due to privacy concerns or memory constraints. In contrast, our approach does not require storing raw data from previous tasks. Instead, we utilize pseudolabels generated by the frozen previous model and introduce a novel regularization term to preserve the structure of shared representations. This enables the model to retain and improve upon prior knowledge without explicit rehearsal.


\paragraph{Knowledge Distillation} Methods leveraging Knowledge Distillation \cite{hinton2015distilling} address the issue of forgetting by using a previous iteration of the model as a teacher. Learning Without Forgetting (LwF) \cite{li2017learning} generates a softened version of the model’s current outputs on new data at the onset of each task, minimizing output drift throughout training. iCaRL \cite{rebuffi2017icarl} combines distillation with replay techniques, using a memory buffer to train a nearest-mean-of-exemplars classifier while applying a self-distillation loss to preserve learned representations across tasks. Moreover, Li et al. \cite{li2019compositional} proposed a continual learning method tailored for sequence-to-sequence tasks, leveraging compositionality to enable knowledge transfer and prevent catastrophic forgetting. Their approach extends traditional label prediction continual learning methods to handle more complex tasks like machine translation and instruction learning. 

While these methods use knowledge distillation to maintain performance on old tasks, they typically focus on preserving output logits or feature representations without considering the underlying relational structure between data points. Our method extends this idea by not only preserving the output predictions via pseudolabels but also maintaining the pairwise relationships in the representation space through our Dynamically Weighted Distance Preservation (DWDP) loss. This helps in better retaining the learned structure and prevents the model from drifting away from previously acquired knowledge.


\paragraph{Regularization-based techniques} These methods modify the loss function to include a penalty that restricts changes to the model's parameters. Examples include Elastic Weight Consolidation (EWC) \cite{kirkpatrick_overcoming_2017}, its online variant (oEWC) \cite{schwarz2018progress}, and Synaptic Intelligence (SI) \cite{zenke2017continual}. Complementary to parameter-based regularization, Duncker et al. \cite{duncker2020organizing} demonstrate how organizing network dynamics by task computation can enable continual learning in recurrent networks.
Moreover, Adel et al. \cite{adel2019continual} introduced Continual Learning with Adaptive Weights (CLAW), which employs a probabilistic modeling approach to adaptively identify which parts of the network should be shared across tasks in a data-driven manner. This method balances between modeling each task separately to prevent catastrophic forgetting and sharing components to allow transfer learning and reduce model size. 

Our approach differs from these methods as we do not rely on parameter regularization or expanding architectures. Instead, we focus on preserving the learned representations and their relational structure between tasks through the DWDP loss, which provides a more scalable solution without incurring additional memory overhead.

\subsection{Continual and Multitask Learning}
Our work distinguishes itself from existing approaches in CMTL by introducing a new problem domain where new tasks and their associated datasets arrive incrementally. In this setting, the model is not only required to adapt to new tasks but also to utilize these new datasets to enhance its performance on previous tasks. Specifically, when new data for additional tasks becomes available, it is used to further train the existing model. This training process enables the model to reinforce and improve its understanding of prior tasks, effectively allowing it to remember and perform better on both past and current tasks.

Building upon the extensive research in multitask learning (MTL)~\cite{caruana1997multitask,sener2018multi,yu2020gradient,navon2022multi,li2017learning} and continual learning (CL)~\cite{ratcliff1990connectionist,robins1995catastrophic,riemer2018learning,aljundi2019gradient,chaudhry2021using,li2017learning}, the emerging field of continual multitask learning seeks to bridge the two paradigms to effectively manage performance across sequential and concurrent tasks~\cite{wu2023multi,chen2025logidebrief,chen2024auto311}. 

One of the most related works to ours, Mirzadeh et al.~\cite{mirzadeh2020linear}, focus on the linear mode connectivity between solutions obtained through sequential and simultaneous training. While they demonstrate that a linear path of low error exists for more than twenty tasks and introduce algorithms like Mode Connectivity SGD (MC-SGD) to enhance continual learning, their work does not address the use of \textit{new tasks} to improve performance on previous ones, particularly using a similar setup to traditional continual learning, which means their works fit more on the realm of CL.

Similarly, Liao et al.~\cite{liao2022muscle} propose MUSCLE, a multitask self-supervised continual learning framework designed to pre-train deep models on diverse X-ray datasets. This work, similar to ours, operates in the domain of medical imaging to process classification and segmentation in different body areas. However, their work differs from ours because their focus is on pre-training the model on different tasks for better generalization, which they refer to as ``multitask continual learning.'' We specifically differentiate our CMTL approach from theirs in that our tasks are seen \textit{iteratively}; we do not have access to all tasks at the same time, and the tasks themselves could be orthogonal to previously seen tasks.

In summary, our approach introduces a novel aspect to CMTL by leveraging new tasks and their data not only to learn the new tasks but also to generalize on prior tasks, all within an iterative framework where tasks arrive sequentially and are potentially unrelated. This sets our work apart from existing CMTL methods, which typically do not utilize new tasks to enhance previous ones in this manner.


\newpage

\section{Learning with Preserving Algorithm Overview}\label{appendix:algo_overview}
In this section, we present the pseudocode for our algorithm presented in Section \ref{section:lwp}.

\begin{algorithm}[!ht]
\caption{Learning with Preserving (LwP)}
\label{alg:lwp}
\begin{algorithmic}[1]
\State \textbf{Input:} Sequence of tasks $\{D_t\}_{t=1}^T$, hyperparameters $\lambda_{\text{c}}$, $\lambda_{\text{o}}$, $\lambda_{\text{d}}$
\State \textbf{Output:} Final model parameters $\theta^{[T]}$

\State Initialize model parameters $\theta^{[0]}$

\For{$t = 1$ to $T$}
    \State Set $\theta^{[t]} \gets \theta^{[t-1]}$
    \State Add new task-specific head $g_{\theta_t}$ for task $t$
    \State Freeze all parameters from the previous step ($\theta^{[t-1]}$)
    \For{each mini-batch $\{(\bm{x}_i, y_i^t)\}_{i=1}^N$ in $D_t$}
        \State // \textbf{Compute representations and outputs}
        \State $\bm{z}_i^{[t]} \gets f_{\theta_s^{[t]}}(\bm{x}_i)$ \quad  \textit{// current representation}
        \State $\hat{y}_i^t \gets g_{\theta_t^{[t]}}(\bm{z}_i^{[t]})$  \quad \textit{// current task output}
        \State $\bm{z}_i^{[t-1]} \gets f_{\theta_s^{[t-1]}}(\bm{x}_i)$ \quad \textit{// frozen representation}

        \State // \textbf{Compute outputs for previous tasks}
        \For{$o = 1$ to $t-1$}
            \State $\hat{y}_i^o \gets g_{\theta_o^{[t]}}(\bm{z}_i^{[t]})$ \quad \textit{// current old task output}
            \State $\tilde{y}_i^o \gets g_{\theta_o^{[t-1]}}(\bm{z}_i^{[t-1]})$ \quad \textit{// pseudolabel}
        \EndFor

        \State // \textbf{Compute losses}
        \State $\mathcal{L}_{\text{cur}} \gets \mathcal{L}_{\text{cur}}\bigl(y_i^t, \hat{y}_i^t\bigr)$
        \State $\mathcal{L}_{\text{old}} \gets \sum_{o=1}^{t-1} \mathcal{L}_{\text{old}}\bigl(\tilde{y}_i^o, \hat{y}_i^o\bigr)$ 
        \State $\mathcal{L}_{\text{DWDP}} \gets \mathcal{L}_{\text{DWDP}}(\bm{z}^{[t]}, \bm{z}^{[t-1]}, y_t)$

        \State // \textbf{Combine all losses into LwP loss}
        \[
            \mathcal{L}_{\text{lwp}}
            \;=\; \lambda_{\text{c}} \,\mathcal{L}_{\text{cur}}
            \;+\; \lambda_{\text{o}} \,\mathcal{L}_{\text{old}}
            \;+\; \lambda_{\text{d}} \,\mathcal{L}_{\text{DWDP}}.
        \]

        \State // \textbf{Update parameters}
        \State $\theta^{[t]} \gets \theta^{[t]} - \eta\, \nabla_{\theta^{[t]}}\mathcal{L}_{\text{lwp}}$
    \EndFor
\EndFor
\end{algorithmic}
\end{algorithm}
\newpage

\section{On Using Euclidean Distance \ref{pres_impl_knowledge}}
\label{appendix:sqdist}

Here, we show that preserving the squared Euclidean distances between the data points in \( Z \) and \( Z' \) is sufficient to achieve the same effect.

\paragraph{Squared Euclidean Distance Preservation}

We define the squared Euclidean distance between two points \( z_i \) and \( z_j \) as:
\begin{equation}
D_{ij}(Z) = \| z_i - z_j \|^2.
\label{eq:euclidean_distance}
\end{equation}
Similarly, we compute \( D_{ij}(Z') \) for \( Z' \).

Our goal is to minimize the difference between the squared distances in \( Z \) and \( Z' \), which we formalize with the following loss function:
\begin{equation}
\mathcal{L}_{\text{dist}}(Z, Z') = \sum_{i=1}^{n} \sum_{j=1}^{n} \left( \| z_i - z_j \|^2 - \| z'_i - z'_j \|^2 \right)^2.
\label{eq:loss_distance}
\end{equation}
Minimizing \( \mathcal{L}_{\text{dist}} \) with respect to \( Z' \) encourages the squared distances between all pairs of points in \( Z' \) to match those in \( Z \):
\begin{equation}
\| z'_i - z'_j \|^2 \approx \| z_i - z_j \|^2, \quad \forall i, j.
\label{eq:distance_alignment}
\end{equation}

Since the exponential function is Lipschitz continuous on compact subsets, small changes in the squared distance result in small changes in the kernel value. Specifically, if the squared distances are preserved within a small error \(\epsilon > 0\):
\begin{equation}
\left| \| z_i - z_j \|^2 - \| z'_i - z'_j \|^2 \right| < \epsilon,
\end{equation}
then the difference in the kernel values can be bounded:

\begin{equation}
\label{eq:kernel-bound}
\begin{aligned}
&\left|\,\| z_i - z_j \|^2 \;-\; \| z'_i - z'_j \|^2 \right|
  \;<\; \epsilon
\\[6pt]
&\implies\quad
\left|\,k(z_i, z_j) \;-\; k(z'_i, z'_j)\right|
\\[4pt]
&\quad
=\;
\Bigl|\,
  \exp\!\Bigl(-\tfrac{\|z_i - z_j\|^2}{2\,\sigma^2}\Bigr)
  \;-\;
  \exp\!\Bigl(-\tfrac{\|z'_i - z'_j\|^2}{2\,\sigma^2}\Bigr)
\Bigr|
\\[5pt]
&\quad
\le
\frac{1}{2\,\sigma^2}\,
\exp\!\Bigl(
   -\tfrac{\min\bigl(\|z_i - z_j\|^2,\;\|z'_i - z'_j\|^2\bigr)}{2\,\sigma^2}
\Bigr)
\\
&\quad\quad\times\;
\Bigl|\,
  \|z_i - z_j\|^2 \;-\; \|z'_i - z'_j\|^2
\Bigr|
\\[6pt]
&\quad
\le
\frac{1}{2\,\sigma^2}\,\epsilon
\\[4pt]
&\quad
\le
L_k\,\epsilon.
\end{aligned}
\end{equation}

where \( L_k \) is a Lipschitz constant dependent on \( \sigma \).

Therefore, preserving the squared Euclidean distances between \( Z \) and \( Z' \) implies that the Gaussian kernel matrices \( K(Z) \) and \( K(Z') \) are approximately equal:
\begin{equation}
k(z_i, z_j) \approx k(z'_i, z'_j), \quad \forall i, j.
\label{eq:kernel_approximation}
\end{equation}


\section{Additional Details on Experiments} \label{appendix:additional_result}
\subsection{Experiments Setup Cont.}\label{appendix:setup_cont}

\paragraph{Datasets} We utilize four datasets from two distinct modalities, each structured for task-incremental learning. In this setting, each task is only exposed to a subset of training samples:

\textit{The BDD100K dataset} \cite{yu2020bdd100k} is a large-scale, diverse driving dataset containing over 100,000 images. For our purposes, we focus on three object detection tasks: "traffic sign", "person", and "truck." These classes were selected to minimize imbalance. The input images are resized to 224x224. A key feature of this dataset is its rich annotation of environmental conditions, which we leverage to validate our model's robustness to distribution shifts across different weather types (e.g., clear, rainy, snowy) and scene types (e.g., city street, highway), as well as their combinations.

\textit{The CelebA dataset} \cite{liu2018large}, consisting of 200,000 images with 40 facial attributes. For our work, we focus on 10 of the most balanced attributes. The train dataset is equally subdivided for each task, leading to 20,000 images per task. For simplicity, input images are resized to 32x32.

\textit{The PhysiQ dataset} \cite{wang_physiq_2023}, which contains approximately 4,500 samples collected using inertial measurement units (IMUs) to capture the quality of physical exercises. The data is collected on accelerometer and gyroscope modality at a 50 Hz sampling rate for 31 participants with three attributes. Each task corresponds to one of these attributes with around 1,500 samples.

\textit{The Fairface dataset} \cite{karkkainenfairface}, which includes 100,000 images with three attributes. Following the same subdivision procedure, the dataset results in approximately 33,333 images with a resolution of 128x128 per task. Not only do the tasks differ from those of CelebA, but also the images are not resized in order to show that our approach is scalable.

\subsection{Hyperparameters}

In the following section, we provide an extensive description of the hyperparameters utilized during the training phase. Across all datasets and models, the Adam optimizer \cite{kingma2017adammethodstochasticoptimization} was employed universally. For the CelebA and FairFace datasets, a consistent learning rate of 0.0001 was maintained, coupled with a batch size configuration of 256. In contrast, for the PhysiQ dataset, a higher learning rate of 0.01 was utilized alongside a smaller batch size of 32. Furthermore, we adhered to fixed model-specific hyperparameters for all datasets and models to ensure uniformity and consistency, including the LwP parameters. In the case of LwP, the parameters set as follows: $\lambda_n$ as a value of 1, $\lambda_o$ as a value of 1, and $\lambda_d$ with a value of 0.01. Additionally, the 10 tasks used for CelebA are wearing lipsticks, smiling, mouth slightly open, high cheekbones, attractive, heavy makeup, male, young, wavy hair, and straight hair. PhysiQ dataset includes three attributes assessing exercise quality: stability, range of motion, and exercise variation.

Details of all models and their hyperparameter selection have been documented in the codebase at [ANONYMOUS LINK].

\subsection{Representation Space Visualization via t-SNE}

\begin{figure}[htbp]
\centering
\begin{subfigure}[b]{0.22\textwidth}
    \centering
    \includegraphics[width=\textwidth]{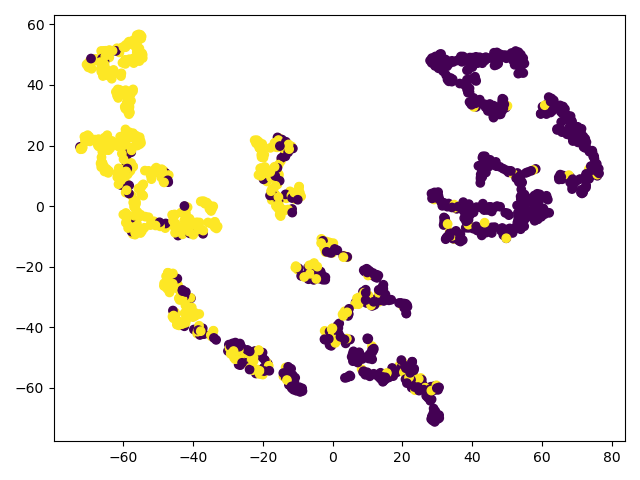}
    \caption{$\bm{z}$ of LwP, after $\mathcal{T}_0$}
    \label{fig:lwp_tsne_task0}
\end{subfigure}
\hspace{0.02\textwidth} 
\begin{subfigure}[b]{0.22\textwidth}
    \centering
    \includegraphics[width=\textwidth]{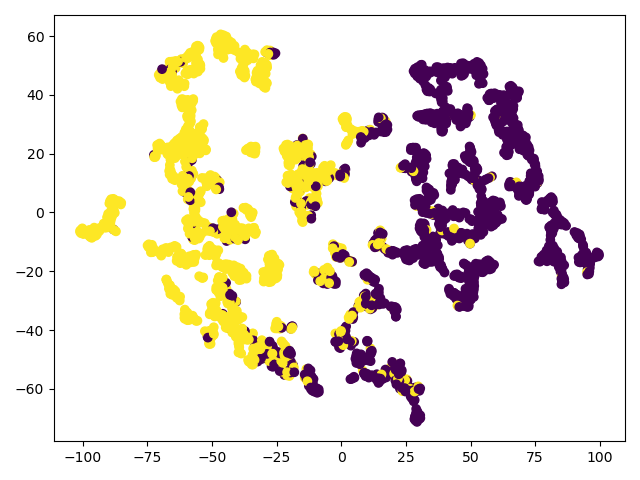}
    \caption{$\bm{z}$ of LwP, after $\mathcal{T}_1$}
    \label{fig:lwp_tsne_task1}
\end{subfigure}
\hspace{0.02\textwidth} 
\begin{subfigure}[b]{0.22\textwidth} 
    \centering
    \includegraphics[width=\textwidth]{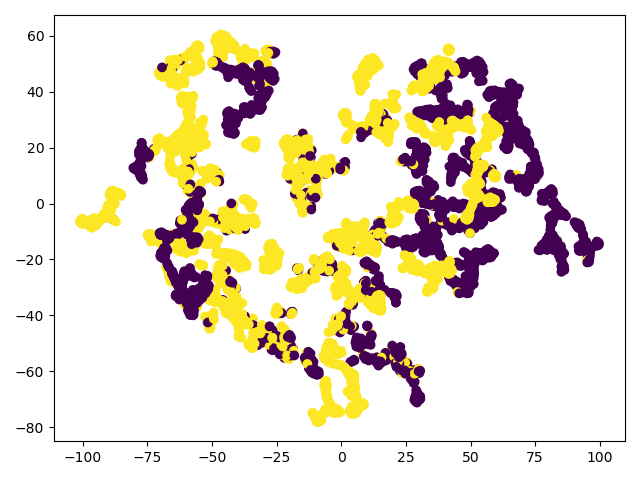}
    \caption{$\bm{z}$ of LwP, after $\mathcal{T}_2$}
    \label{fig:lwp_tsne_task2}
\end{subfigure}
\caption{Representation space progression over task iteration. Colors indicate different label values.}
\label{fig:representation_tsne}
\end{figure}

Fig. \ref{fig:representation_tsne} shows the 2D visualization of the representation $\bm{z}$ for each model trained on the PhysiQ dataset. It was constructed using the dimensionality reduction algorithm t-SNE \cite{tsnepaper} on the PhysiQ test dataset at the end of each task iteration. x and y axes represent the two new dimensions created by the algorithm to project the high-dimensional data onto a 2D plane. Note these dimensions do not have an intrinsic meaning and rather constructed to reflect the relative distances between data points in the high-dimensional space. As demonstrated, the $\bm{z}$ produced by LwP maintains coherent cluster formations as it progressively learns new tasks without introducing considerable distortions when compared to the baseline model. This behavior is comparable to the example provided with the toy dataset depicted in Fig. \ref{fig:toy_example}.

\subsection{Comparison with MTL methods}\label{appendix:MTL_comparison}

\begin{table}[!ht]
\centering
\small
\caption{Comparison of Accuracy across Different Models and Datasets}
\label{tab:comparison}
\begin{tabular}{@{}l l c c c@{}}
\toprule
\textbf{Method Type} & \textbf{Model} & \textbf{CelebA} & \textbf{PhysiQ} & \textbf{FairFace} \\
\midrule
STL & - & 72.230 $\pm$ 7.297 & 87.167 $\pm$ 10.102 & 64.435 $\pm$ 3.660 \\
\midrule
\multirow{4}{*}{MTL} & MTL & \textbf{76.526 $\pm$ 7.616} & \textbf{93.536 $\pm$ 5.739} & 71.418 $\pm$ 4.169 \\
                     & PCGrad & 75.506 $\pm$ 8.146 & 91.910 $\pm$ 8.491 & 70.061 $\pm$ 4.892 \\
                     & IMTL & 76.280 $\pm$ 7.248 & 92.661 $\pm$ 6.617 & 71.399 $\pm$ 3.887 \\
                     & NashMTL & 75.506 $\pm$ 8.146 & 91.518 $\pm$ 7.118 & \textbf{71.607 $\pm$ 3.577} \\
\midrule
CMTL & \textbf{LwP} & 73.484 $\pm$ 8.019 & 88.242 $\pm$ 12.010 & 68.545 $\pm$ 4.454 \\
\bottomrule
\end{tabular}
\end{table}

All MTL approaches utilize the same model architecture as LwP. Despite being supplied with all labels for every input data point, the amount of training samples for MTL models matches that seen by CL models per task iteration. Aligning with earlier studies, MTL approaches frequently represent the upper bound for all CL models. An interesting discovery is that all MTL models deliver nearly identical performance on the benchmark.

\subsection{Accuracy Progression for Each Task Iteration} \label{appendix:accuracy_progression}

\begin{figure*}[!t]
\centering
\begin{subfigure}[b]{0.13\textwidth}
    \centering
    \includegraphics[width=\textwidth]{Figures/cm/celeba_LWP_confusion_matrix.png}
    \caption{LwP \textbf{Ours}}
    \label{fig:lwp}
\end{subfigure}
\begin{subfigure}[b]{0.13\textwidth}
    \centering
    \includegraphics[width=\textwidth]{Figures/cm/celeba_LWF_confusion_matrix.png}
    \caption{LWF}
    \label{fig:lwf}
\end{subfigure}
\begin{subfigure}[b]{0.13\textwidth}
    \centering
    \includegraphics[width=\textwidth]{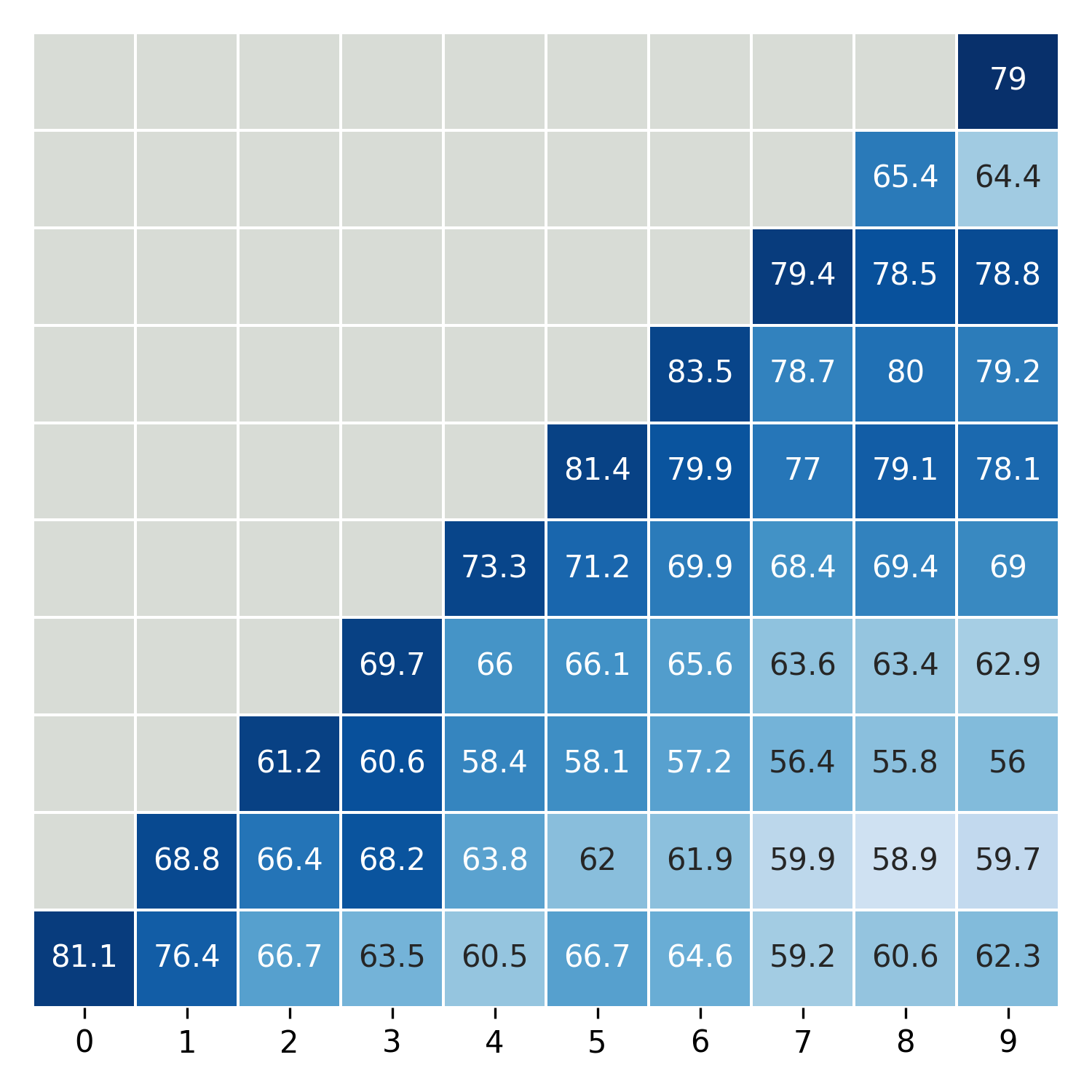}
    \caption{ER}
    \label{fig:er}
\end{subfigure}
\begin{subfigure}[b]{0.13\textwidth}
    \centering
    \includegraphics[width=\textwidth]{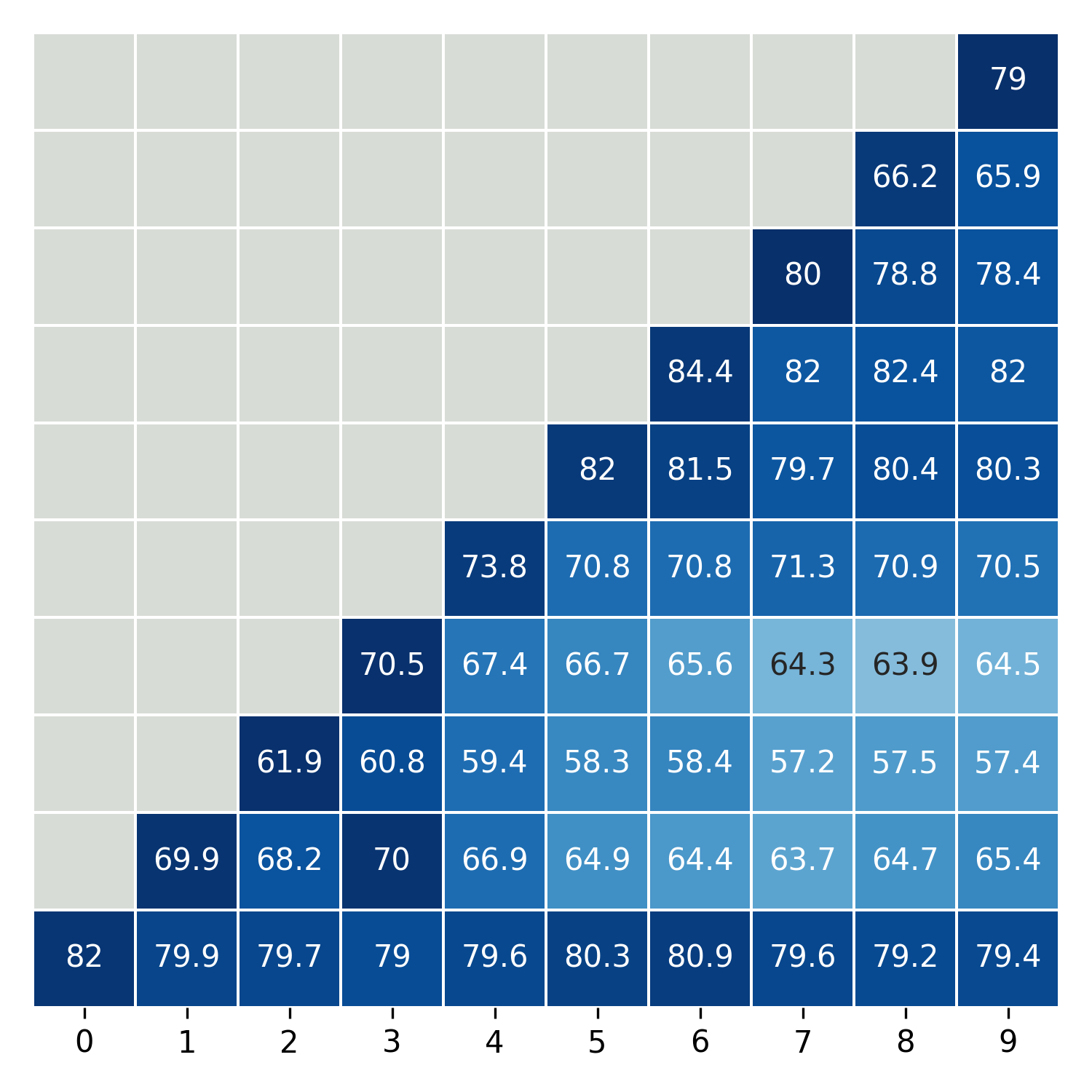}
    \caption{GSS}
    \label{fig:gss}
\end{subfigure}
\begin{subfigure}[b]{0.13\textwidth}
    \centering
    \includegraphics[width=\textwidth]{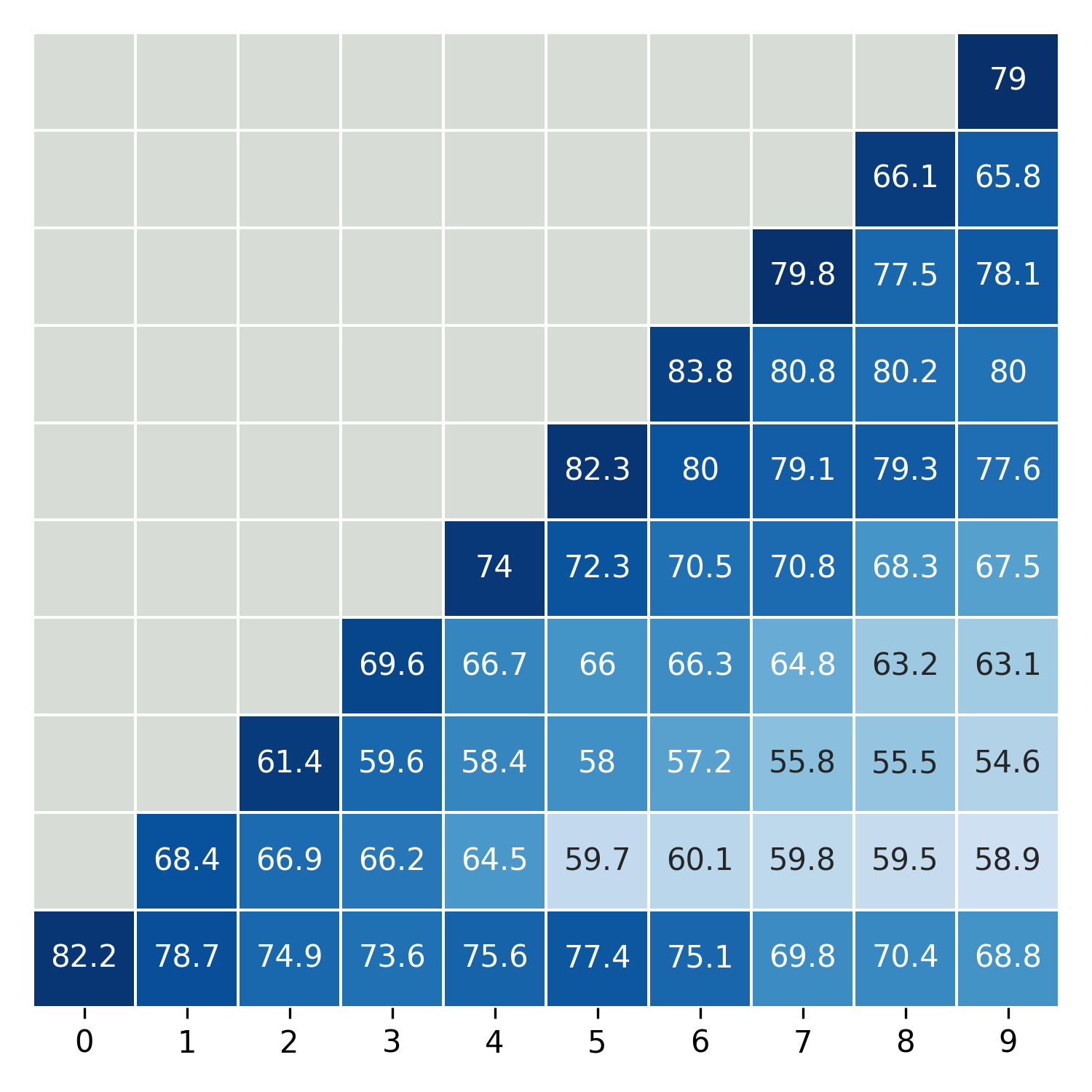}
    \caption{FDR}
    \label{fig:fdr}
\end{subfigure}

\begin{subfigure}[b]{0.13\textwidth}
    \centering
    \includegraphics[width=\textwidth]{Figures/cm/celeba_OBC_confusion_matrix.png}
    \caption{OBC}
    \label{fig:obc}
\end{subfigure}
\begin{subfigure}[b]{0.13\textwidth}
    \centering
    \includegraphics[width=\textwidth]{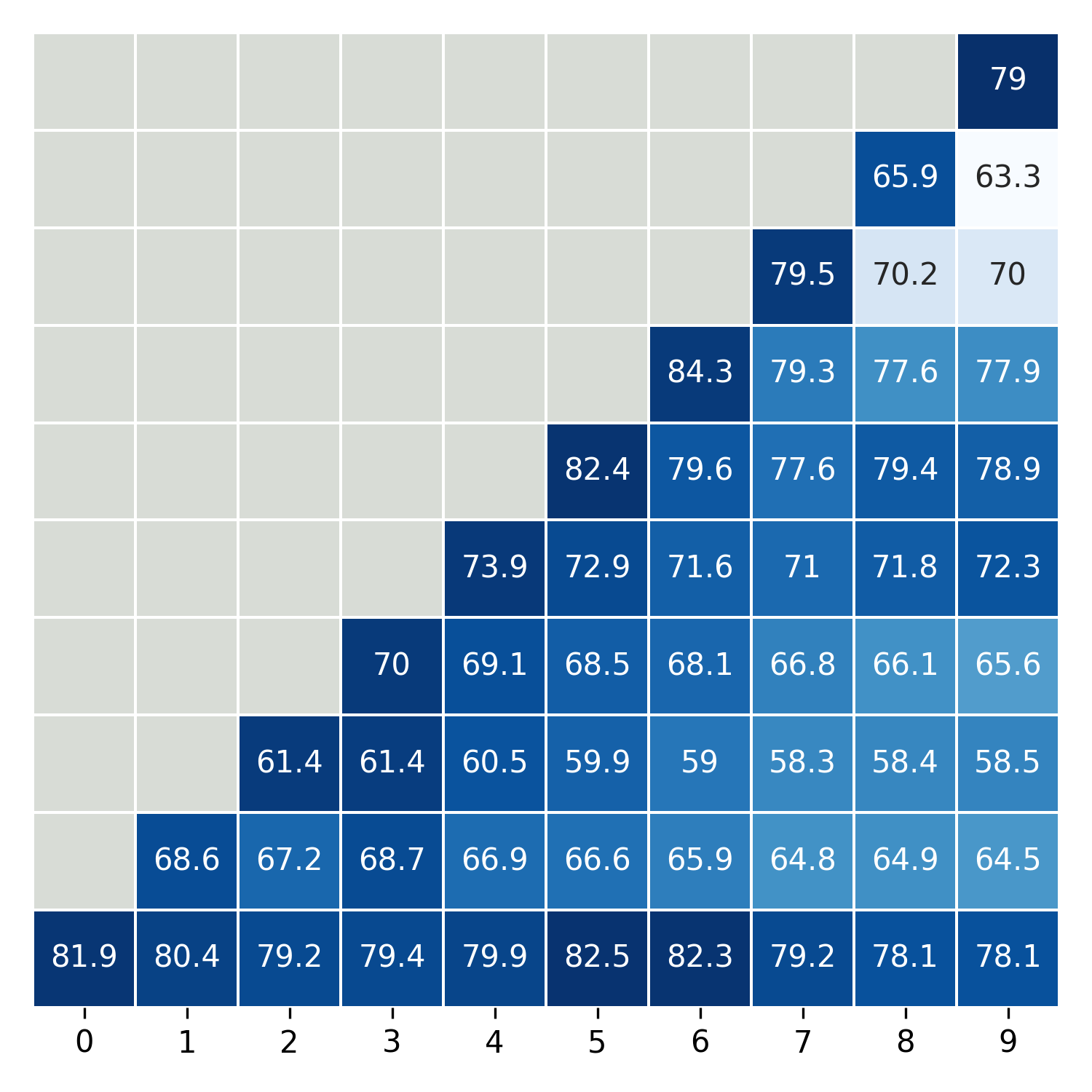}
    \caption{DVC}
    \label{fig:dvc}
\end{subfigure}
\begin{subfigure}[b]{0.13\textwidth}
    \centering
    \includegraphics[width=\textwidth]{Figures/cm/celeba_DER_confusion_matrix.png}
    \caption{DER}
    \label{fig:der}
\end{subfigure}
\begin{subfigure}[b]{0.13\textwidth}
    \centering
    \includegraphics[width=\textwidth]{Figures/cm/celeba_DERPP_confusion_matrix.png}
    \caption{DERPP}
    \label{fig:derpp}
\end{subfigure}
\begin{subfigure}[b]{0.13\textwidth}
    \centering
    \includegraphics[width=\textwidth]{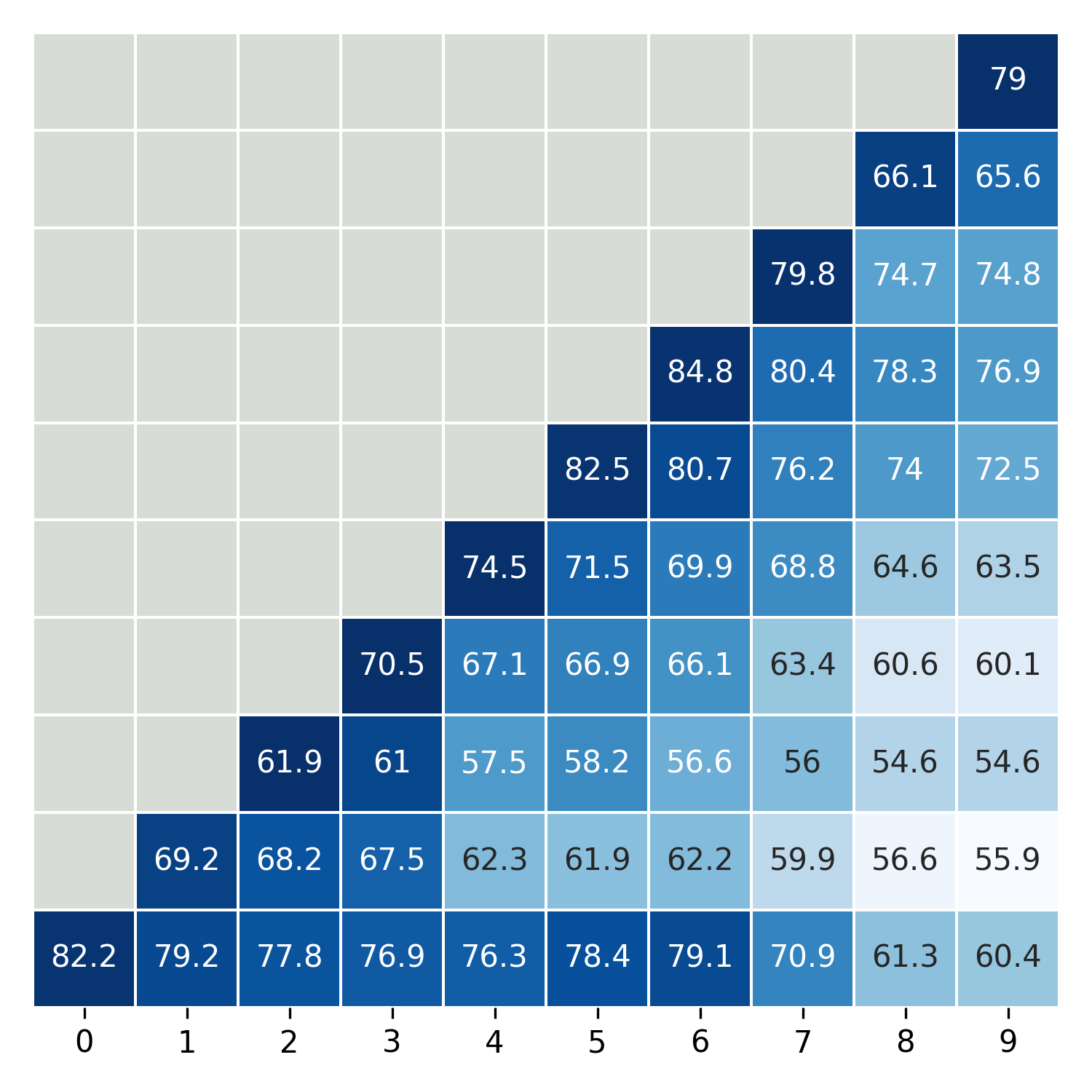}
    \caption{SI}
    \label{fig:si}
\end{subfigure}
\begin{subfigure}[b]{0.13\textwidth}
    \centering
    \includegraphics[width=\textwidth]{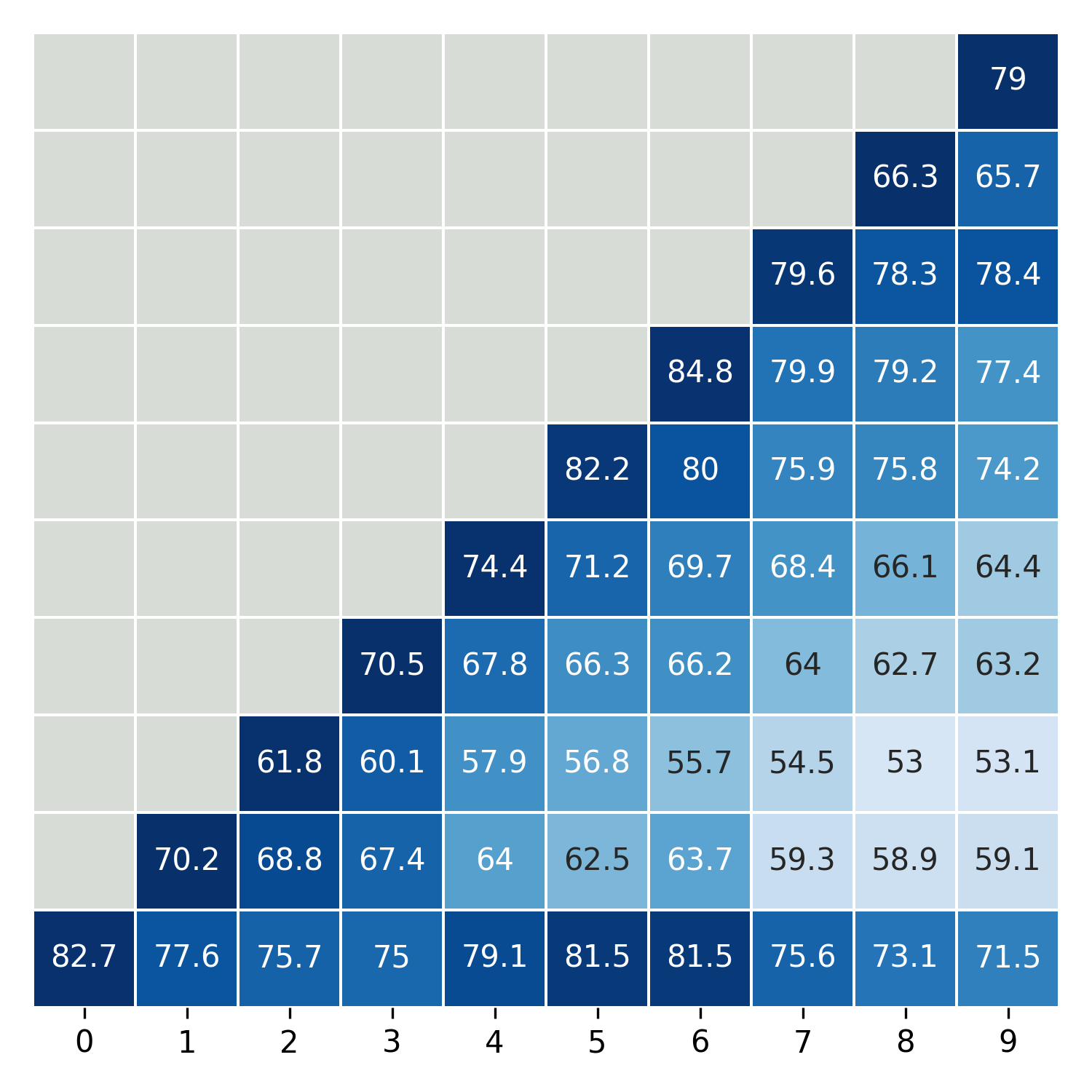}
    \caption{oEWC}
    \label{fig:oewc}
\end{subfigure}

\caption{Matrices showcasing the accuracy progression for various models for the dataset CelebA. Each column corresponds to an iteration of the task, arranged sequentially from left to right. We generate confusion matrices normalized across all tasks for all models to ensure consistency.}

\label{fig:celeba_confusion_matrices}
\end{figure*}

\begin{figure*}[htbp]
\centering
\begin{subfigure}[b]{0.13\textwidth}
    \centering
    \includegraphics[width=\textwidth]{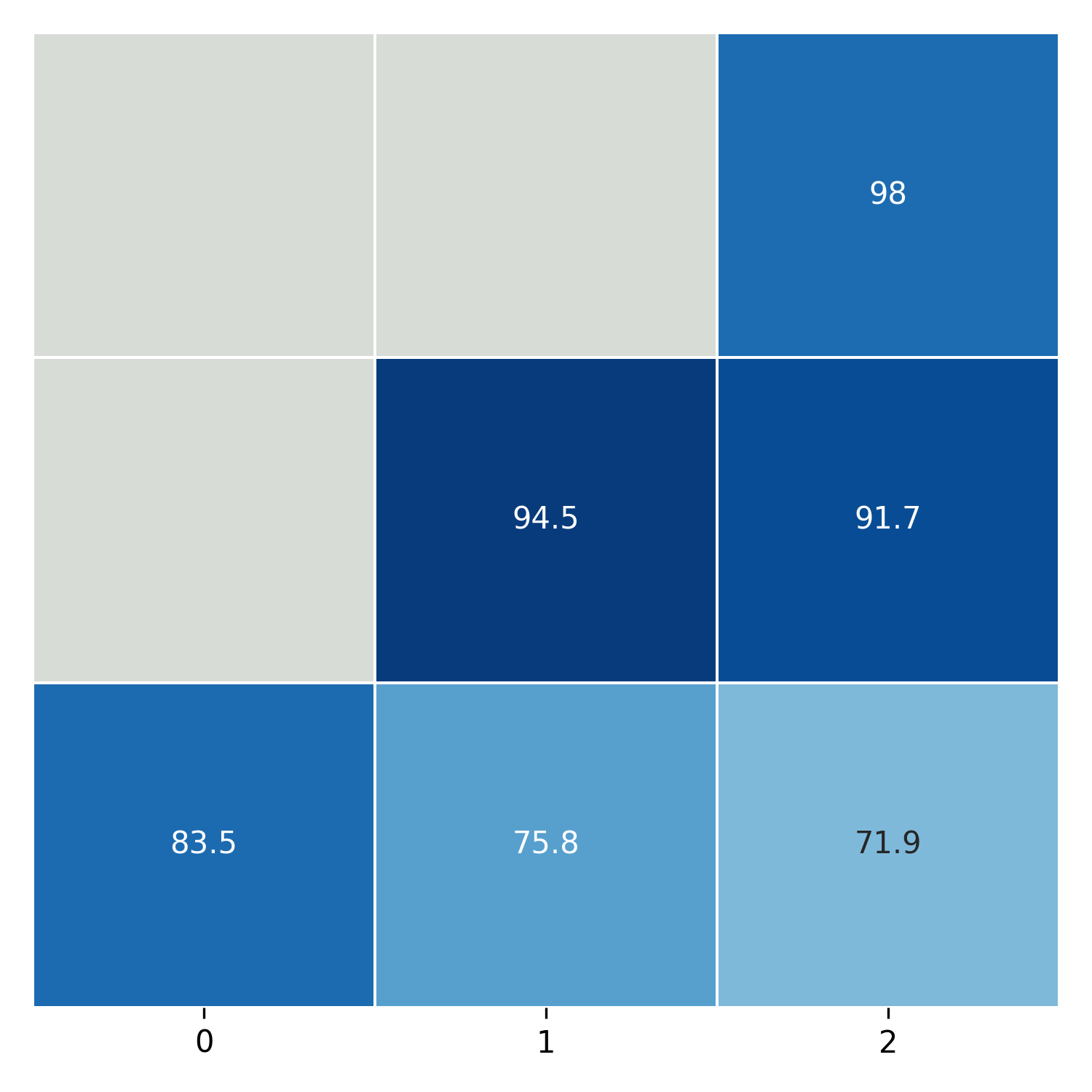}
    \caption{LwP \textbf{Ours}}
    \label{fig:physiq_lwp}
\end{subfigure}
\begin{subfigure}[b]{0.13\textwidth}
    \centering
    \includegraphics[width=\textwidth]{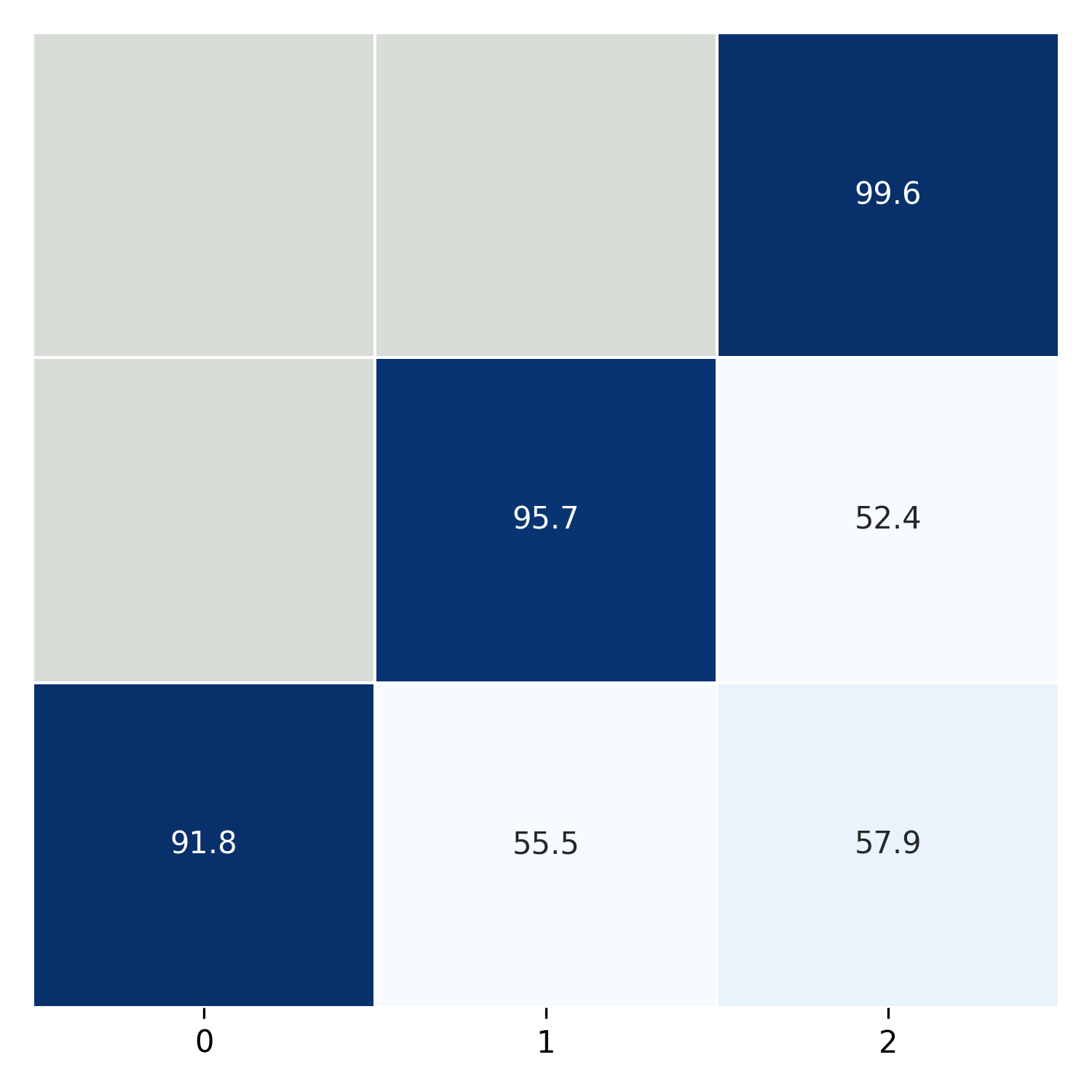}
    \caption{LwF}
    \label{fig:physiq_lwf}
\end{subfigure}
\begin{subfigure}[b]{0.13\textwidth}
    \centering
    \includegraphics[width=\textwidth]{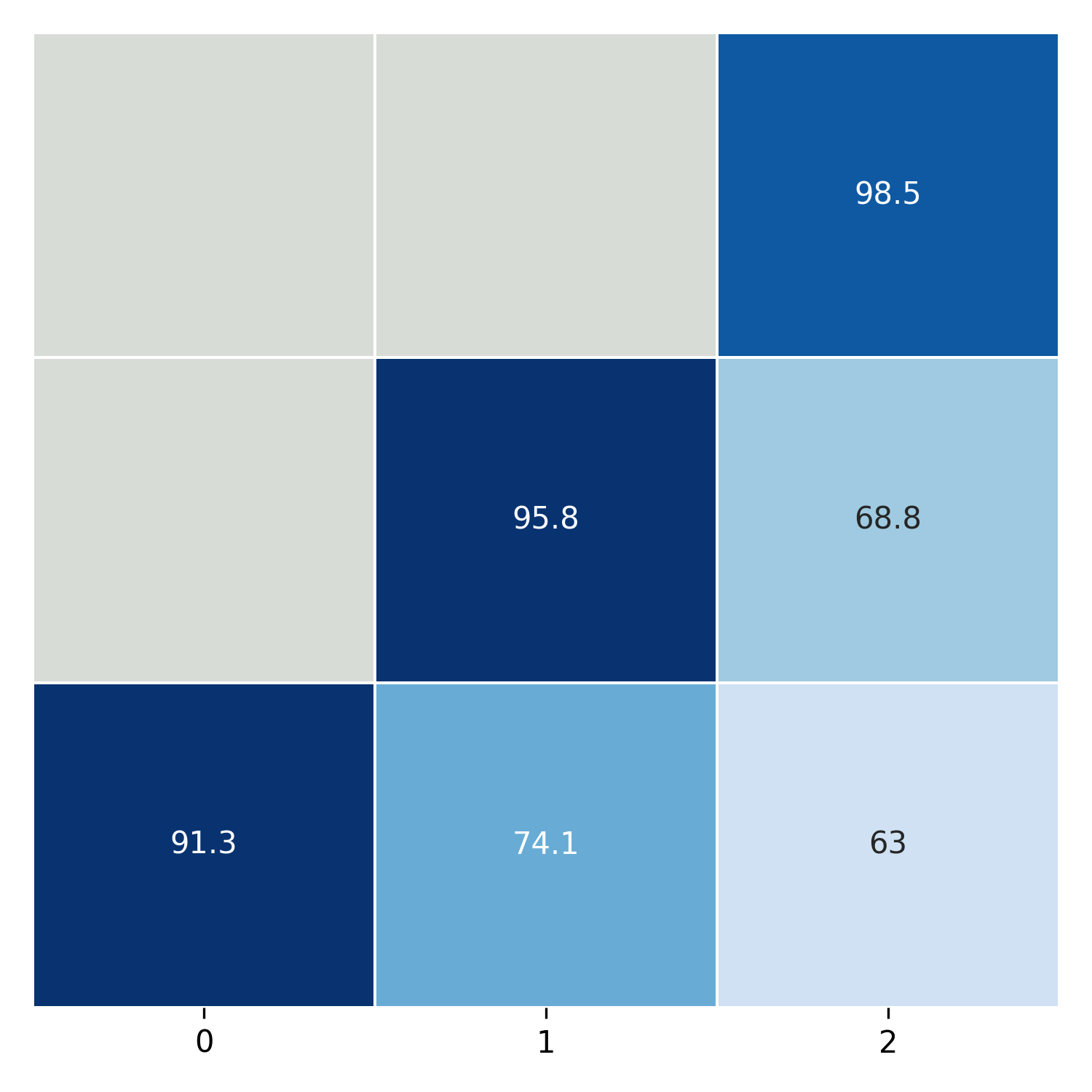}
    \caption{ER}
    \label{fig:physiq_er}
\end{subfigure}
\begin{subfigure}[b]{0.13\textwidth}
    \centering
    \includegraphics[width=\textwidth]{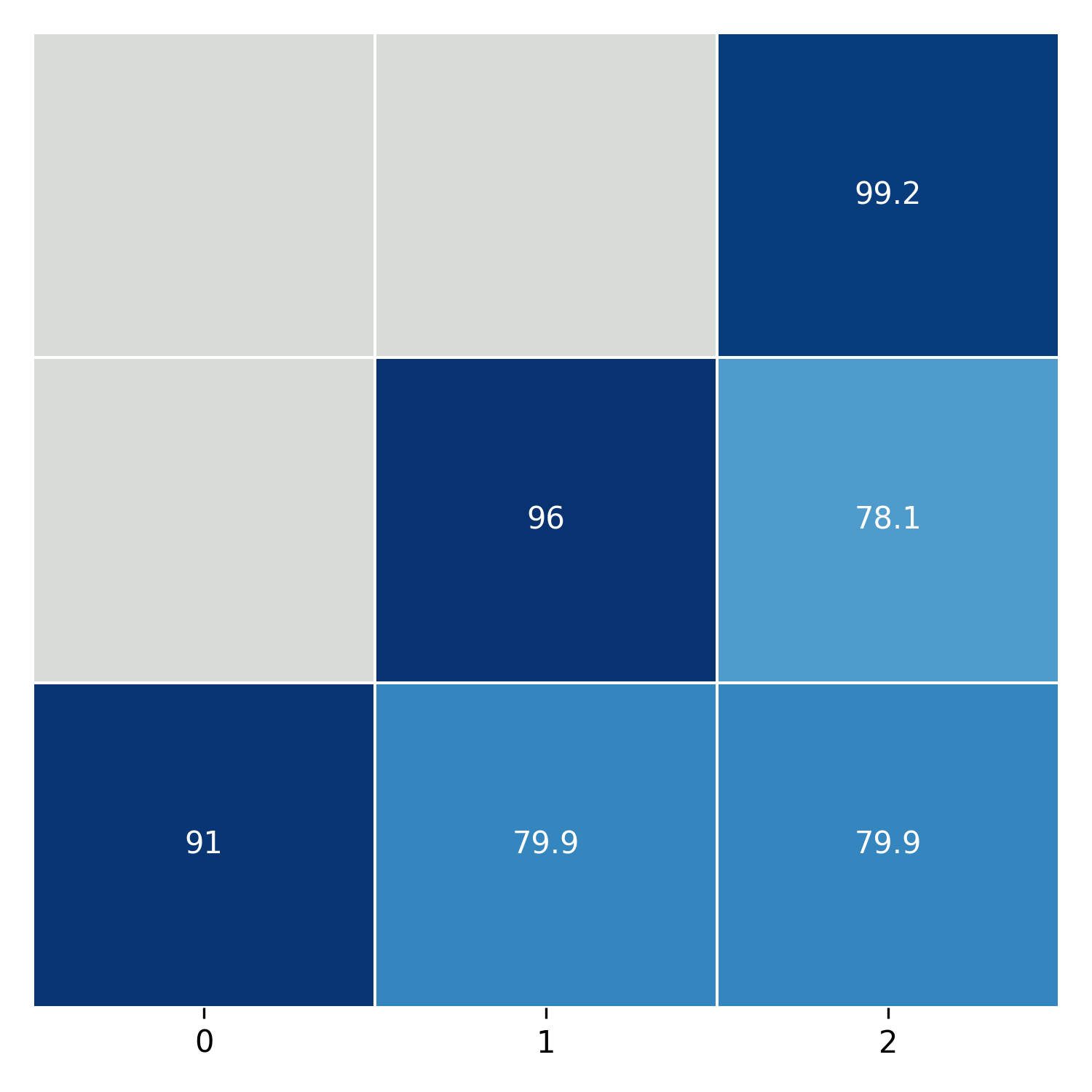}
    \caption{GSS}
    \label{fig:physiq_gss}
\end{subfigure}
\begin{subfigure}[b]{0.13\textwidth}
    \centering
    \includegraphics[width=\textwidth]{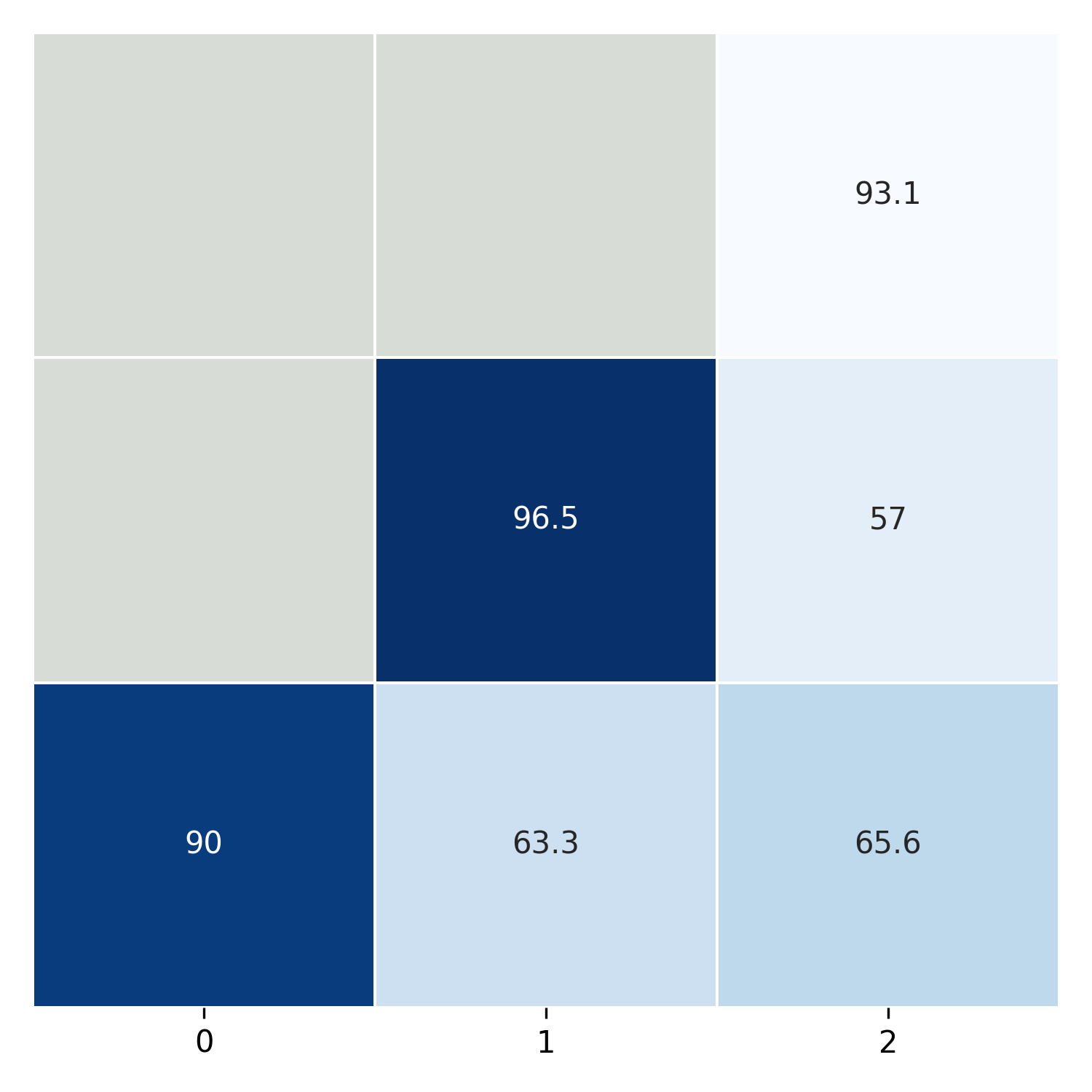}
    \caption{FDR}
    \label{fig:physiq_fdr}
\end{subfigure}

\begin{subfigure}[b]{0.13\textwidth}
    \centering
    \includegraphics[width=\textwidth]{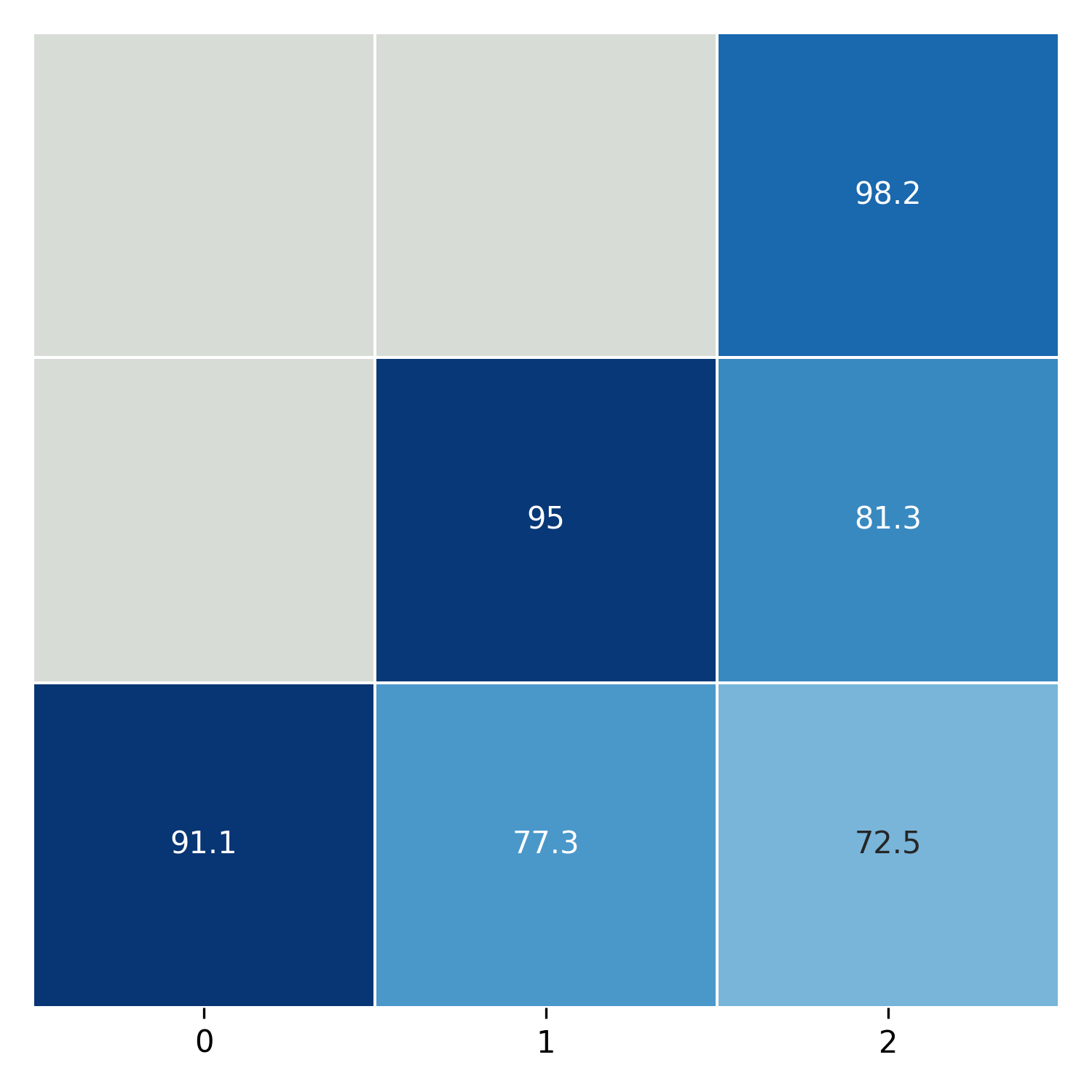}
    \caption{OBC}
    \label{fig:physiq_obc}
\end{subfigure}
\begin{subfigure}[b]{0.13\textwidth}
    \centering
    \includegraphics[width=\textwidth]{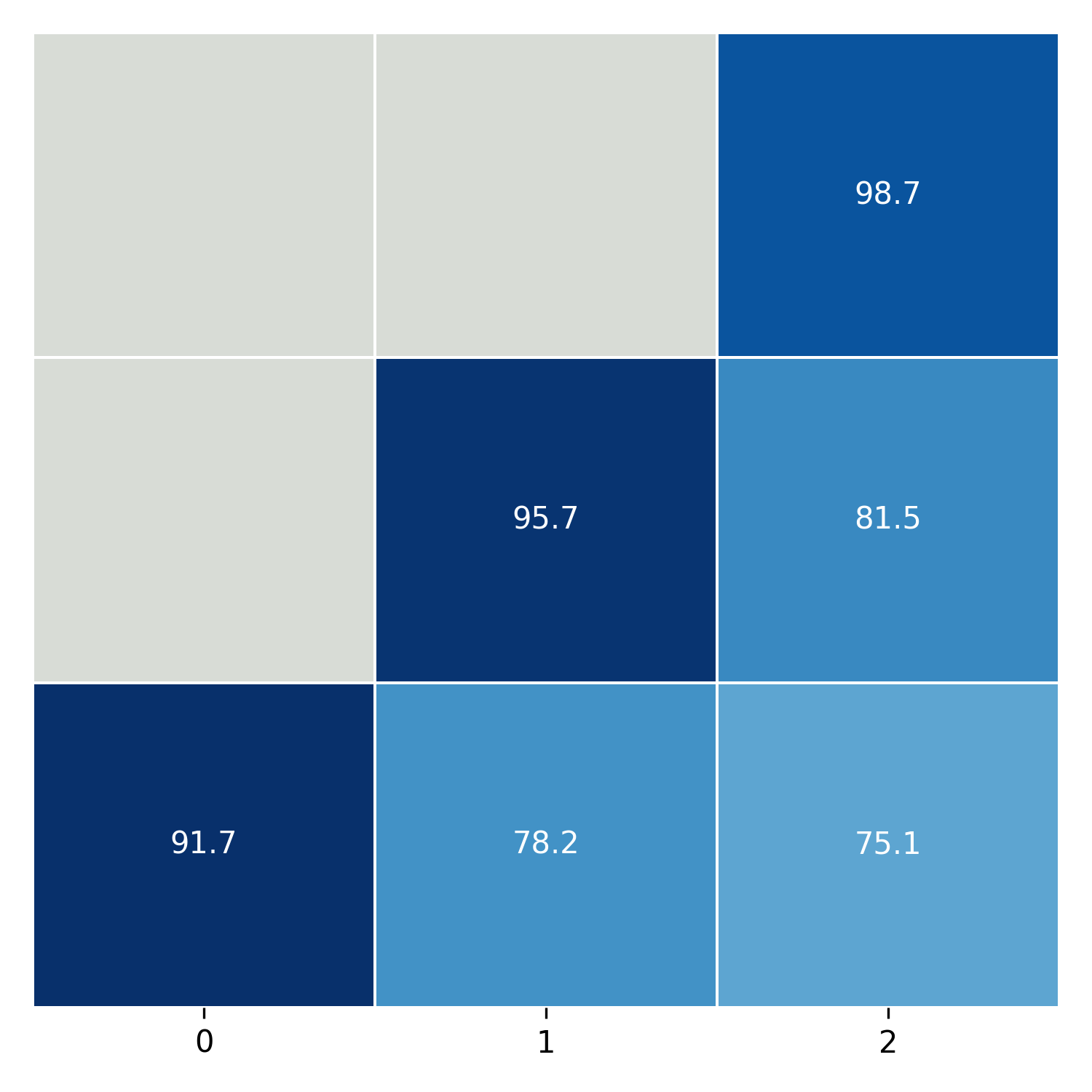}
    \caption{DVC}
    \label{fig:physiq_dvc}
\end{subfigure}
\begin{subfigure}[b]{0.13\textwidth}
    \centering
    \includegraphics[width=\textwidth]{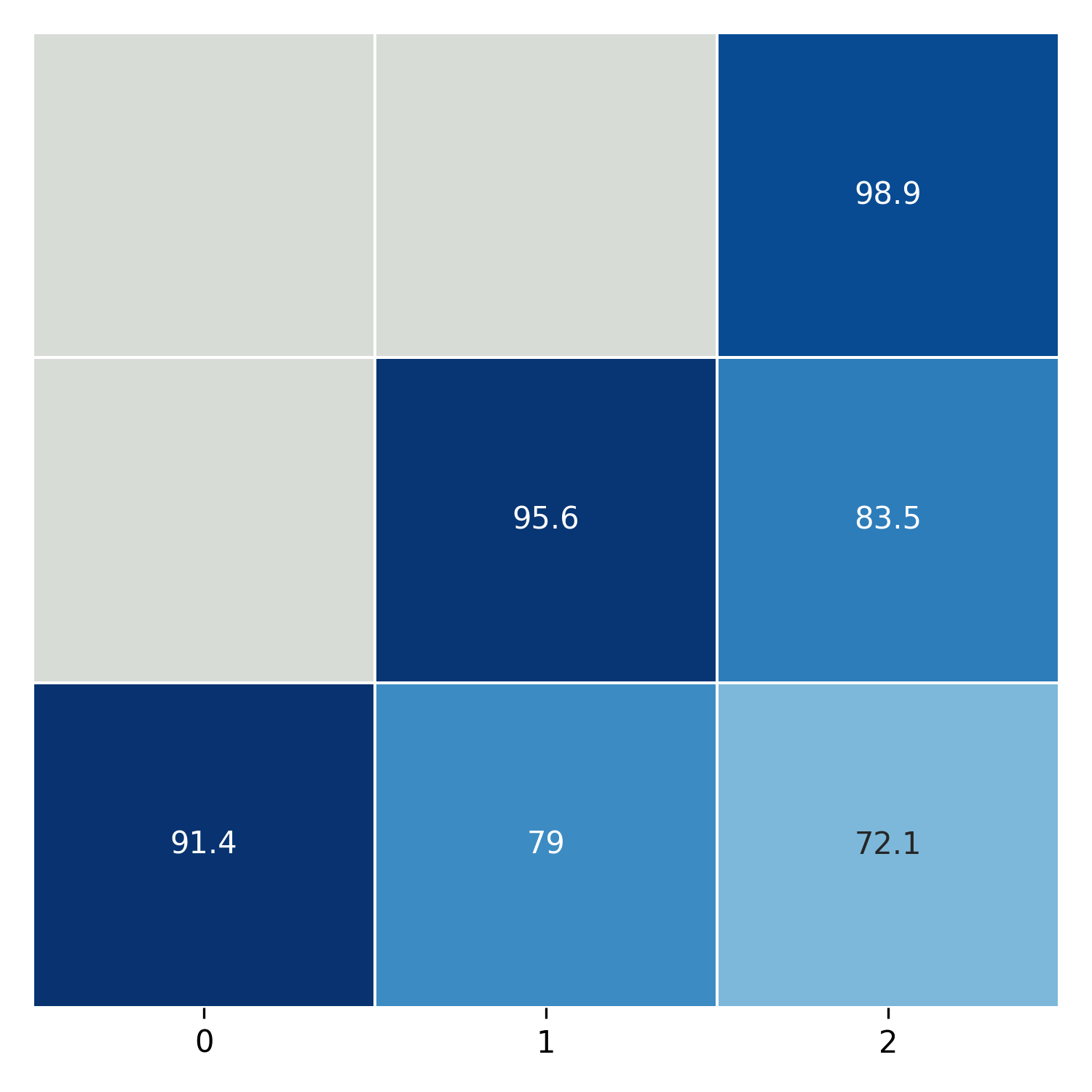}
    \caption{DER}
    \label{fig:physiq_der}
\end{subfigure}
\begin{subfigure}[b]{0.13\textwidth}
    \centering
    \includegraphics[width=\textwidth]{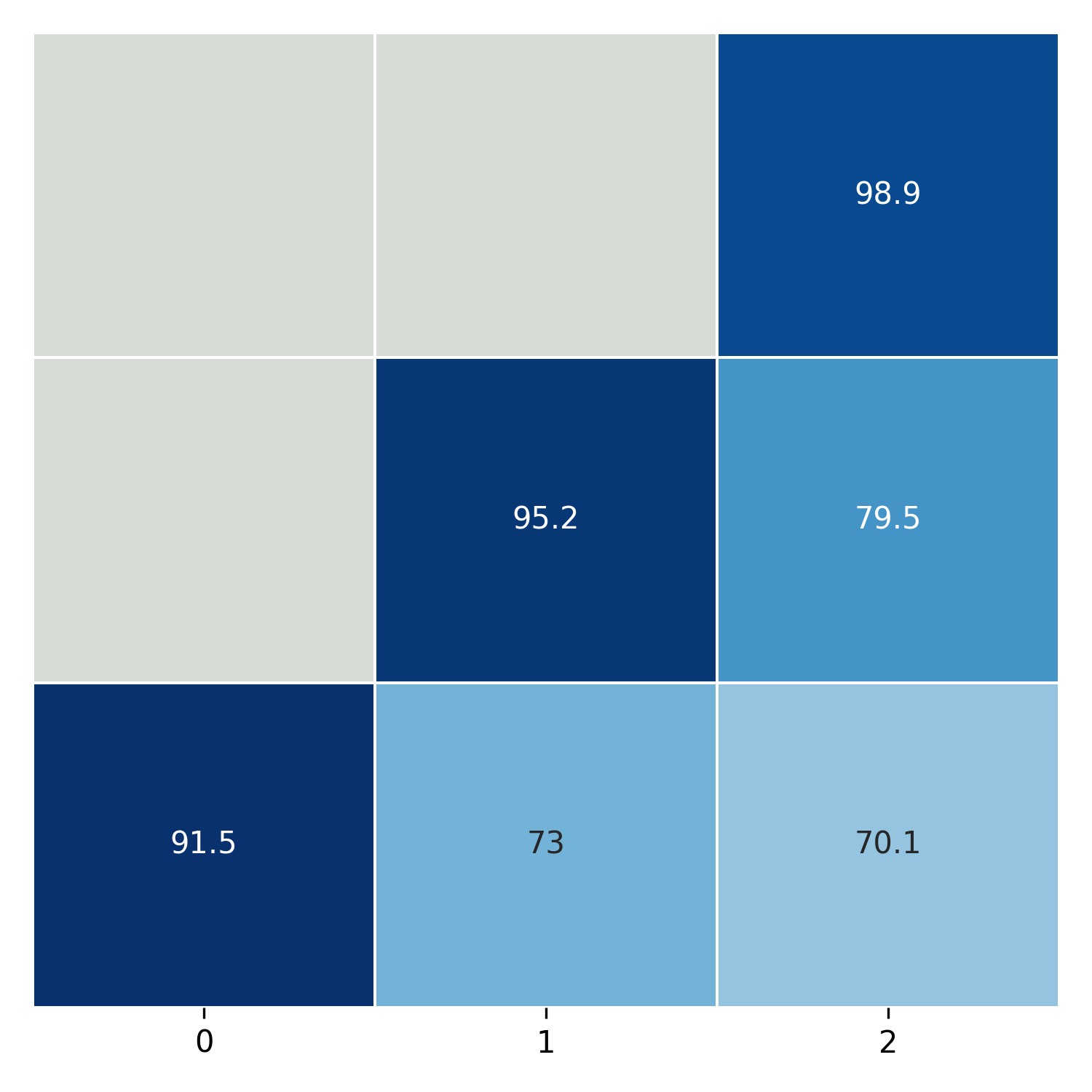}
    \caption{DERPP}
    \label{fig:physiq_derpp}
\end{subfigure}
\begin{subfigure}[b]{0.13\textwidth}
    \centering
    \includegraphics[width=\textwidth]{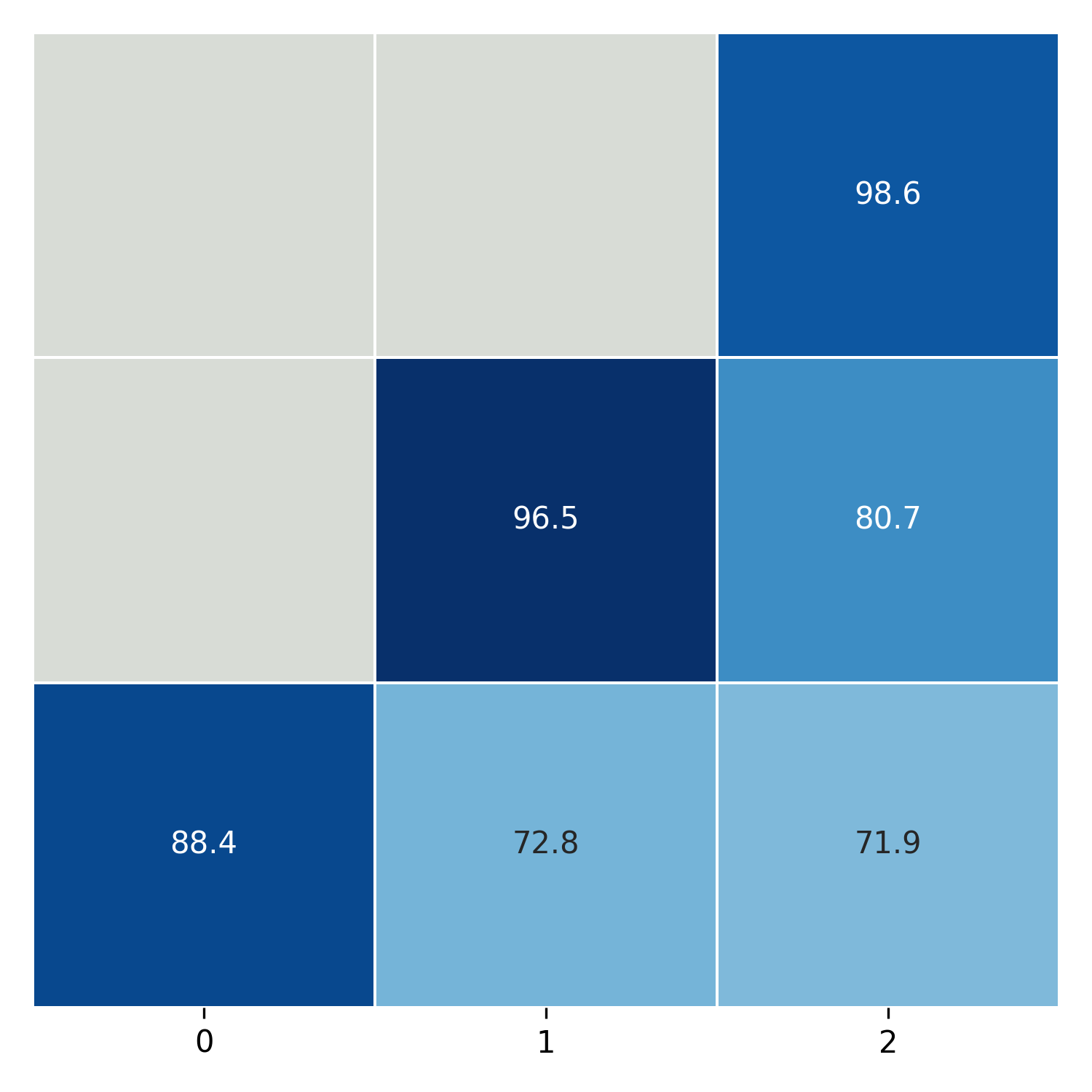}
    \caption{SI}
    \label{fig:physiq_SI}
\end{subfigure}
\begin{subfigure}[b]{0.13\textwidth}
    \centering
    \includegraphics[width=\textwidth]{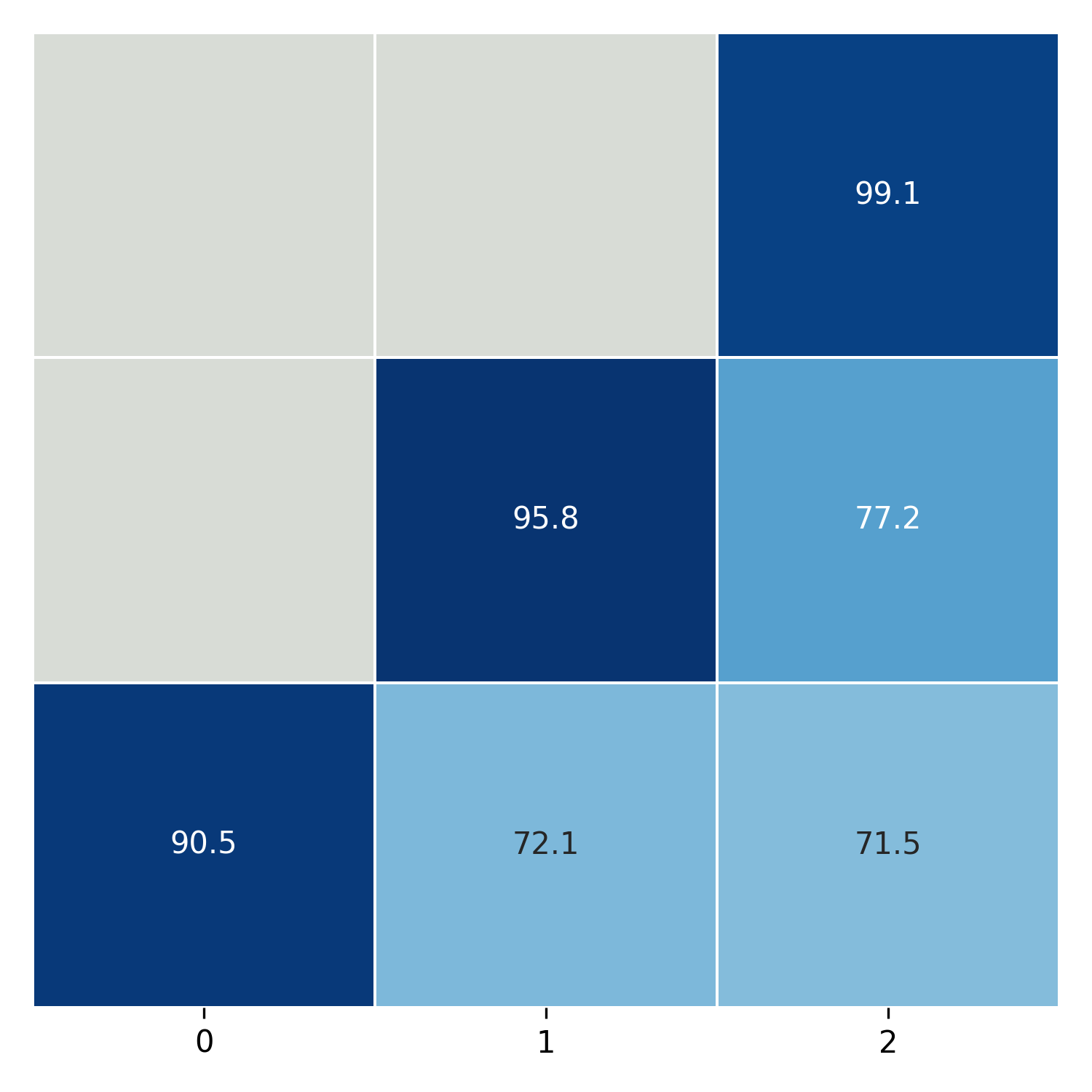}
    \caption{oEWC}
    \label{fig:physiq_EWC}
\end{subfigure}
\caption{Confusion matrices for different models on the PhysiQ dataset}
\label{fig:physiq_confusion_matrices}
\end{figure*}

\begin{figure*}[htbp]
\centering
\begin{subfigure}[b]{0.13\textwidth}
    \centering
    \includegraphics[width=\textwidth]{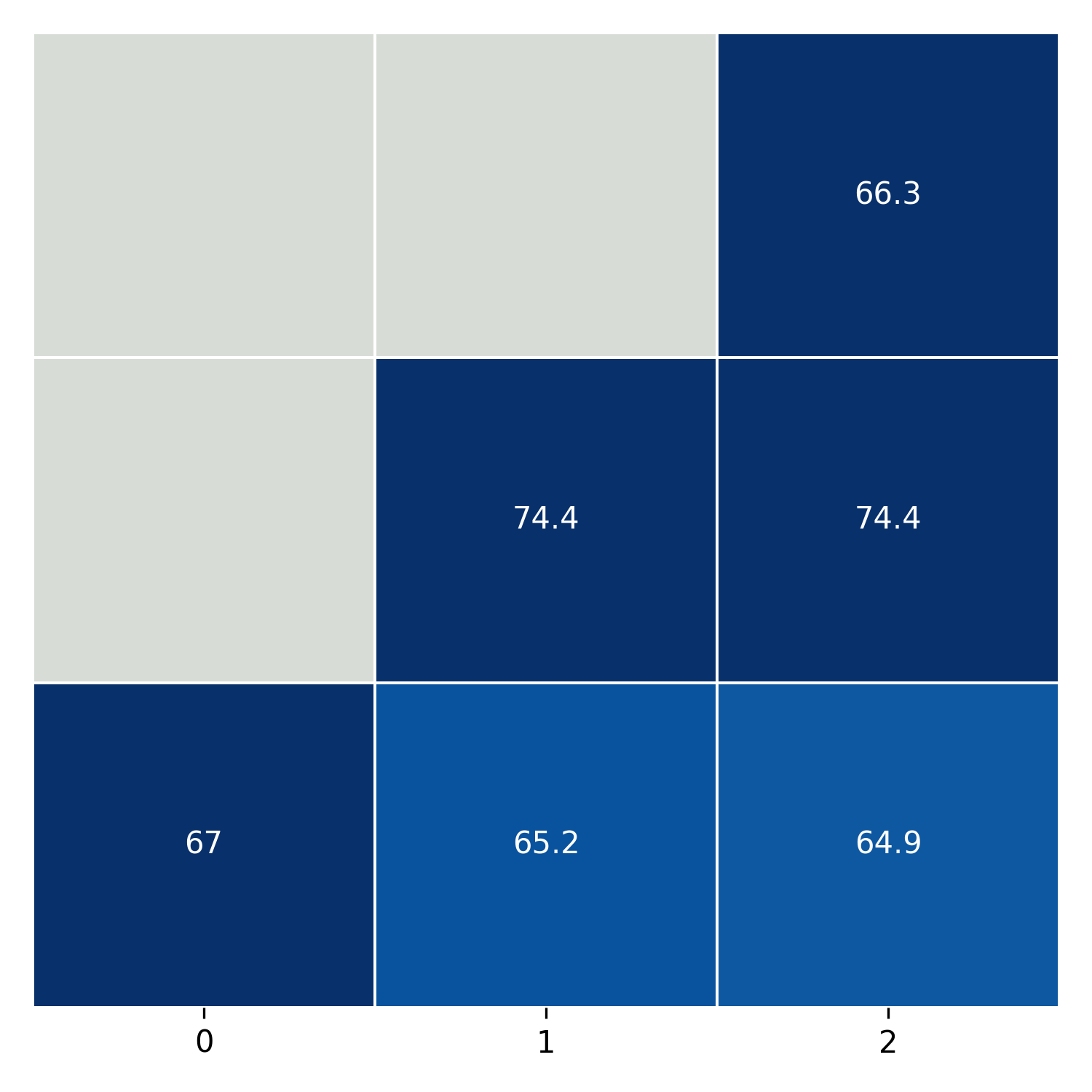}
    \caption{LwP \textbf{Ours}}
    \label{fig:fairface_lwp}
\end{subfigure}
\begin{subfigure}[b]{0.13\textwidth}
    \centering
    \includegraphics[width=\textwidth]{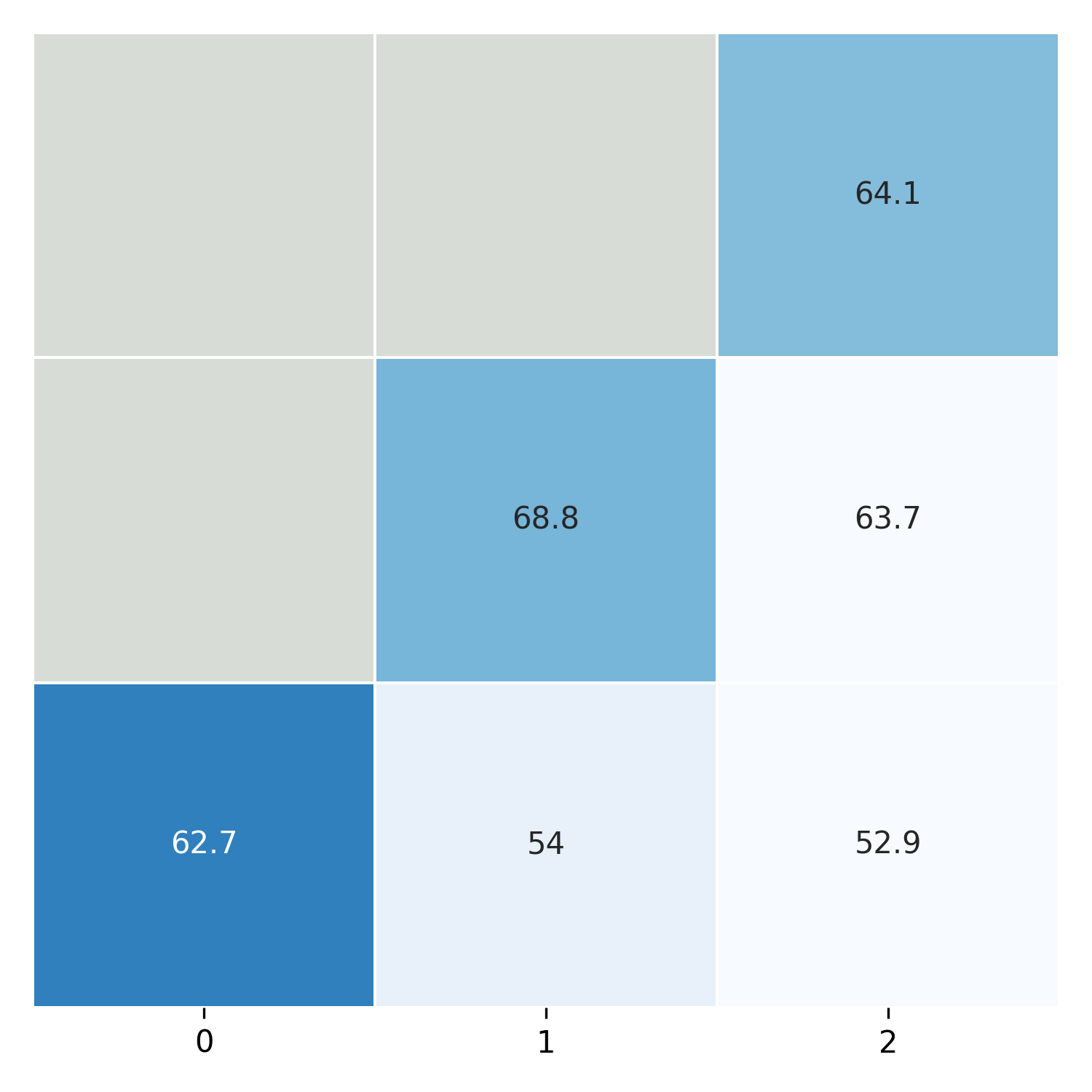}
    \caption{LwF}
    \label{fig:fairface_lwf}
\end{subfigure}
\begin{subfigure}[b]{0.13\textwidth}
    \centering
    \includegraphics[width=\textwidth]{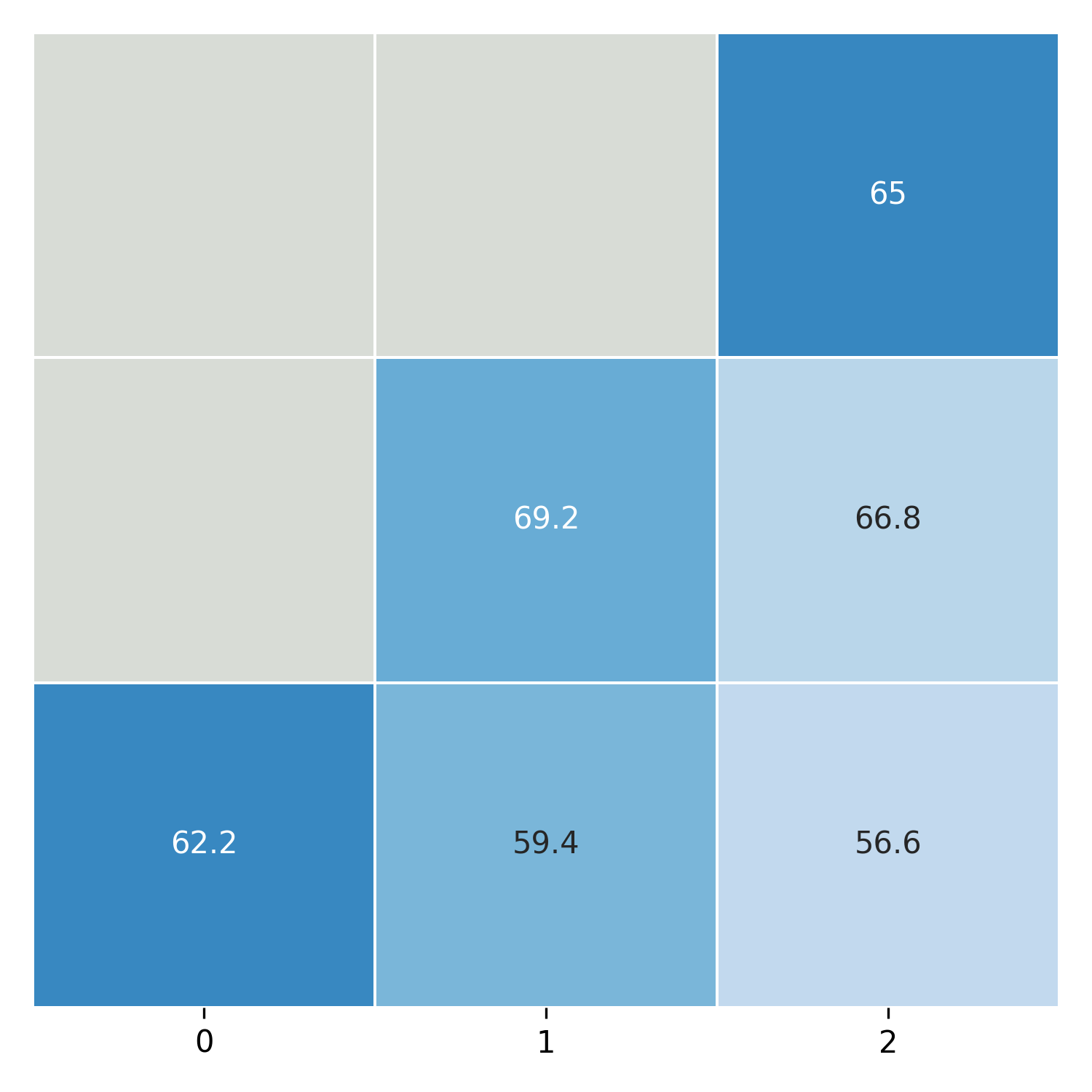}
    \caption{ER}
    \label{fig:fairface_er}
\end{subfigure}
\begin{subfigure}[b]{0.13\textwidth}
    \centering
    \includegraphics[width=\textwidth]{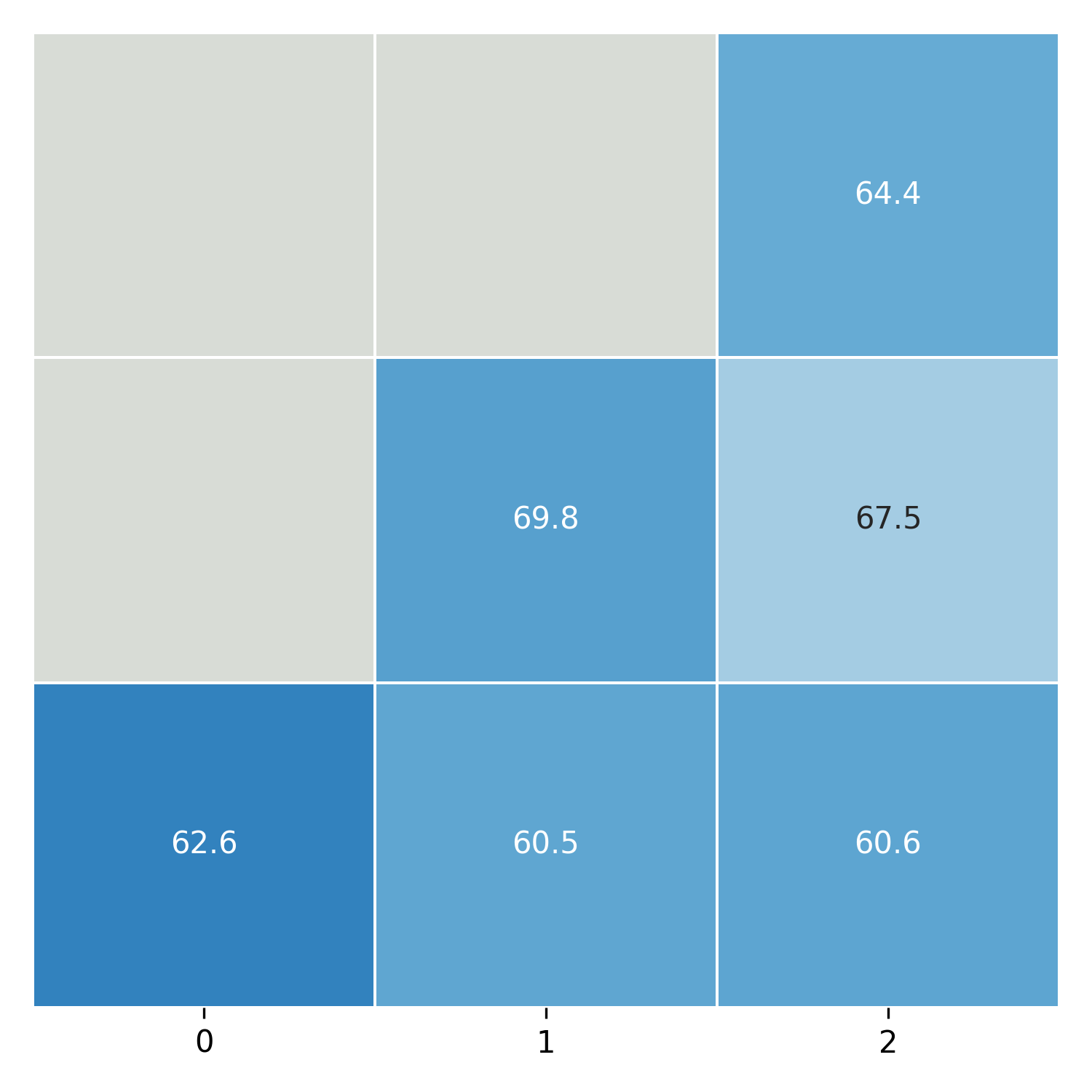}
    \caption{GSS}
    \label{fig:fairface_gss}
\end{subfigure}
\begin{subfigure}[b]{0.13\textwidth}
    \centering
    \includegraphics[width=\textwidth]{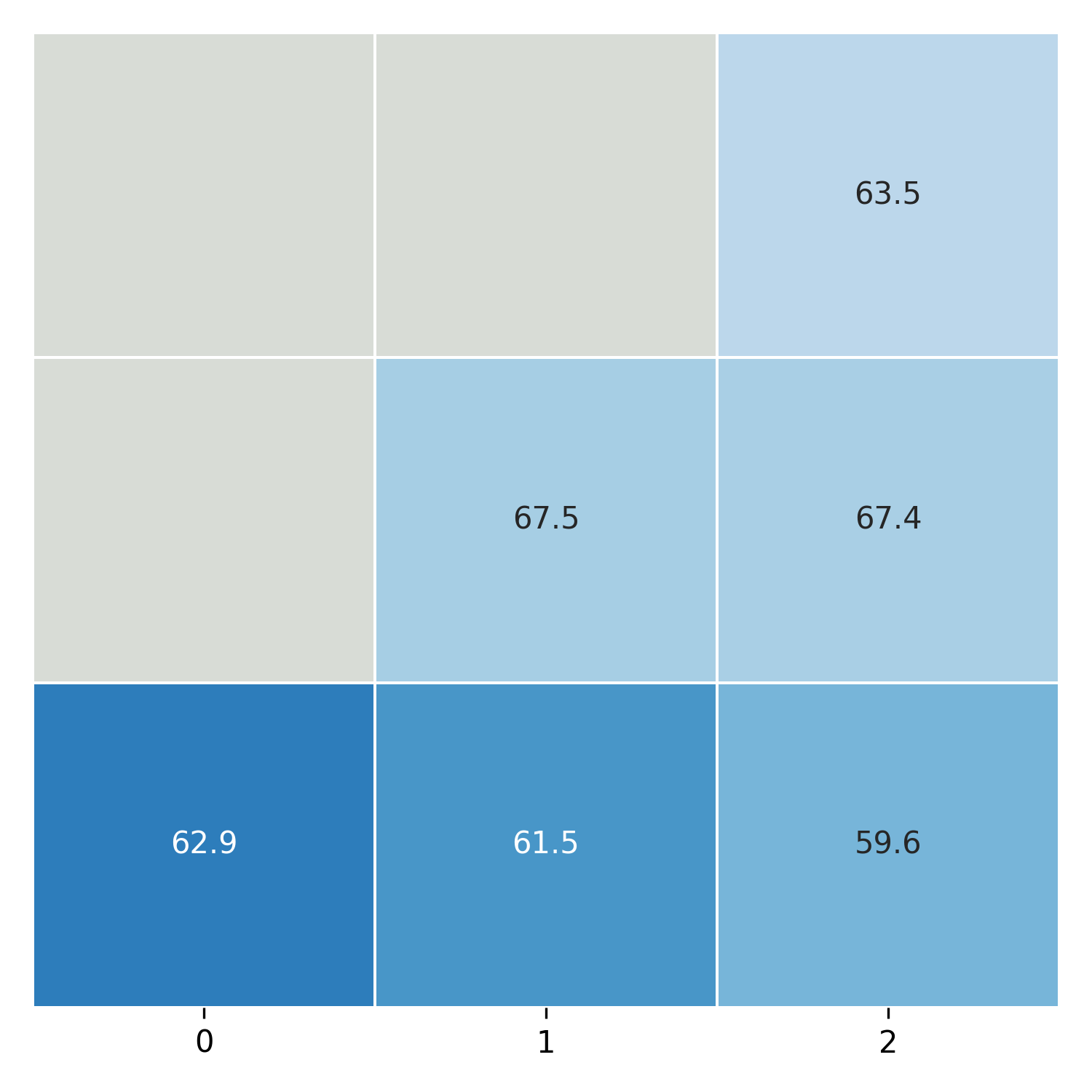}
    \caption{FDR}
    \label{fig:fairface_fdr}
\end{subfigure}

\begin{subfigure}[b]{0.13\textwidth}
    \centering
    \includegraphics[width=\textwidth]{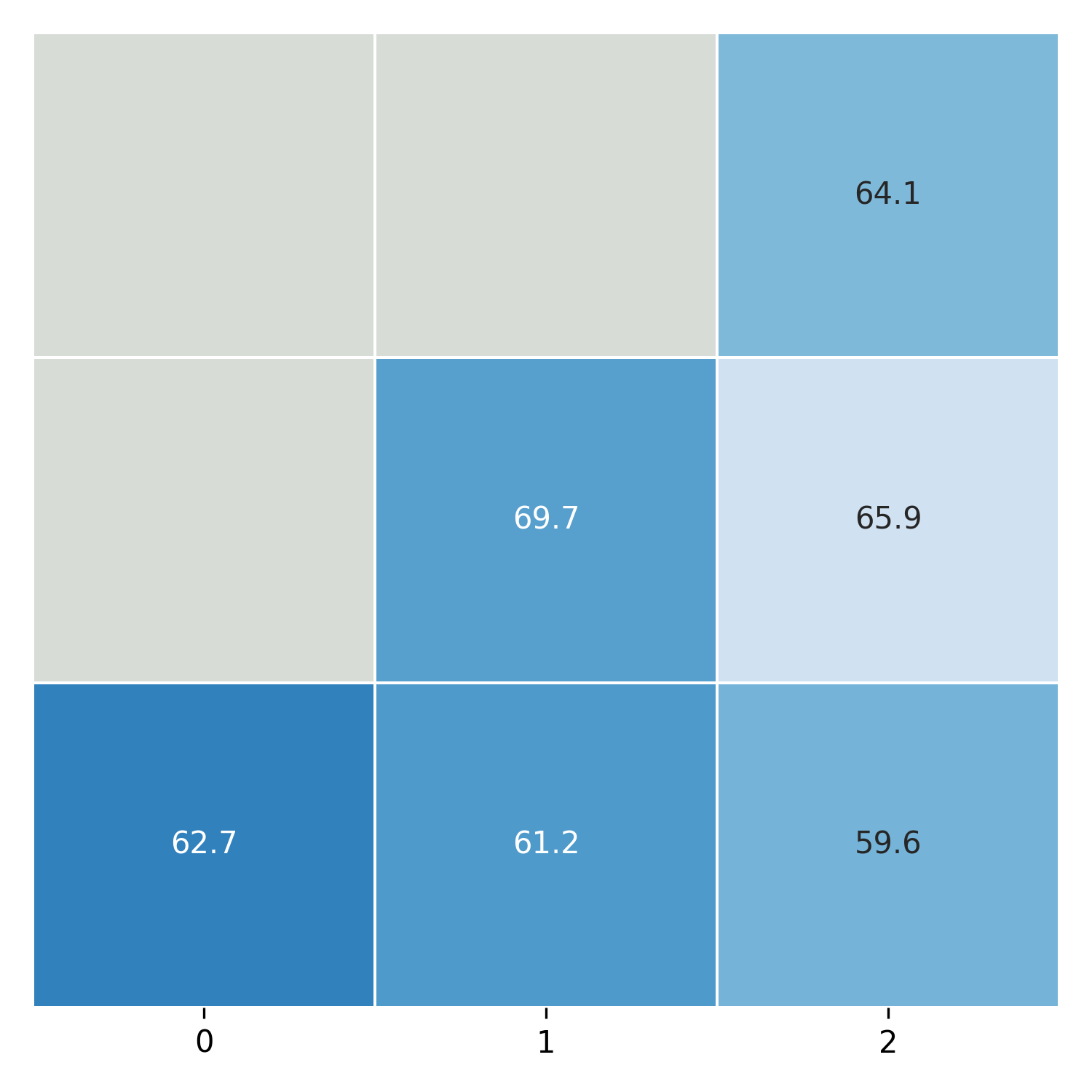}
    \caption{OBC}
    \label{fig:fairface_OBC}
\end{subfigure}
\begin{subfigure}[b]{0.13\textwidth}
    \centering
    \includegraphics[width=\textwidth]{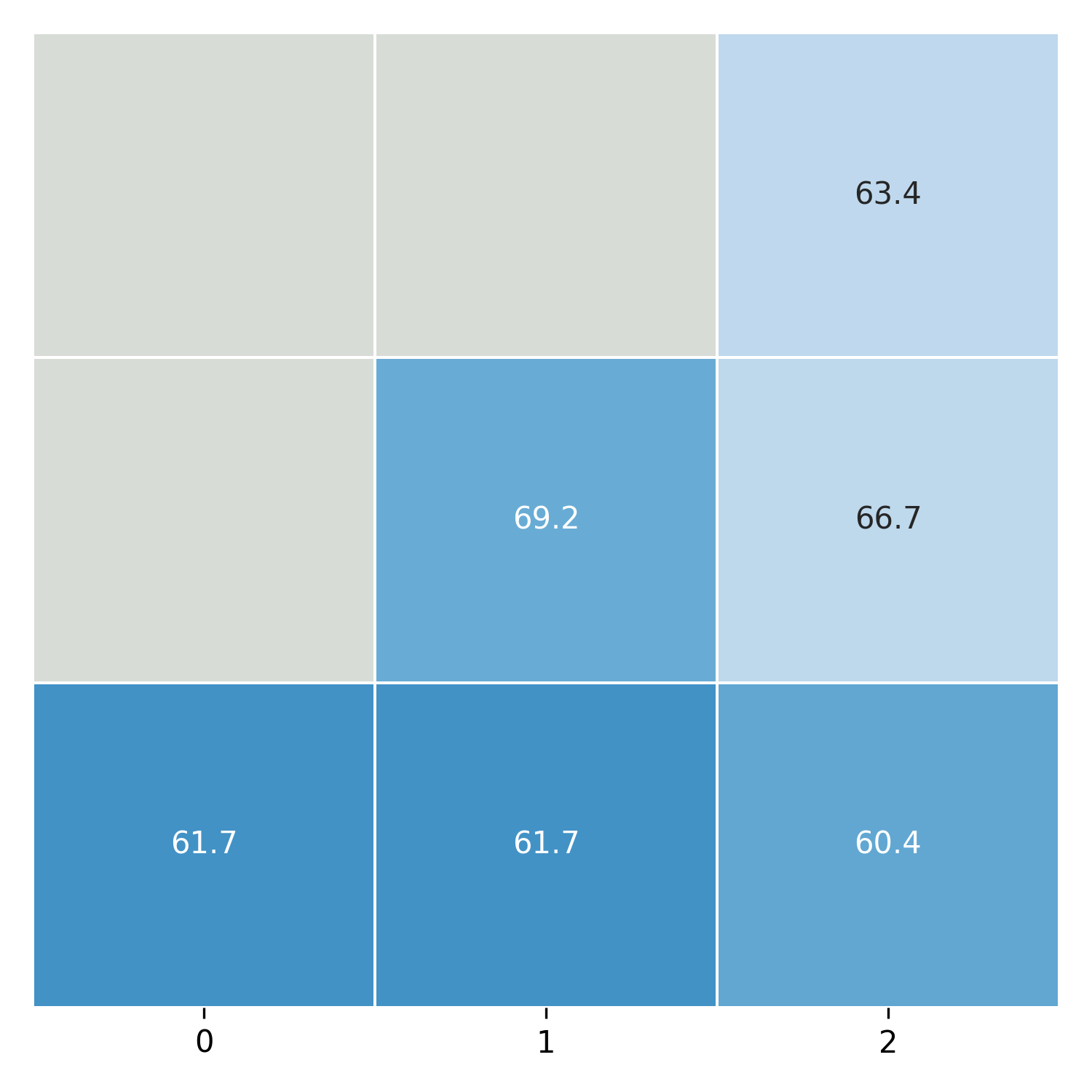}
    \caption{DVC}
    \label{fig:fairface_DVC}
\end{subfigure}
\begin{subfigure}[b]{0.13\textwidth}
    \centering
    \includegraphics[width=\textwidth]{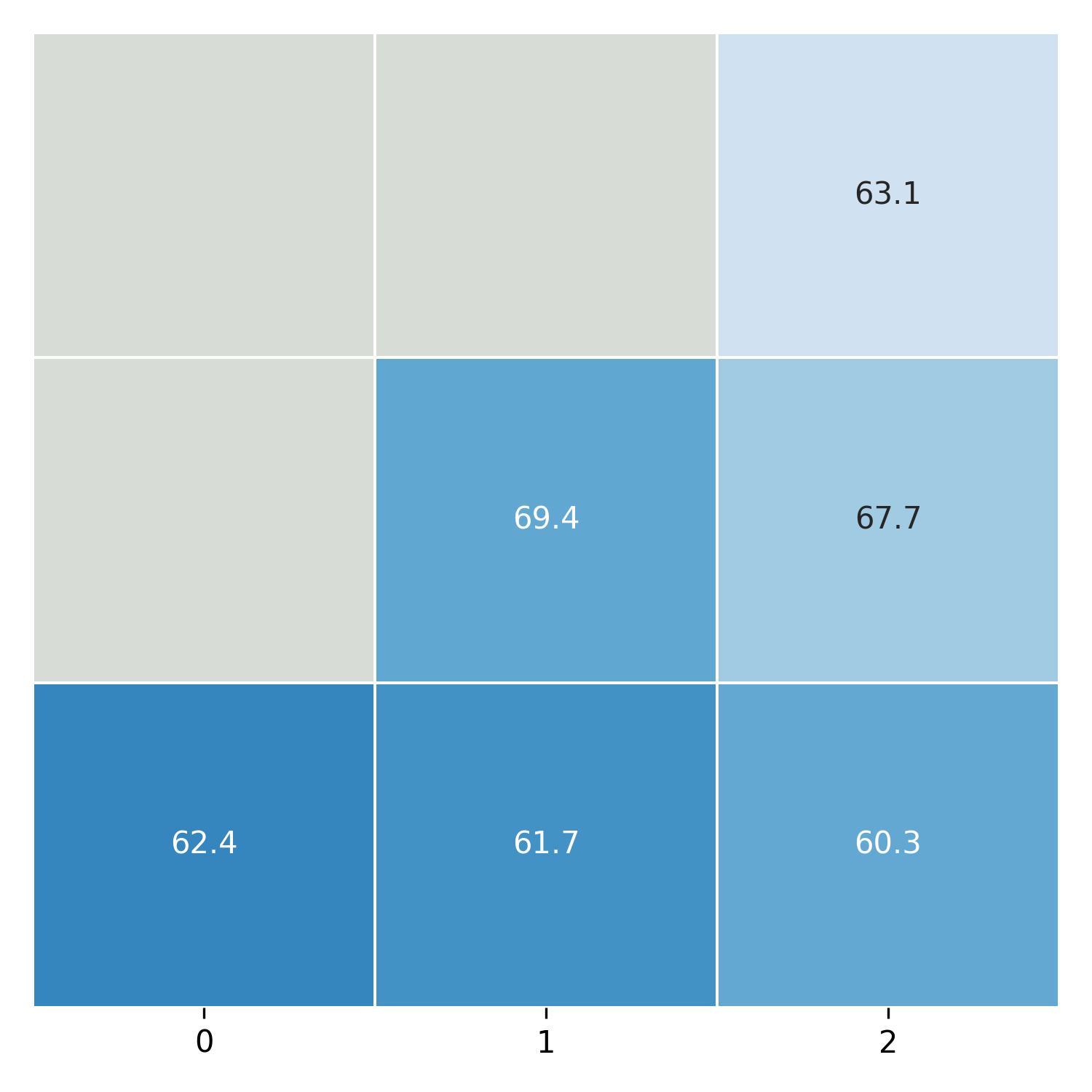}
    \caption{DER}
    \label{fig:fairface_der}
\end{subfigure}
\begin{subfigure}[b]{0.13\textwidth}
    \centering
    \includegraphics[width=\textwidth]{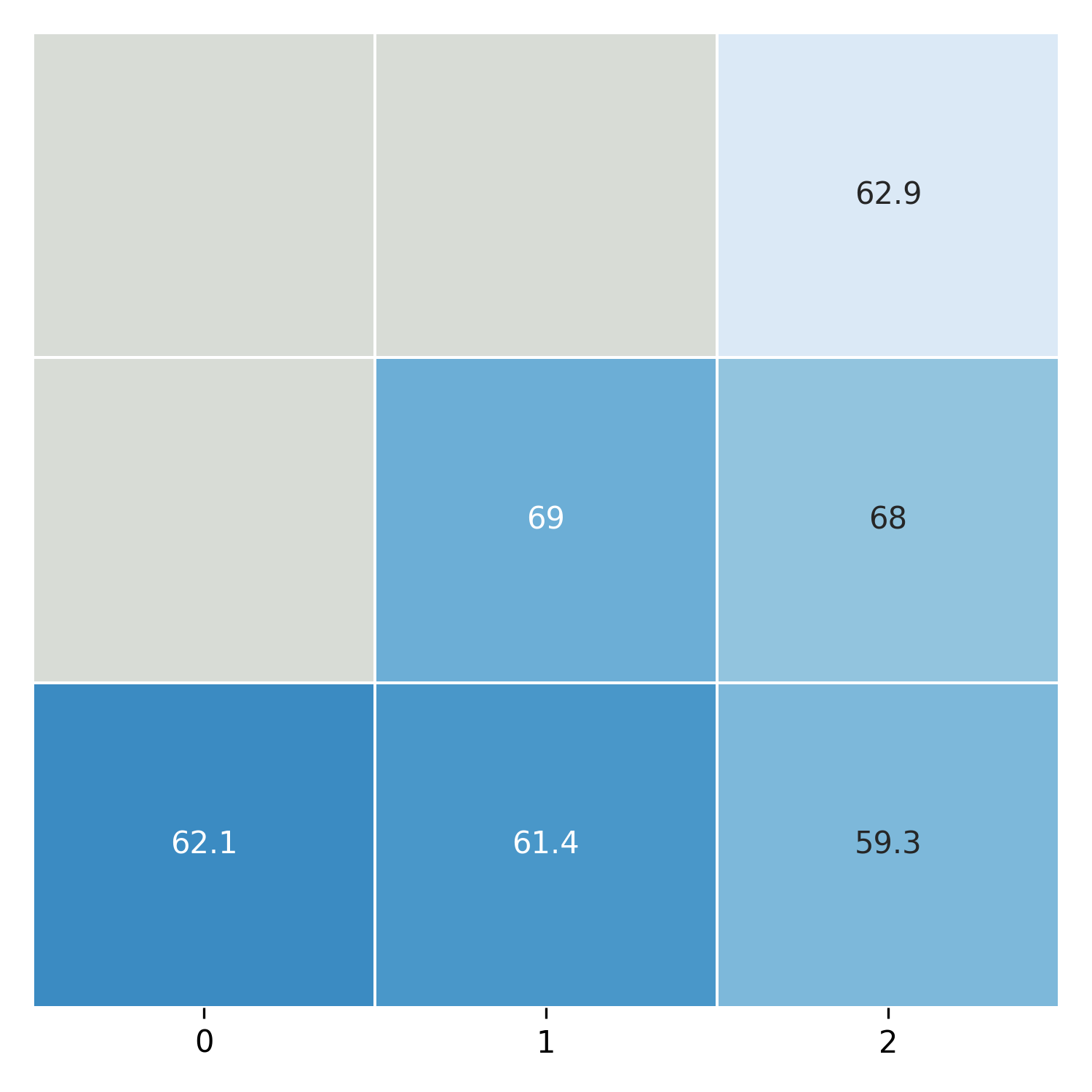}
    \caption{DERPP}
    \label{fig:fairface_derpp}
\end{subfigure}
\begin{subfigure}[b]{0.13\textwidth}
    \centering
    \includegraphics[width=\textwidth]{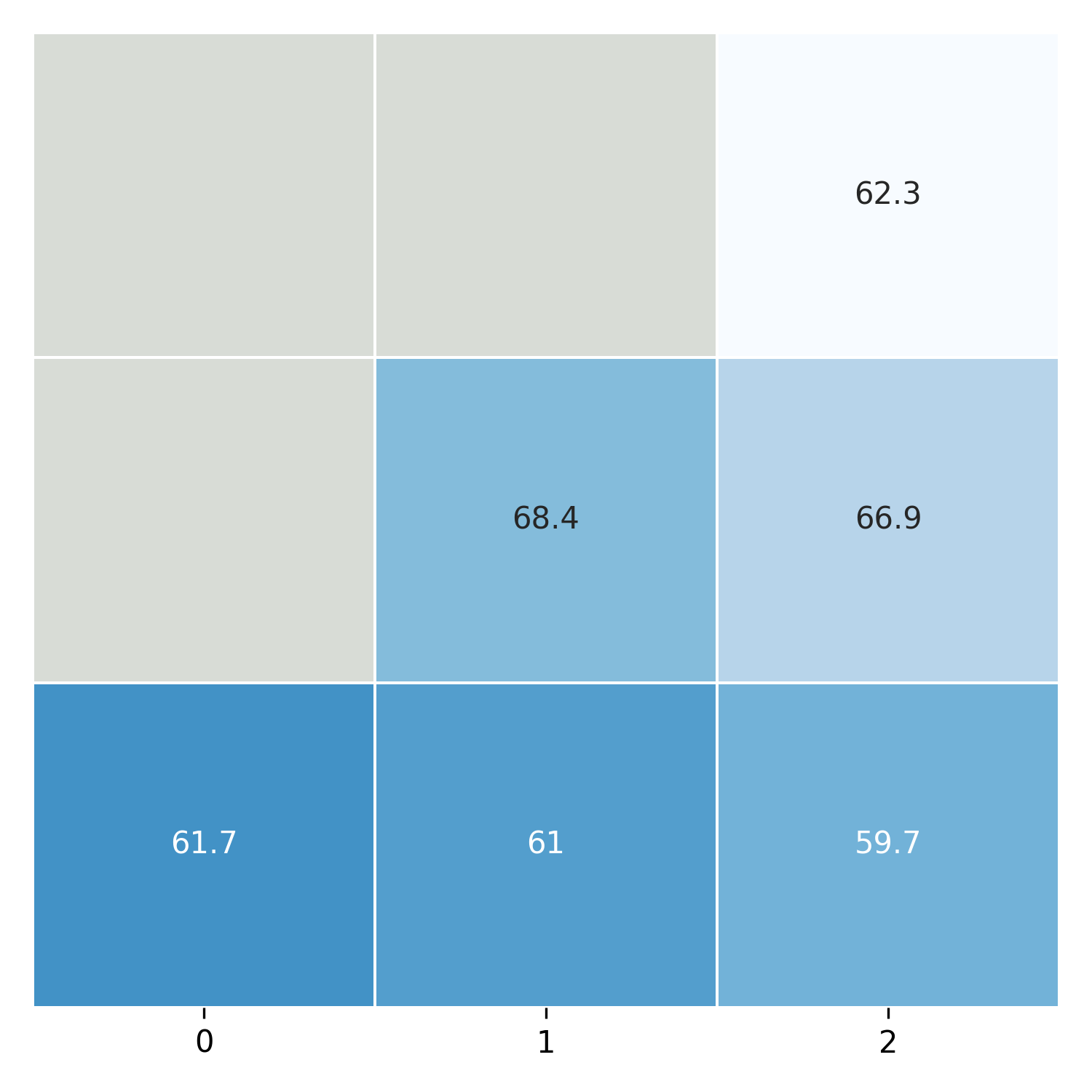}
    \caption{SI}
    \label{fig:fairface_si}
\end{subfigure}
\begin{subfigure}[b]{0.13\textwidth}
    \centering
    \includegraphics[width=\textwidth]{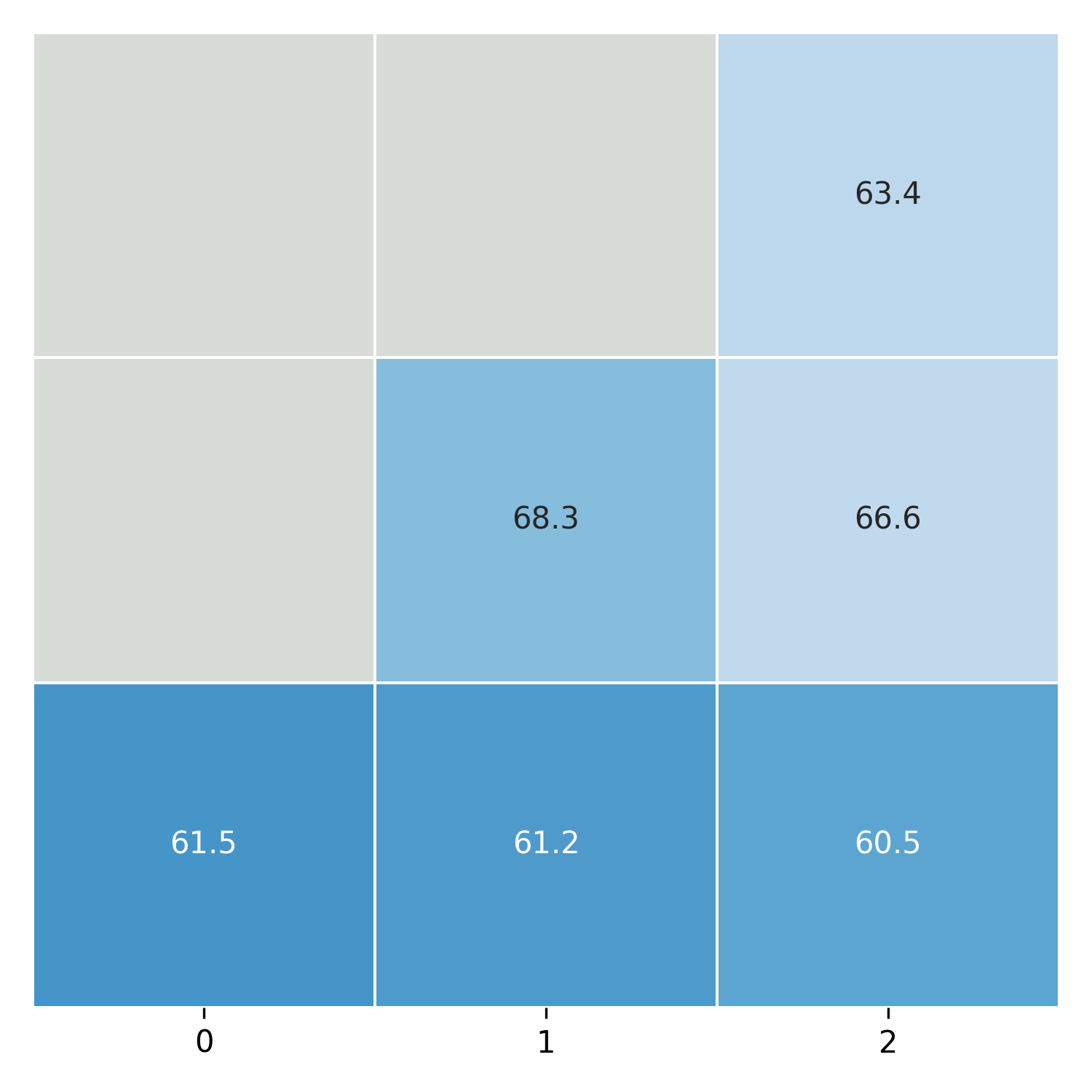}
    \caption{oEWC}
    \label{fig:fairface_ewc}
\end{subfigure}
\caption{Confusion matrices for different models on the FairFace dataset}
\label{fig:fairface_confusion_matrices}
\end{figure*}

\begin{figure*}[htbp]
\centering
\begin{subfigure}[b]{0.13\textwidth}
    \centering
    \includegraphics[width=\textwidth]{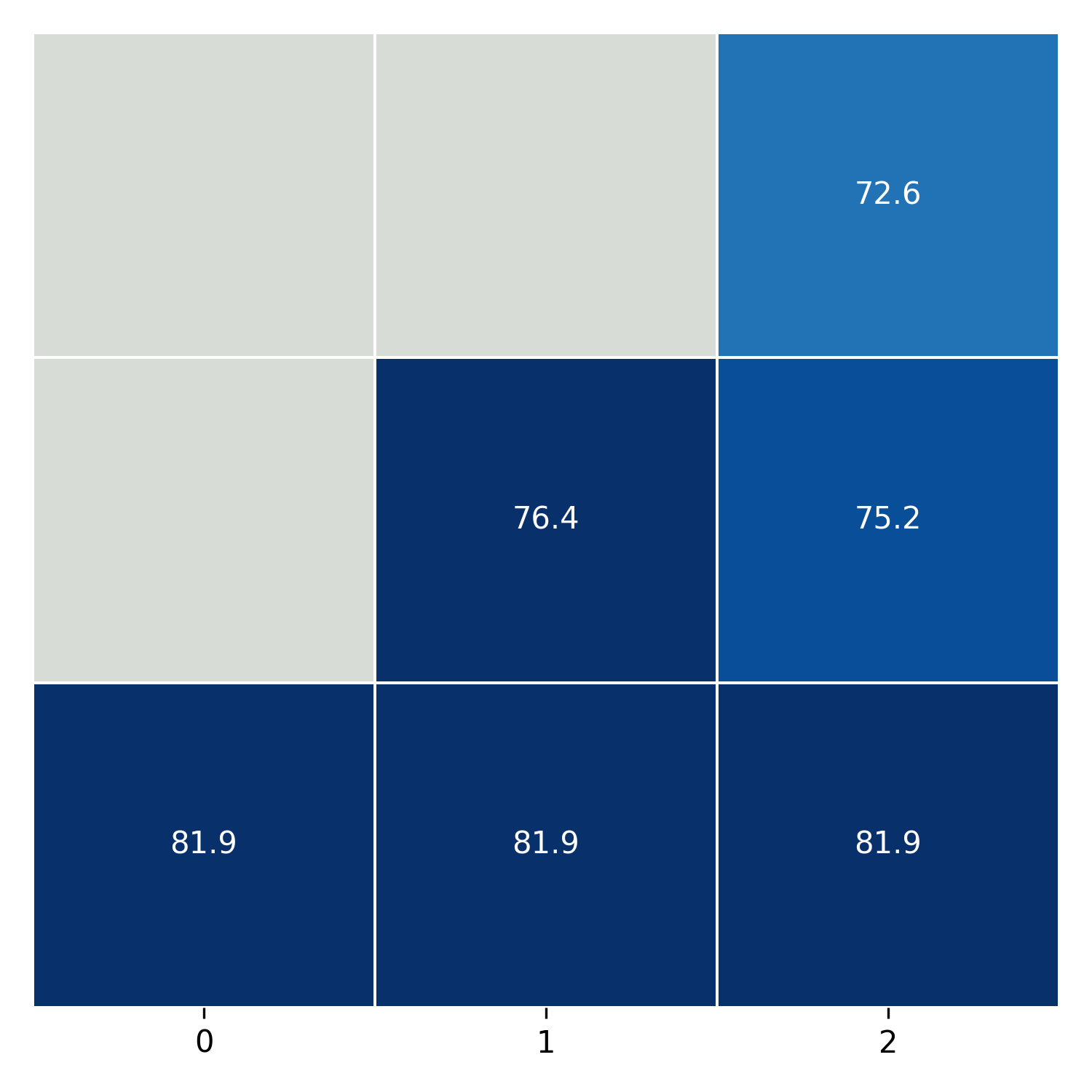}
    \caption{LwP \textbf{Ours}}
    \label{fig:bdd100k_lwp}
\end{subfigure}
\begin{subfigure}[b]{0.13\textwidth}
    \centering
    \includegraphics[width=\textwidth]{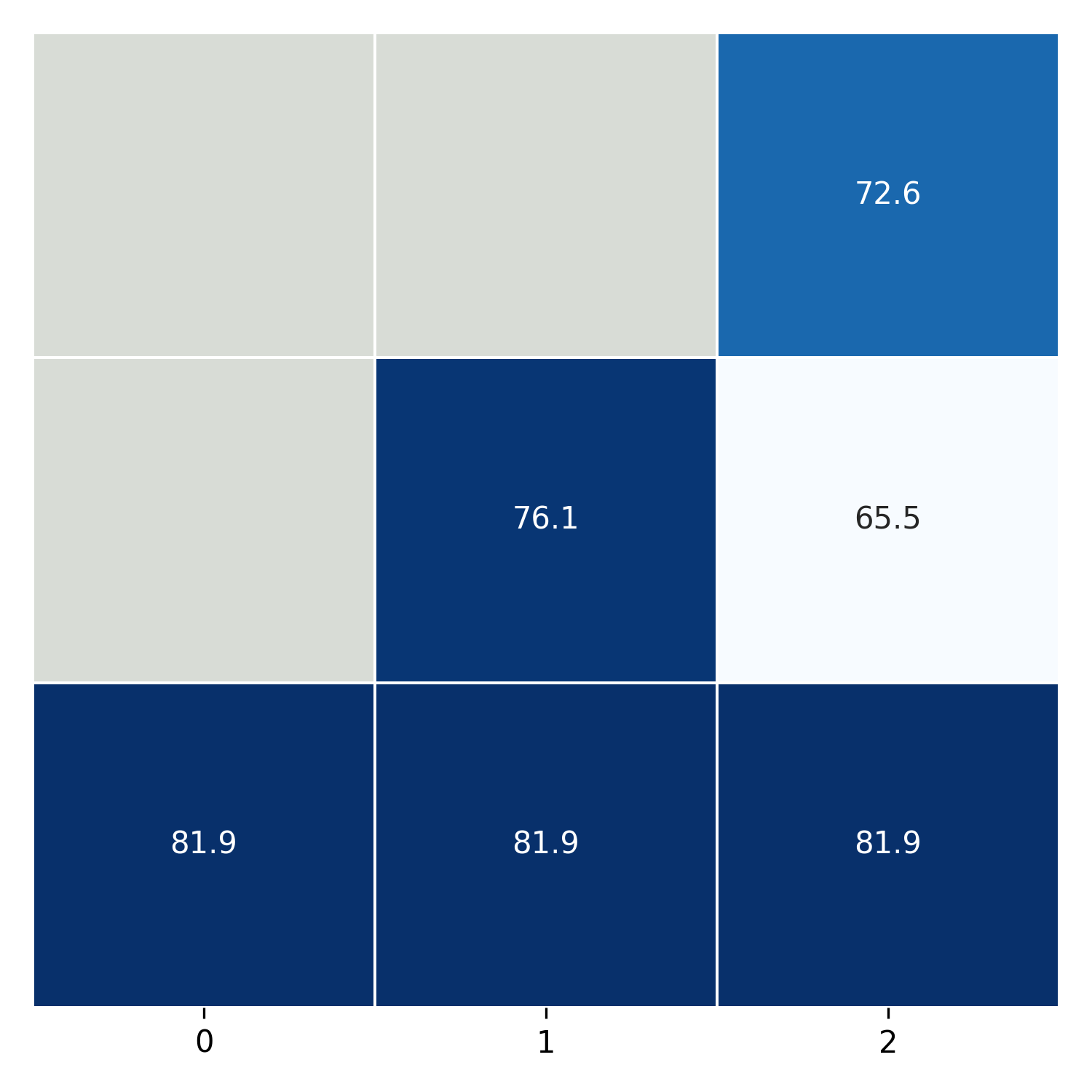}
    \caption{LwF}
    \label{fig:bdd100k_lwf}
\end{subfigure}
\begin{subfigure}[b]{0.13\textwidth}
    \centering
    \includegraphics[width=\textwidth]{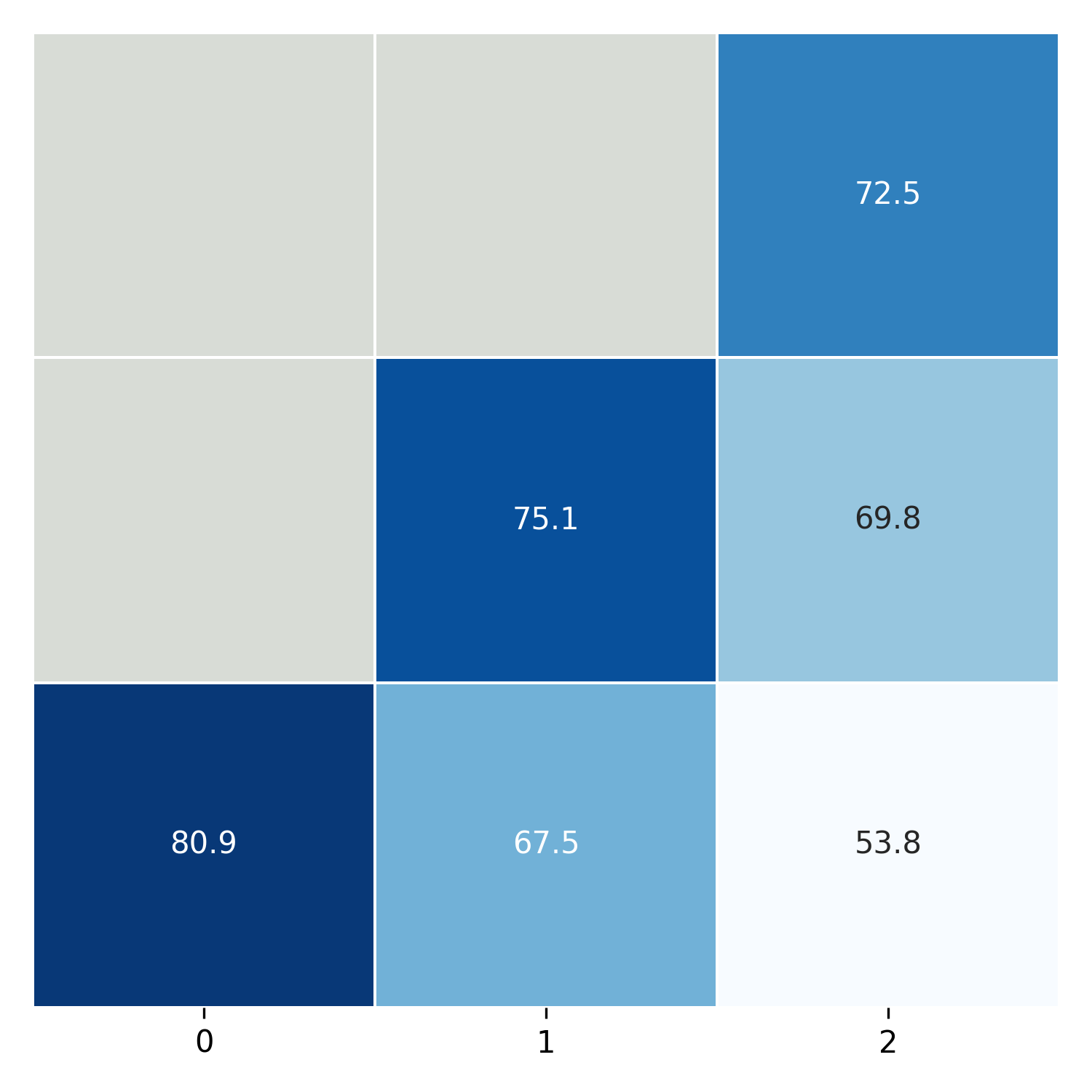}
    \caption{ER}
    \label{fig:bdd100k_er}
\end{subfigure}
\begin{subfigure}[b]{0.13\textwidth}
    \centering
    \includegraphics[width=\textwidth]{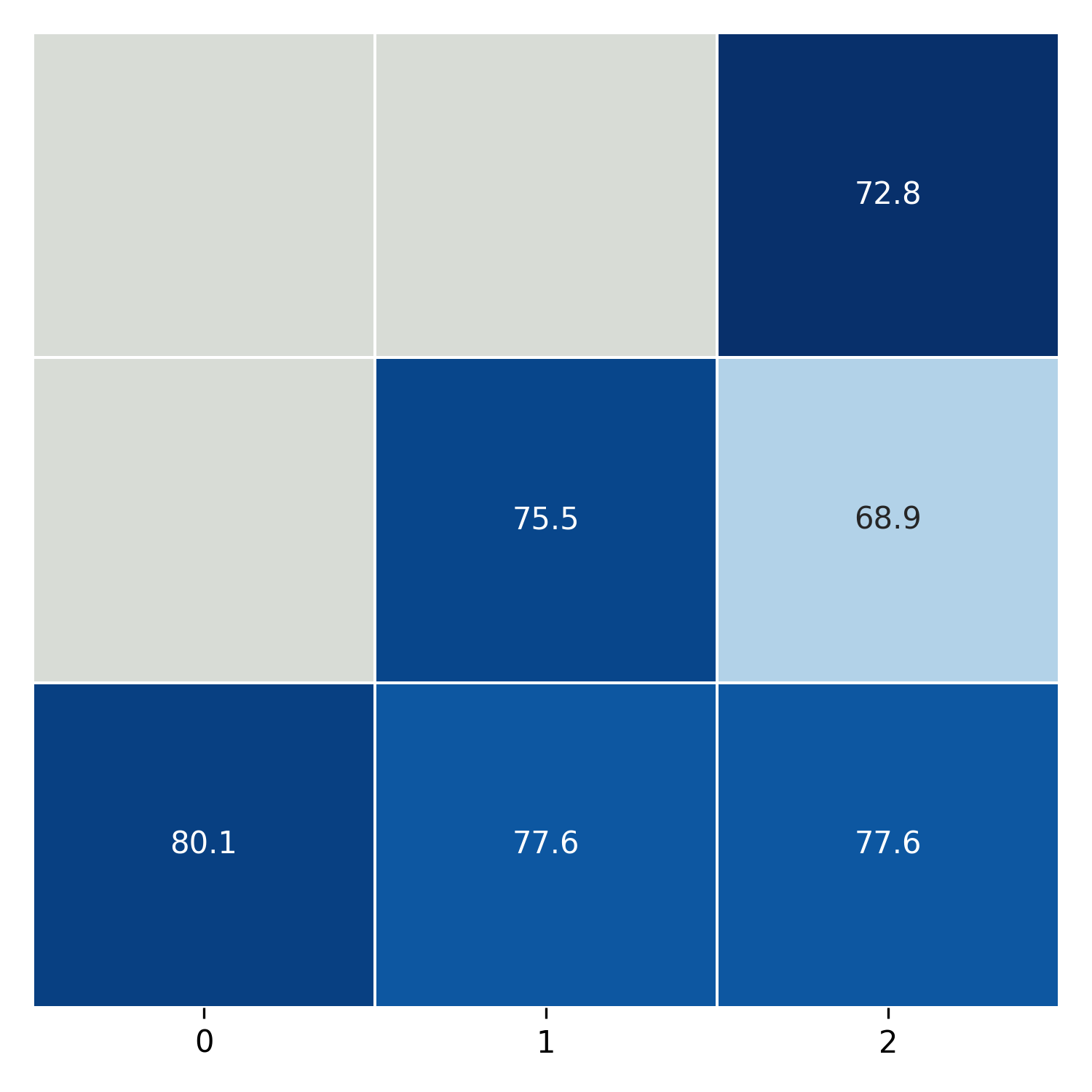}
    \caption{GSS}
    \label{fig:bdd100k_gss}
\end{subfigure}
\begin{subfigure}[b]{0.13\textwidth}
    \centering
    \includegraphics[width=\textwidth]{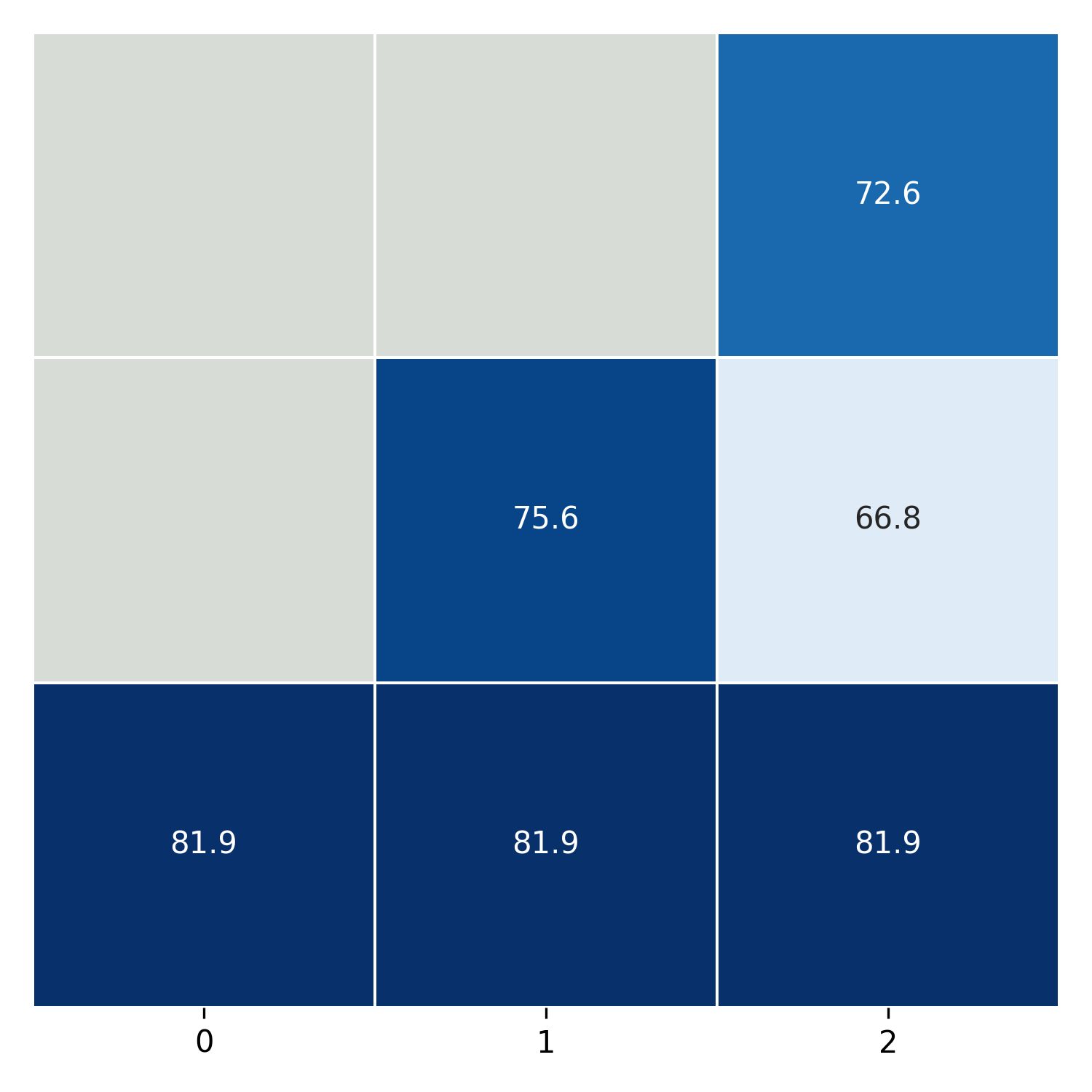}
    \caption{FDR}
    \label{fig:bdd100k_fdr}
\end{subfigure}

\begin{subfigure}[b]{0.13\textwidth}
    \centering
    \includegraphics[width=\textwidth]{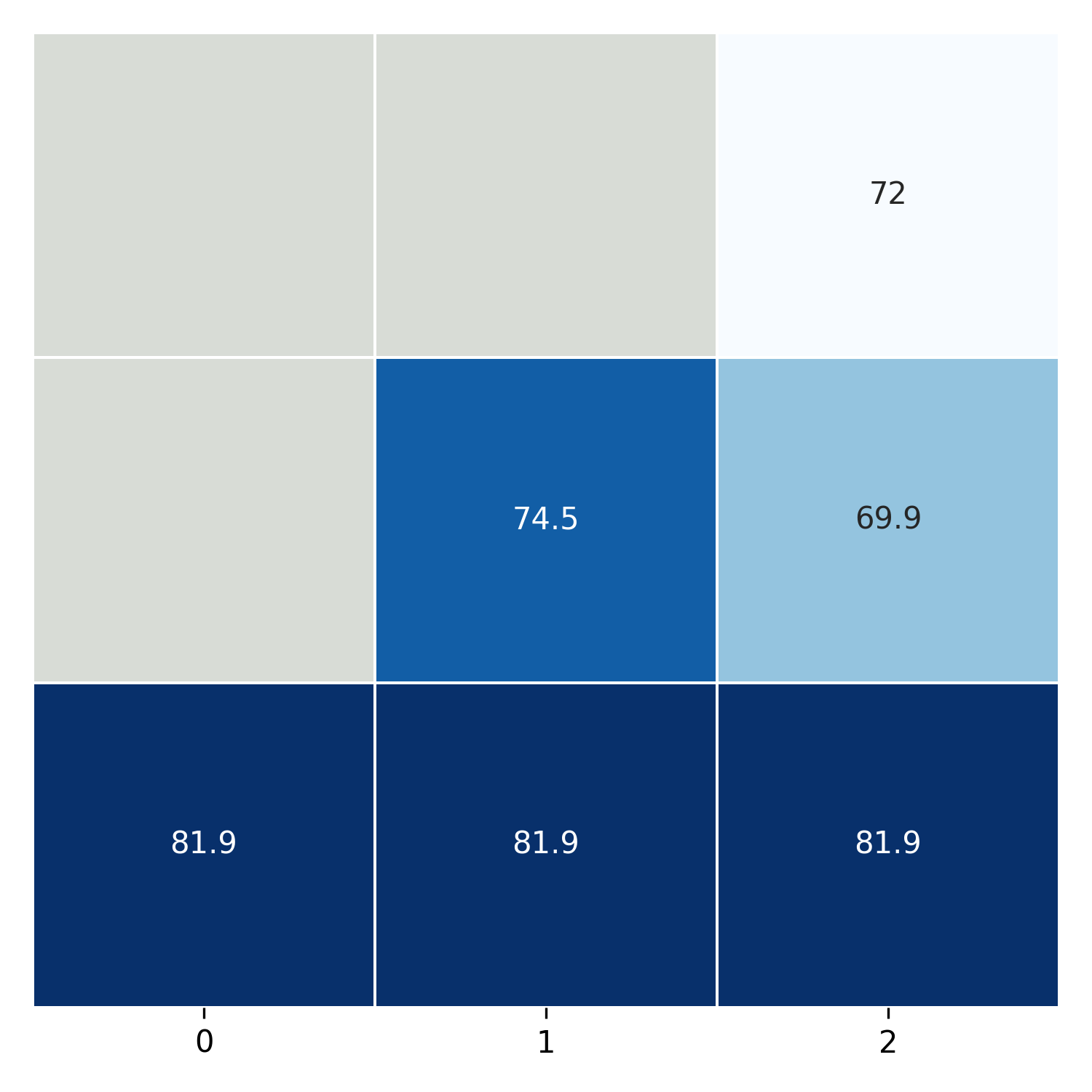}
    \caption{OBC}
    \label{fig:bdd100k_OBC}
\end{subfigure}
\begin{subfigure}[b]{0.13\textwidth}
    \centering
    \includegraphics[width=\textwidth]{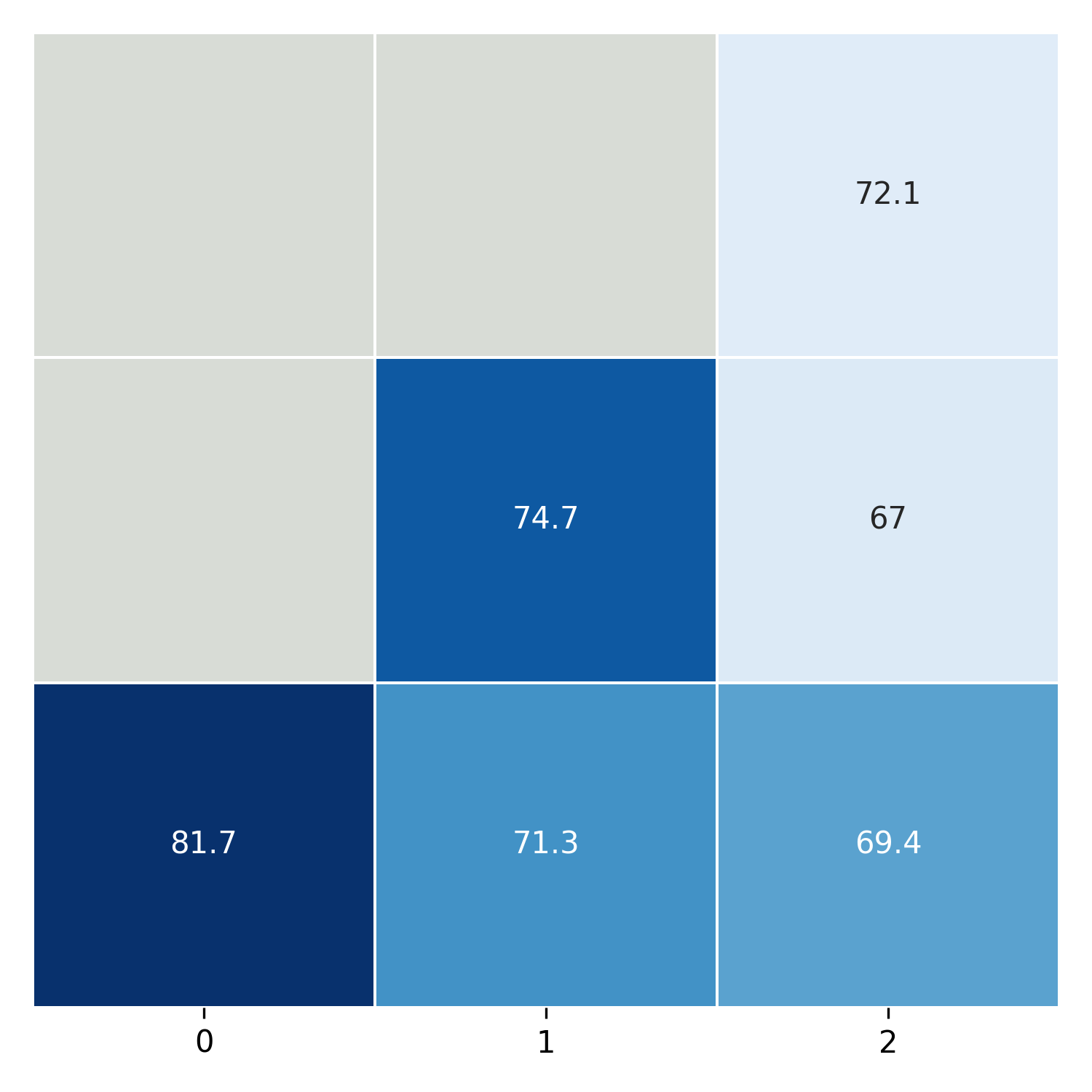}
    \caption{DVC}
    \label{fig:bdd100k_DVC}
\end{subfigure}
\begin{subfigure}[b]{0.13\textwidth}
    \centering
    \includegraphics[width=\textwidth]{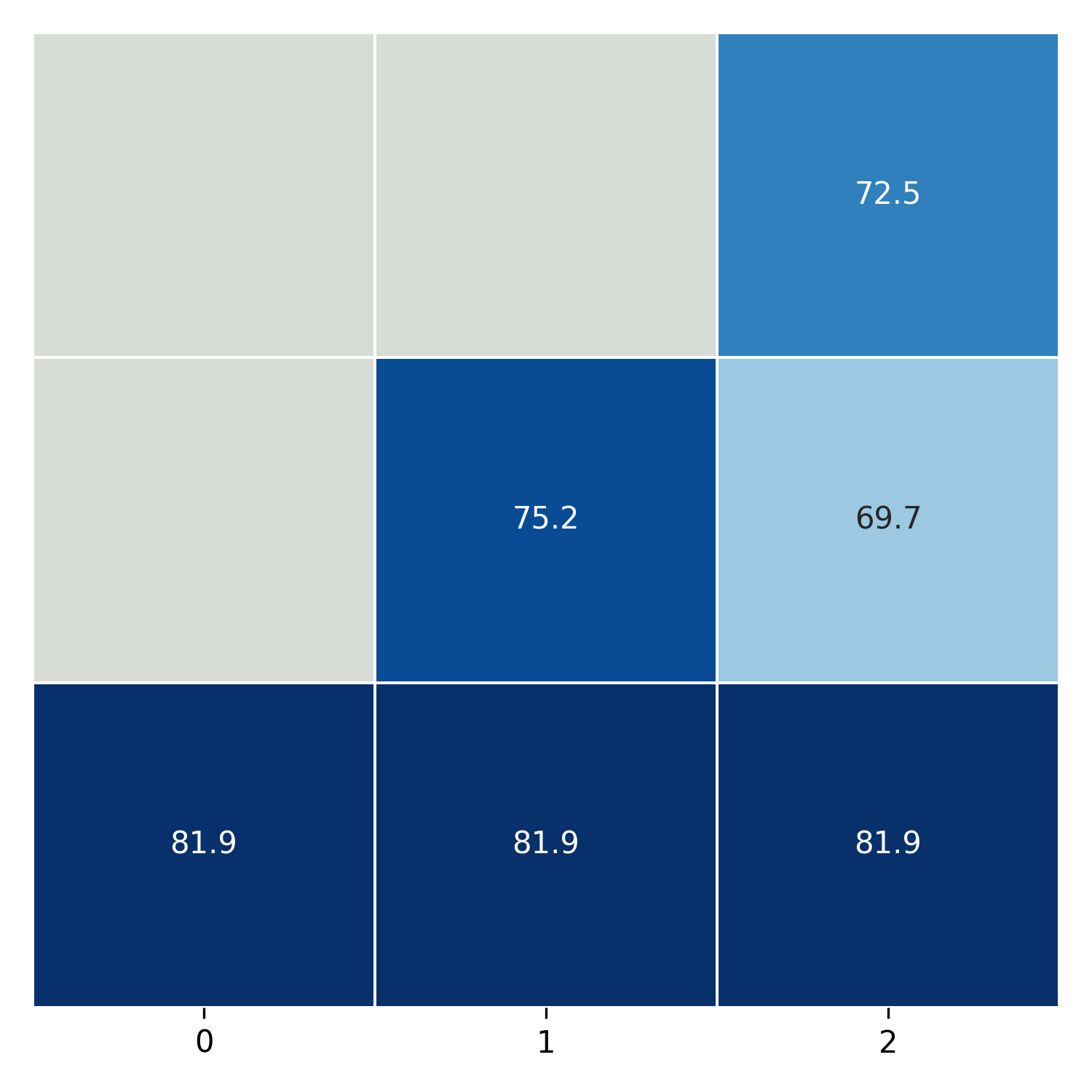}
    \caption{DER}
    \label{fig:bdd100k_der}
\end{subfigure}
\begin{subfigure}[b]{0.13\textwidth}
    \centering
    \includegraphics[width=\textwidth]{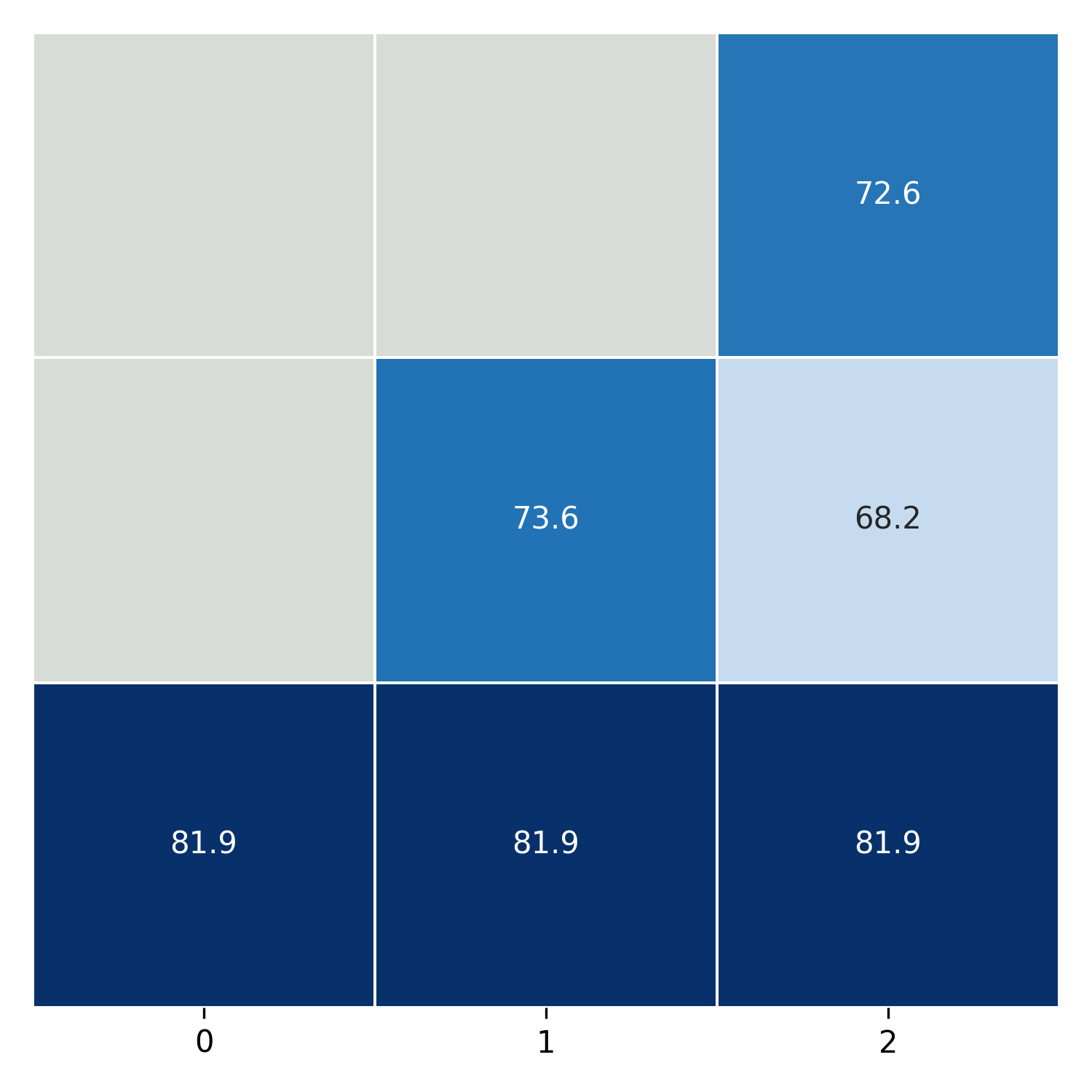}
    \caption{DERPP}
    \label{fig:bdd100k_derpp}
\end{subfigure}
\begin{subfigure}[b]{0.13\textwidth}
    \centering
    \includegraphics[width=\textwidth]{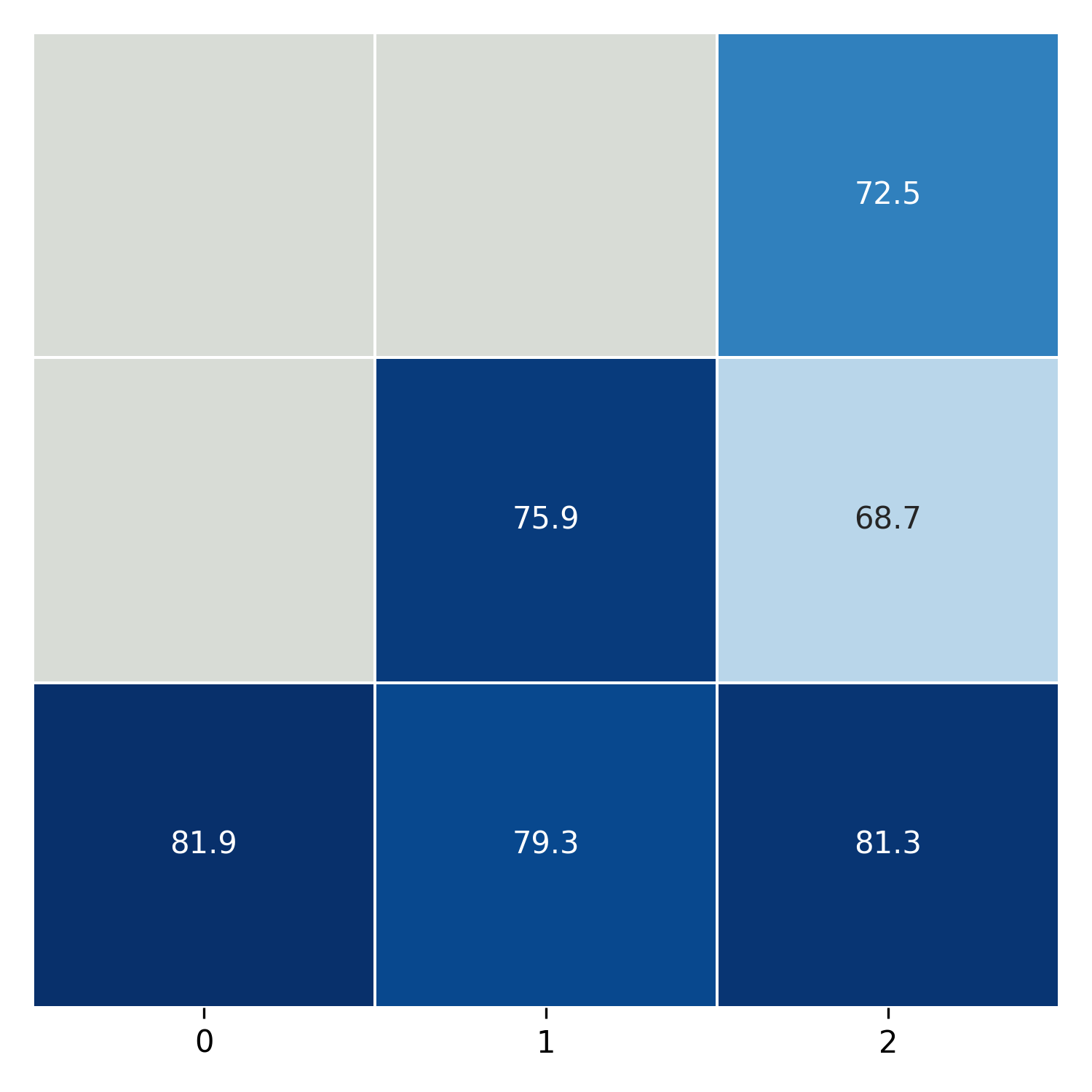}
    \caption{SI}
    \label{fig:bdd100k_si}
\end{subfigure}
\begin{subfigure}[b]{0.13\textwidth}
    \centering
    \includegraphics[width=\textwidth]{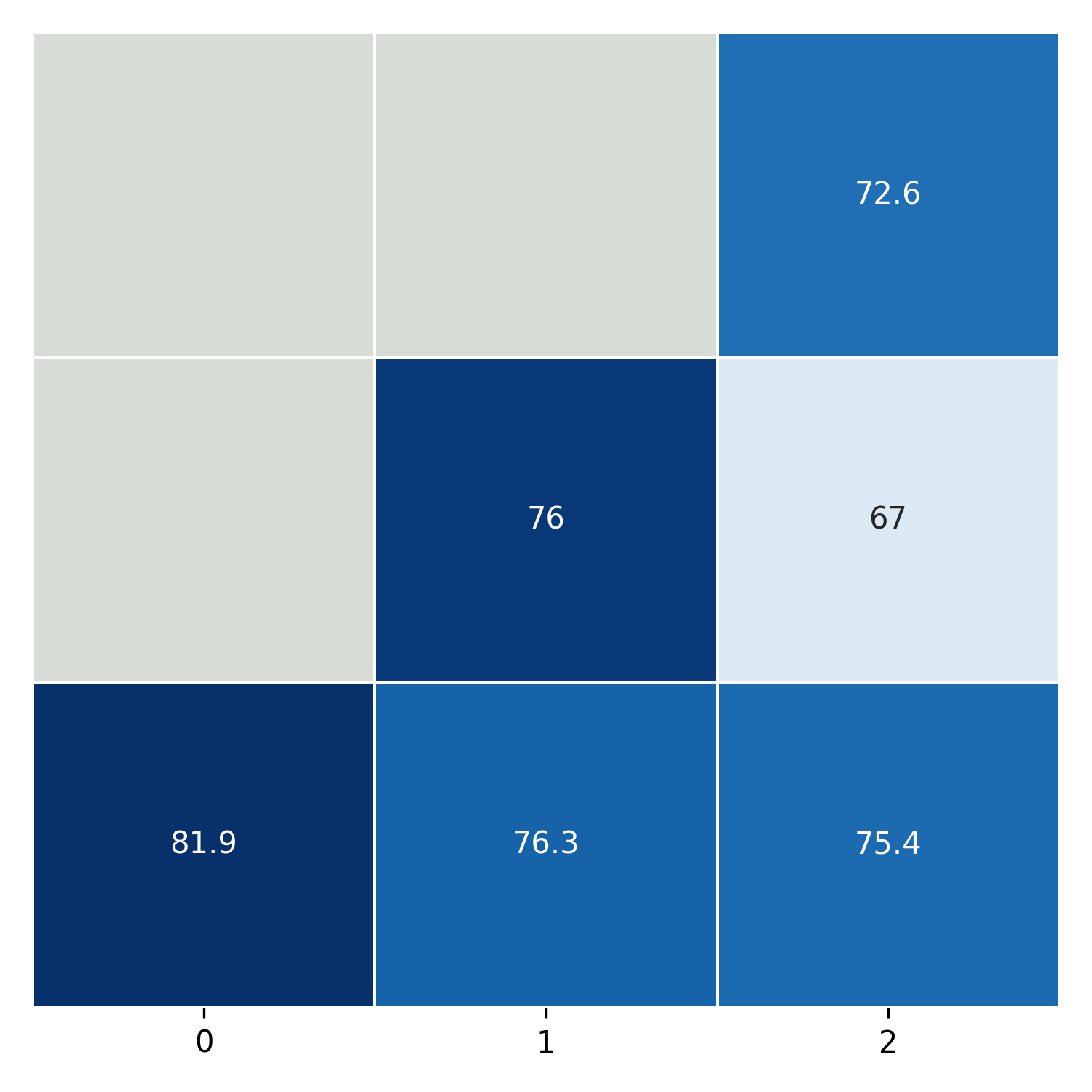}
    \caption{oEWC}
    \label{fig:bdd100k_ewc}
\end{subfigure}
\caption{Confusion matrices for different models on the BDD100K dataset}
\label{fig:bdd100k_confusion_matrices}
\end{figure*}

\begin{figure*}[htbp]
\centering
\begin{subfigure}[b]{0.13\textwidth}
    \centering
    \includegraphics[width=\textwidth]{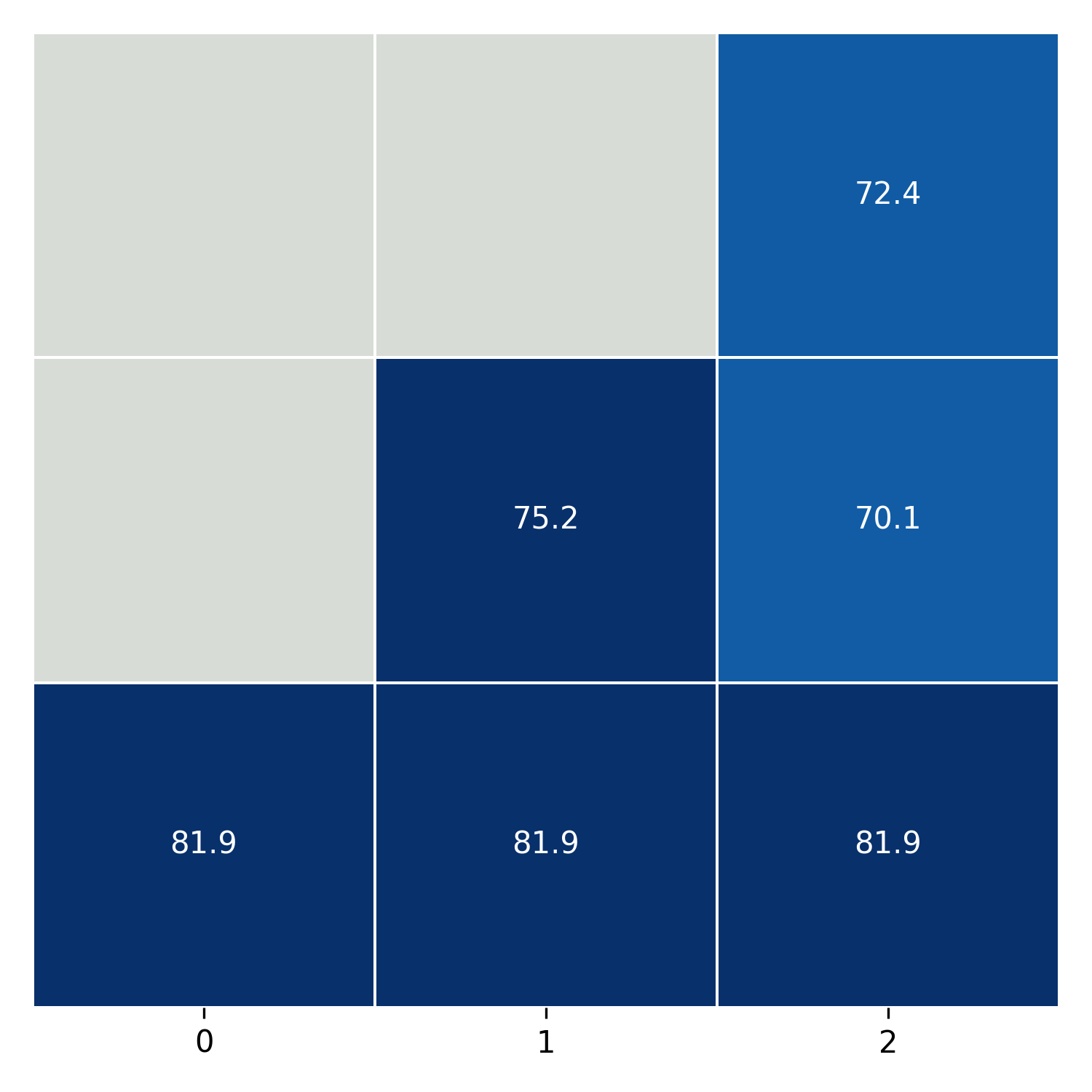}
    \caption{LwP \textbf{Ours}}
    \label{fig:bdd100k_lwp_weather_shift}
\end{subfigure}
\begin{subfigure}[b]{0.13\textwidth}
    \centering
    \includegraphics[width=\textwidth]{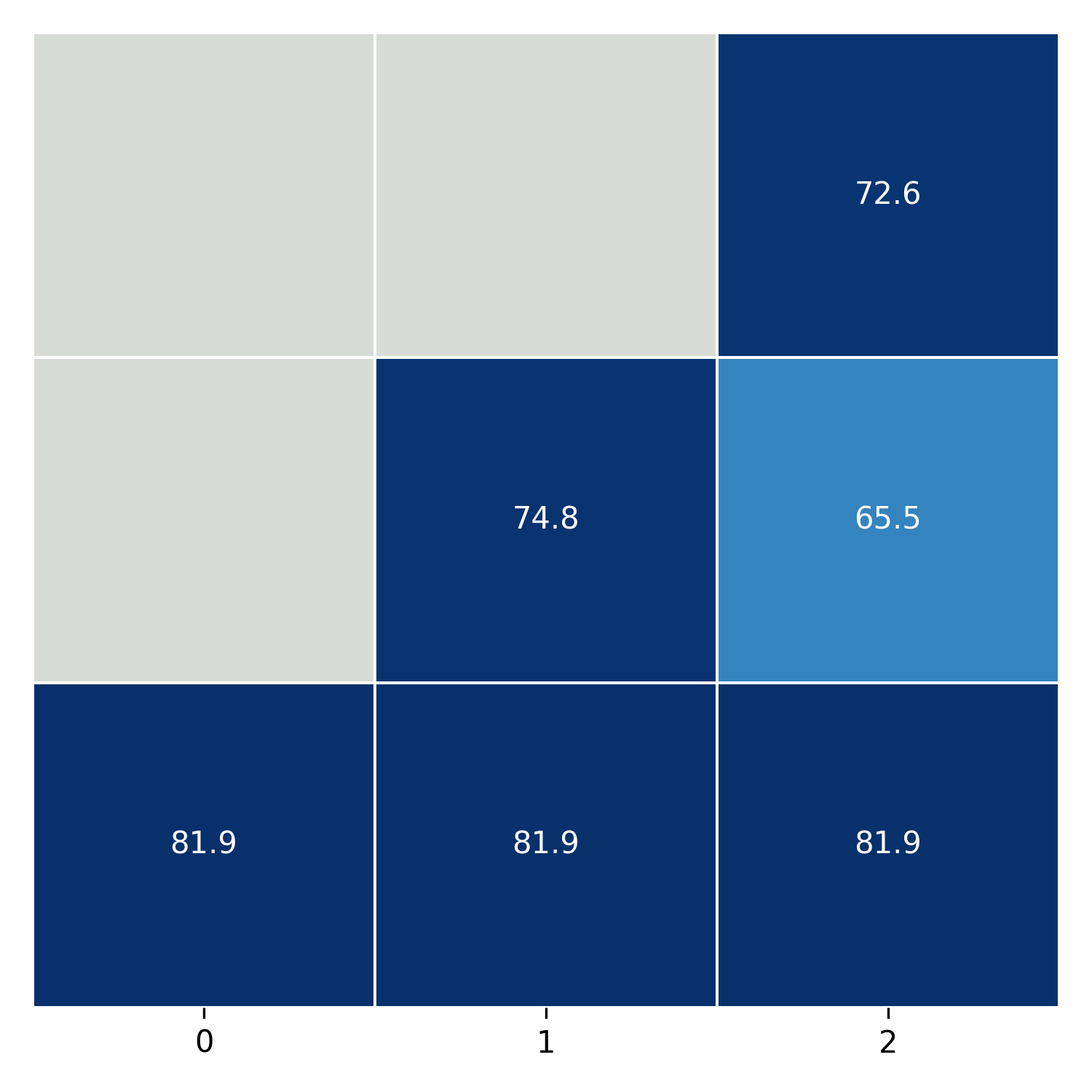}
    \caption{LwF}
    \label{fig:bdd100k_lwf_weather_shift}
\end{subfigure}
\begin{subfigure}[b]{0.13\textwidth}
    \centering
    \includegraphics[width=\textwidth]{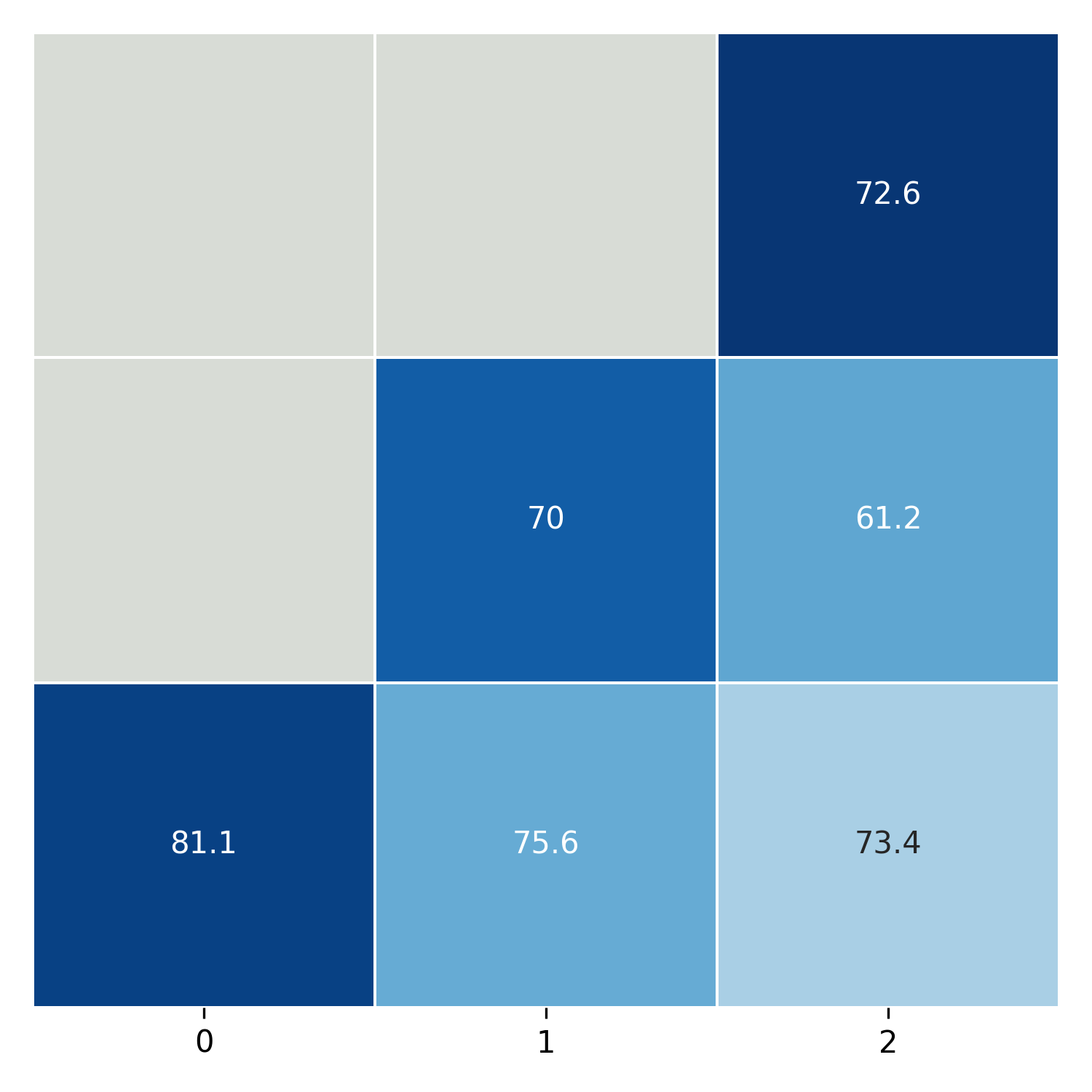}
    \caption{ER}
    \label{fig:bdd100k_er_weather_shift}
\end{subfigure}
\begin{subfigure}[b]{0.13\textwidth}
    \centering
    \includegraphics[width=\textwidth]{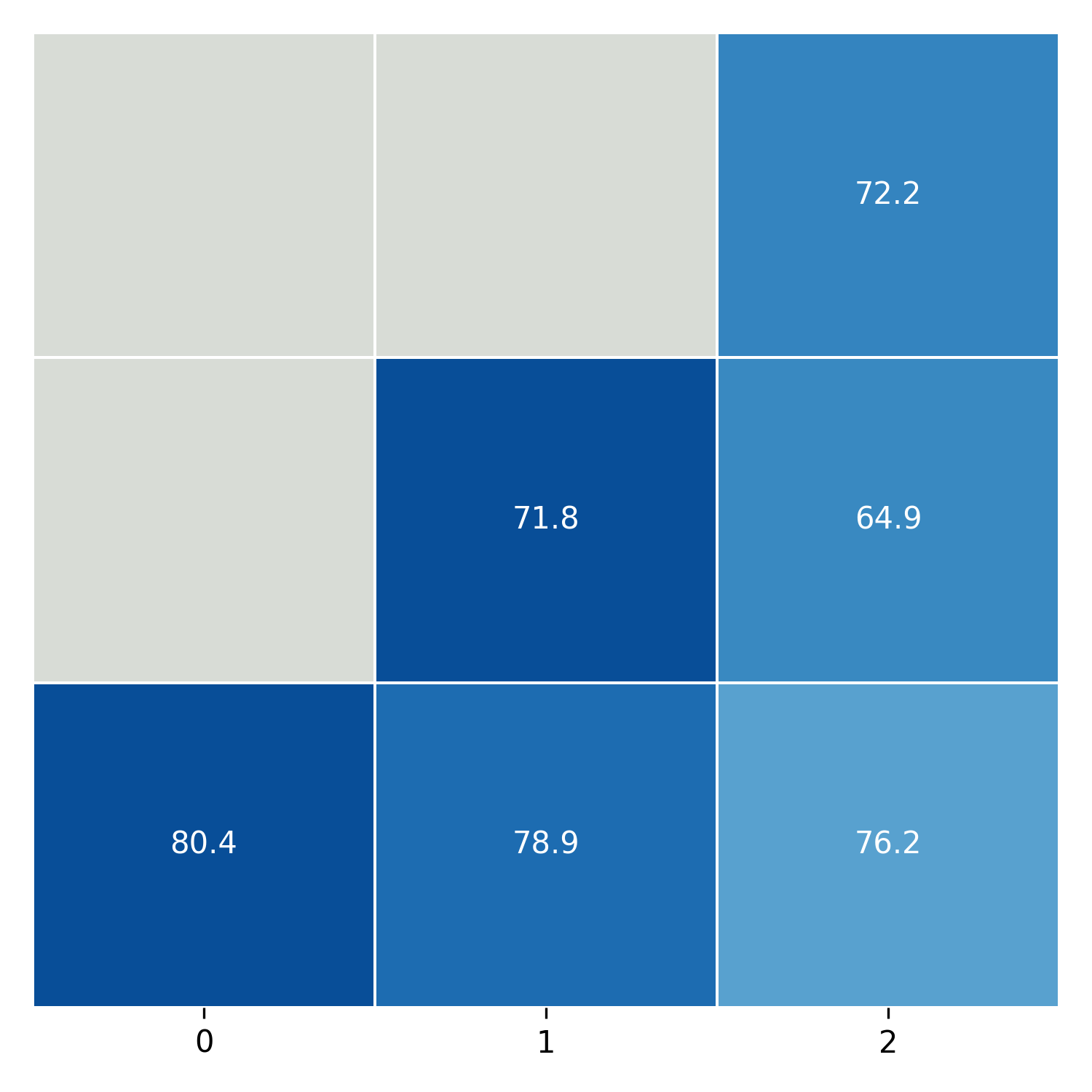}
    \caption{GSS}
    \label{fig:bdd100k_gss_weather_shift}
\end{subfigure}
\begin{subfigure}[b]{0.13\textwidth}
    \centering
    \includegraphics[width=\textwidth]{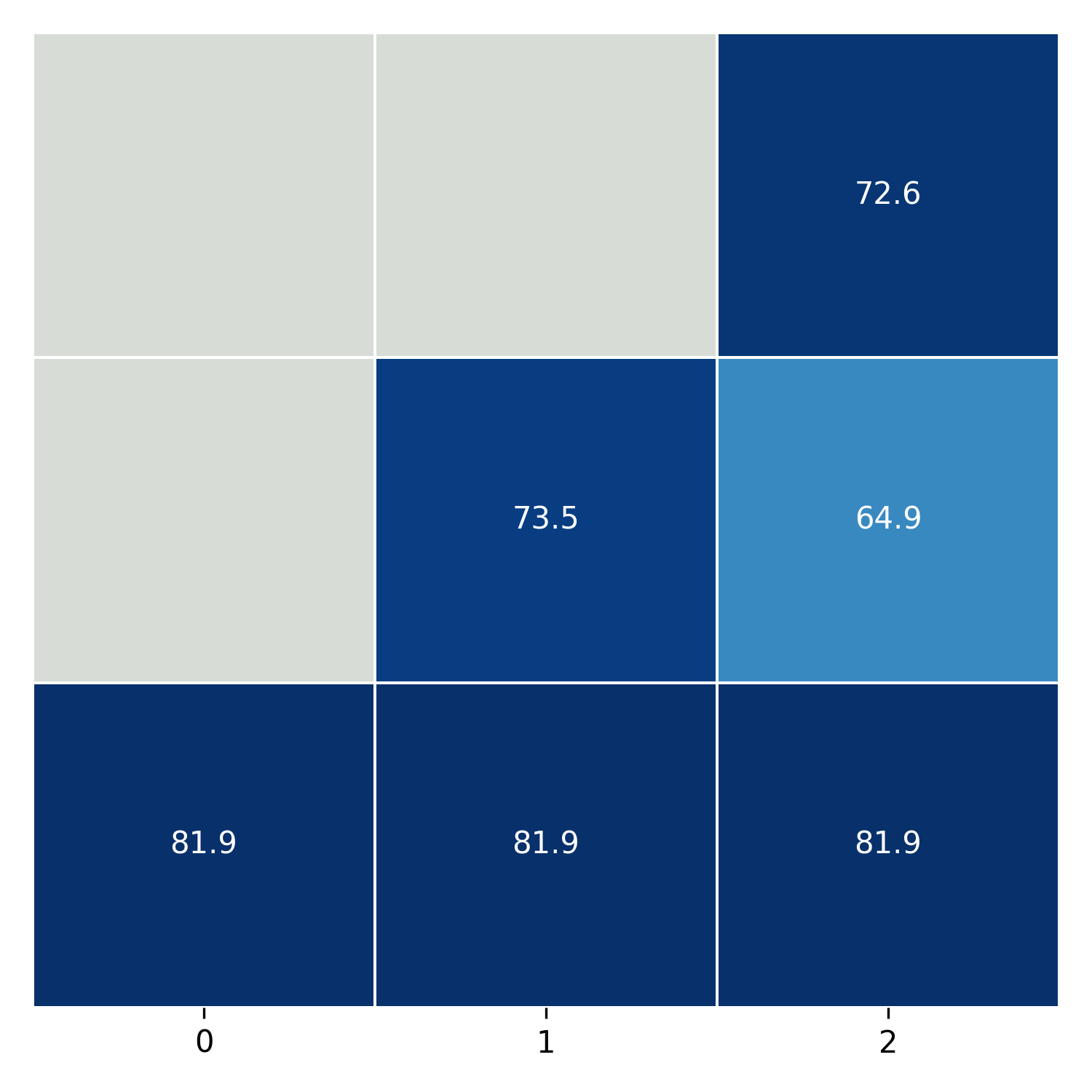}
    \caption{FDR}
    \label{fig:bdd100k_fdr_weather_shift}
\end{subfigure}

\begin{subfigure}[b]{0.13\textwidth}
    \centering
    \includegraphics[width=\textwidth]{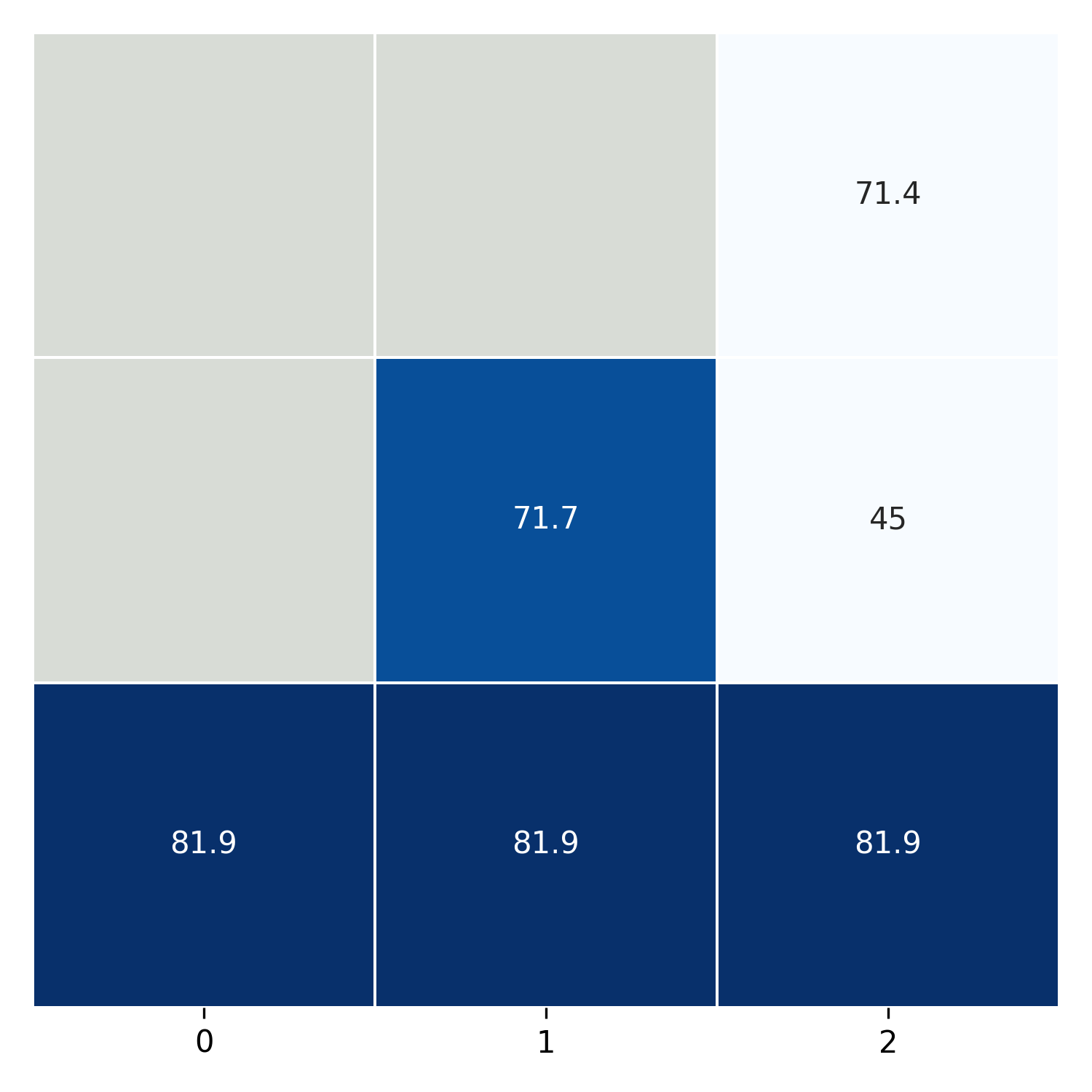}
    \caption{OBC}
    \label{fig:bdd100k_OBC_weather_shift}
\end{subfigure}
\begin{subfigure}[b]{0.13\textwidth}
    \centering
    \includegraphics[width=\textwidth]{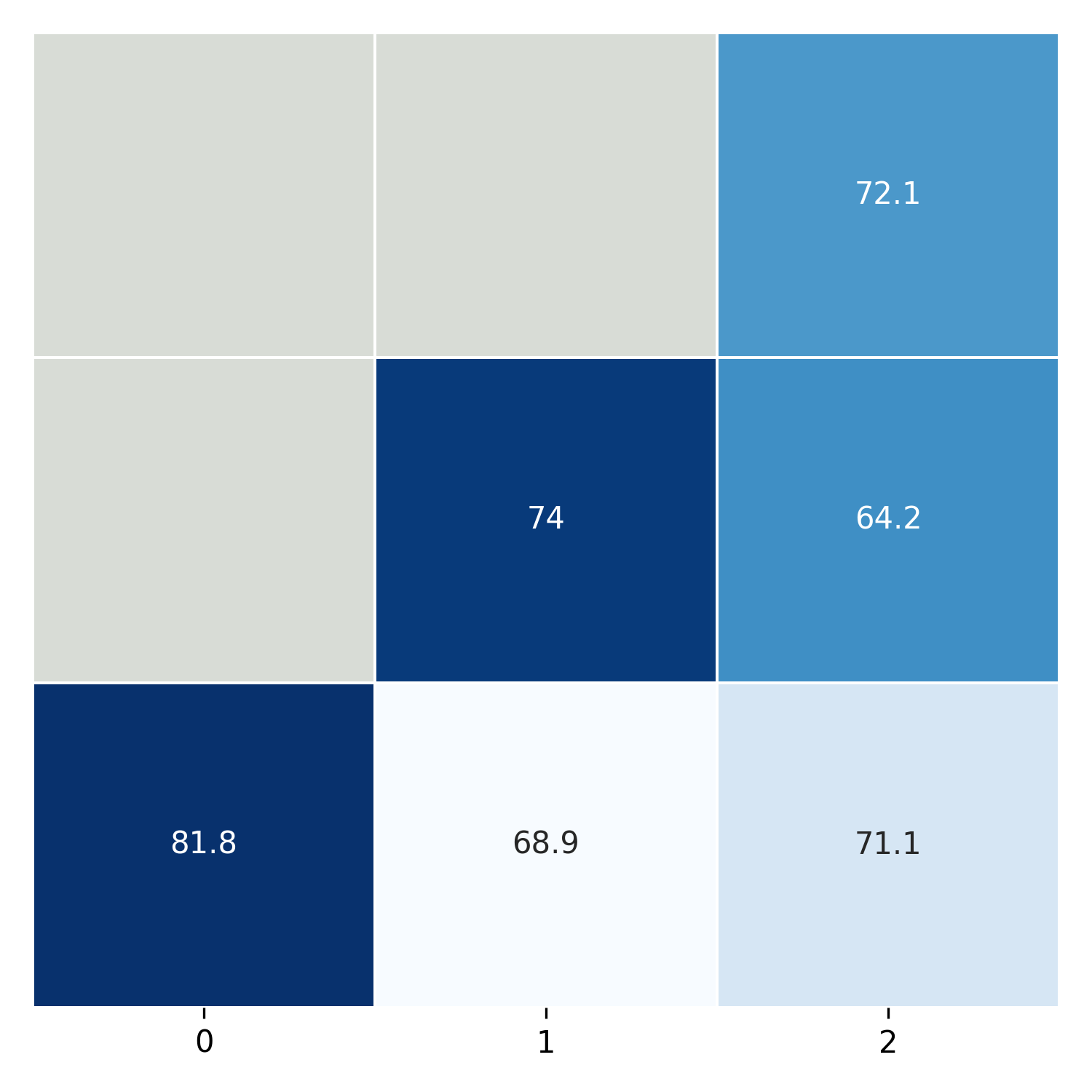}
    \caption{DVC}
    \label{fig:bdd100k_DVC_weather_shift}
\end{subfigure}
\begin{subfigure}[b]{0.13\textwidth}
    \centering
    \includegraphics[width=\textwidth]{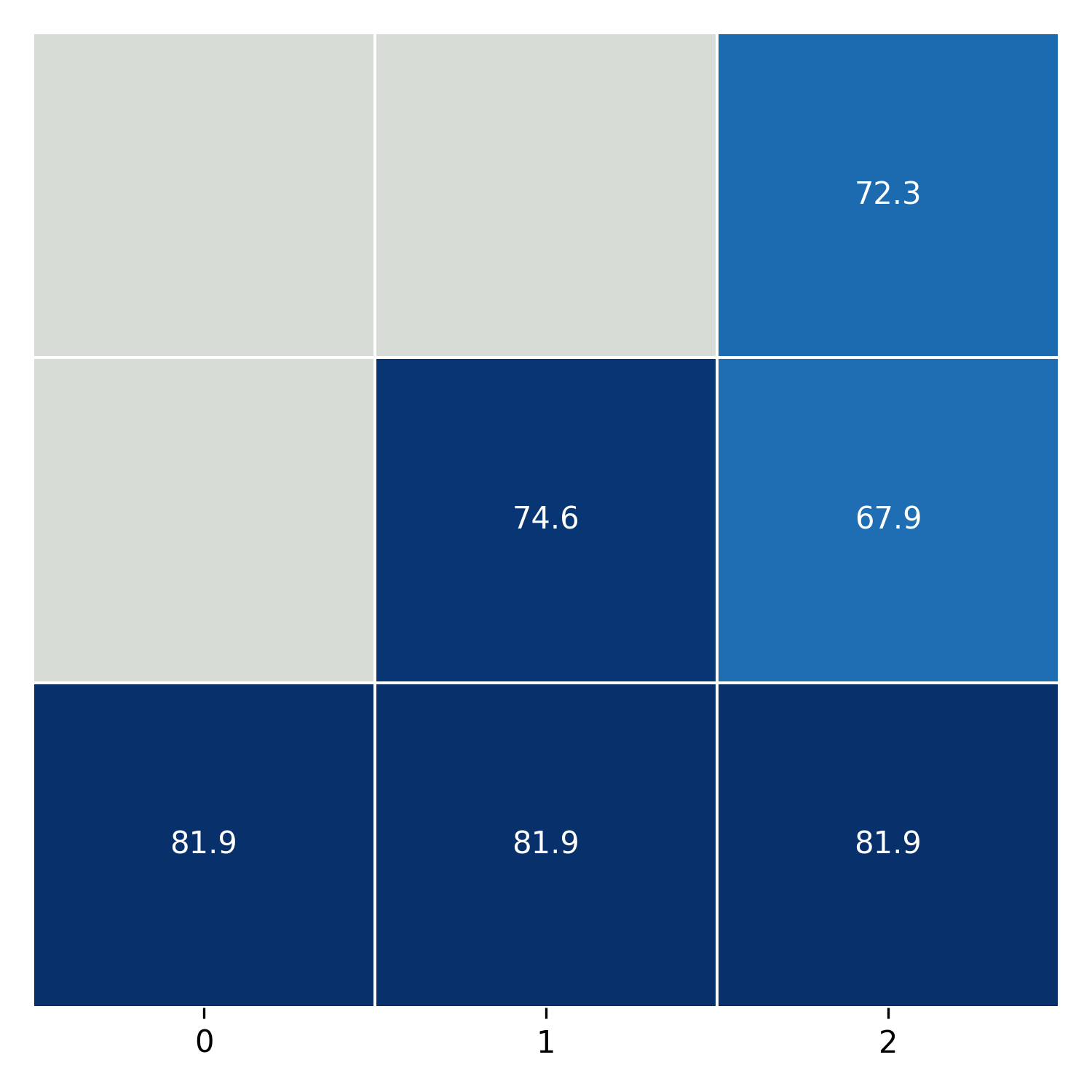}
    \caption{DER}
    \label{fig:bdd100k_der_weather_shift}
\end{subfigure}
\begin{subfigure}[b]{0.13\textwidth}
    \centering
    \includegraphics[width=\textwidth]{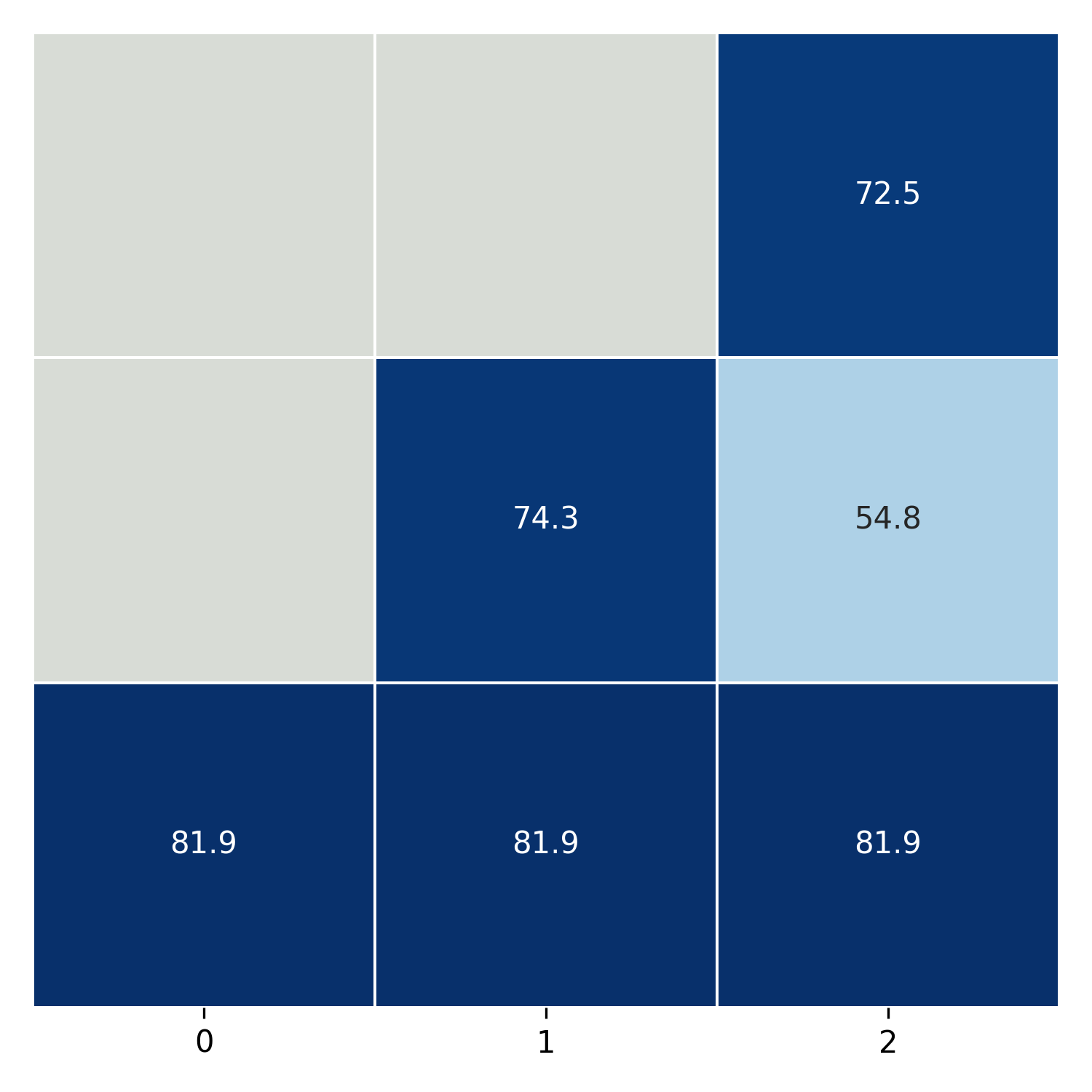}
    \caption{DERPP}
    \label{fig:bdd100k_derpp_weather_shift}
\end{subfigure}
\begin{subfigure}[b]{0.13\textwidth}
    \centering
    \includegraphics[width=\textwidth]{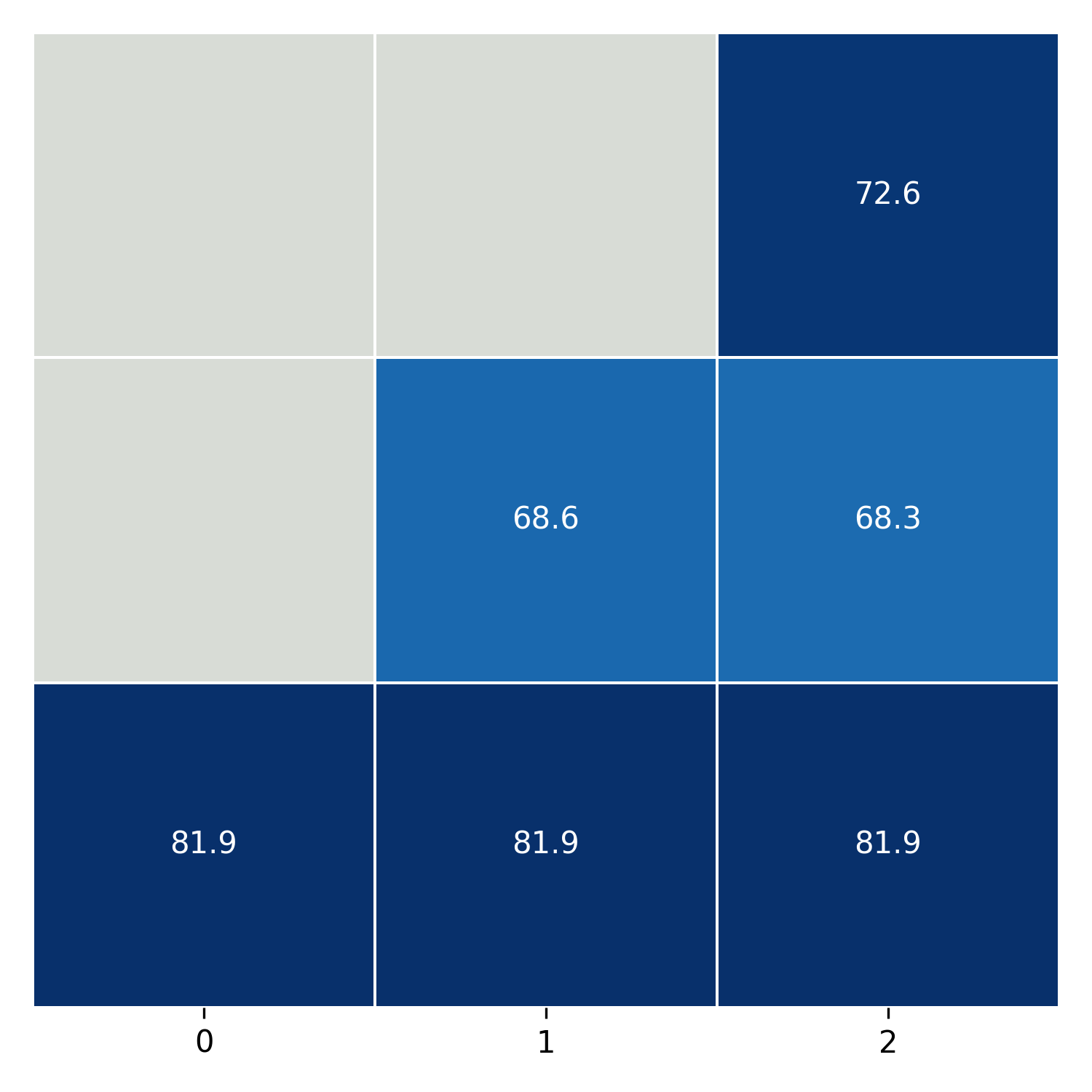}
    \caption{SI}
    \label{fig:bdd100k_si_weather_shift}
\end{subfigure}
\begin{subfigure}[b]{0.13\textwidth}
    \centering
    \includegraphics[width=\textwidth]{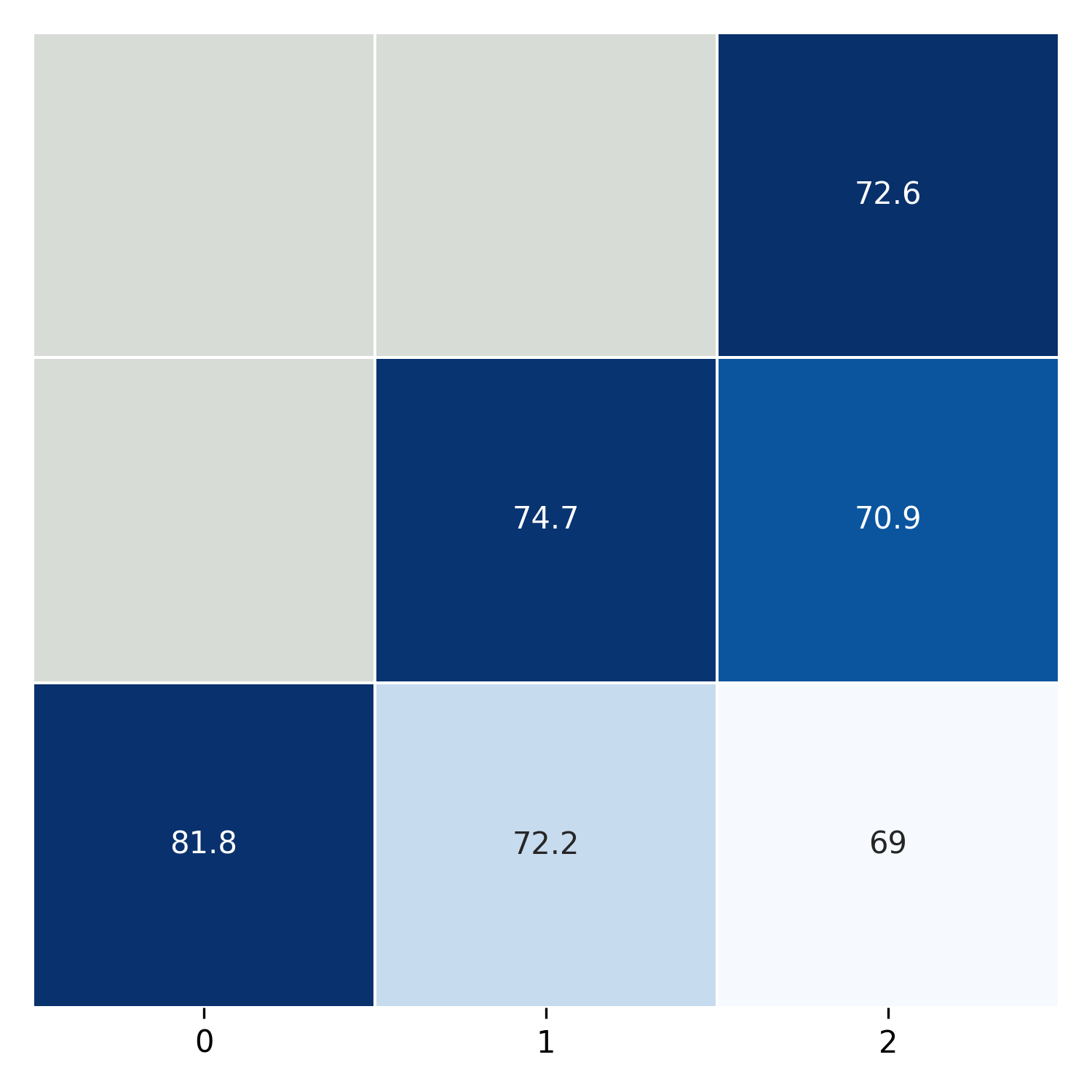}
    \caption{oEWC}
    \label{fig:bdd100k_ewc_weather_shift}
\end{subfigure}
\caption{Confusion matrices for different models on the BDD100K dataset under a \textbf{weather-based distribution shift}.}
\label{fig:bdd100k_weather_shift_confusion_matrices}
\end{figure*}

Figures \ref{fig:physiq_confusion_matrices},  \ref{fig:fairface_confusion_matrices}, and \ref{fig:bdd100k_confusion_matrices} illustrate that the application of LwP reduces the issue of catastrophic forgetting in the PhysiQ and FairFace datasets as well. This effect is particularly pronounced when applied to datasets with a large number of samples, such as Fairface, CelebA, and BDD100K, in comparison to smaller datasets such as PhysiQ. These observations imply that LwP is a scalable and effective solution to mitigate catastrophic forgetting in continual multitask learning models. We further investigate the effect of the number of training samples on performance in \ref{appendix:training_sample}.

\subsection{Influence of Training Sample}\label{appendix:training_sample}

\begin{figure*}[htbp]
    \centering
    \begin{subfigure}{0.4\textwidth}
        \centering
        \includegraphics[width=\linewidth]{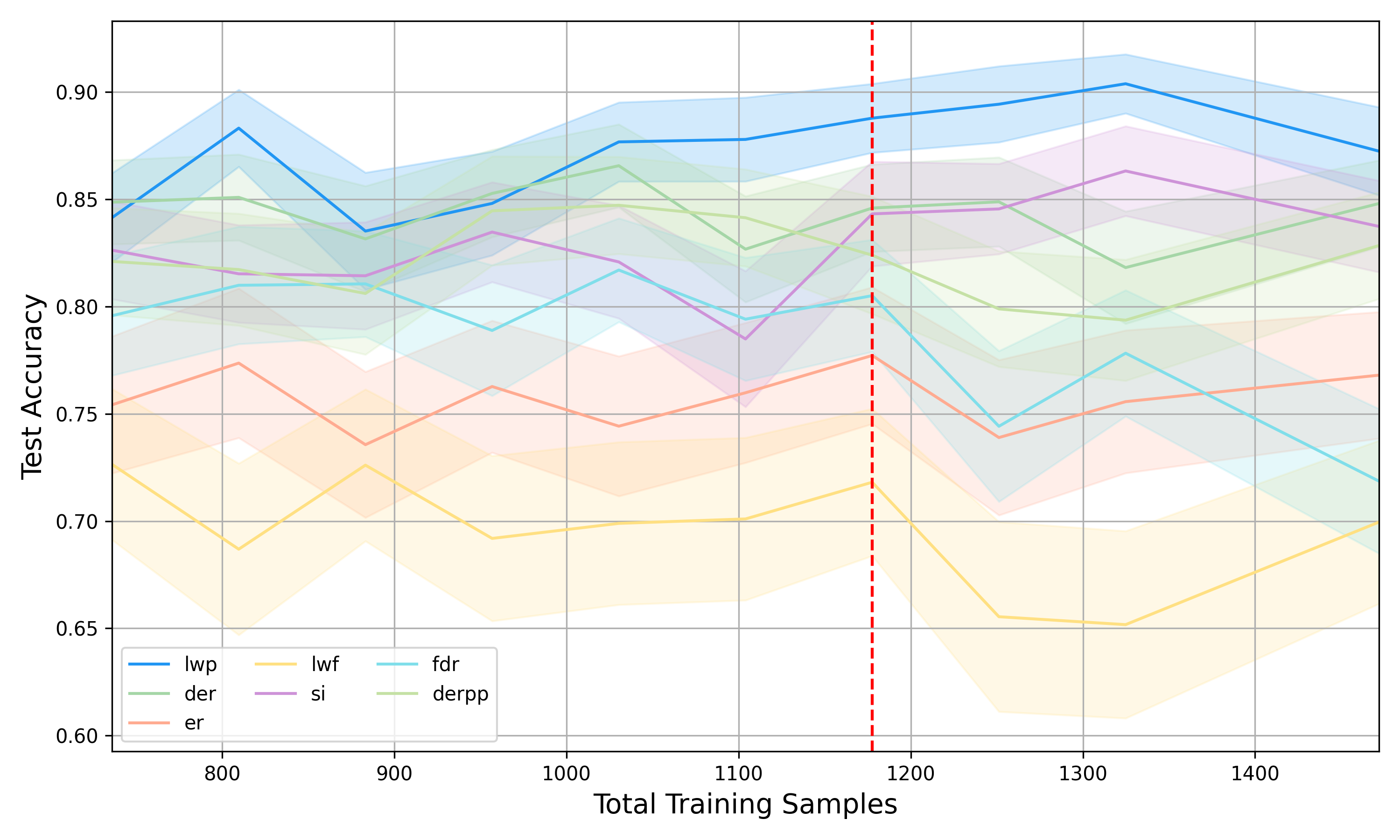}
        \caption{Training Sample Accuracy}
        \label{fig:subfig1_acc}
    \end{subfigure}
    \begin{subfigure}{0.4\textwidth}
        \centering
        \includegraphics[width=\linewidth]{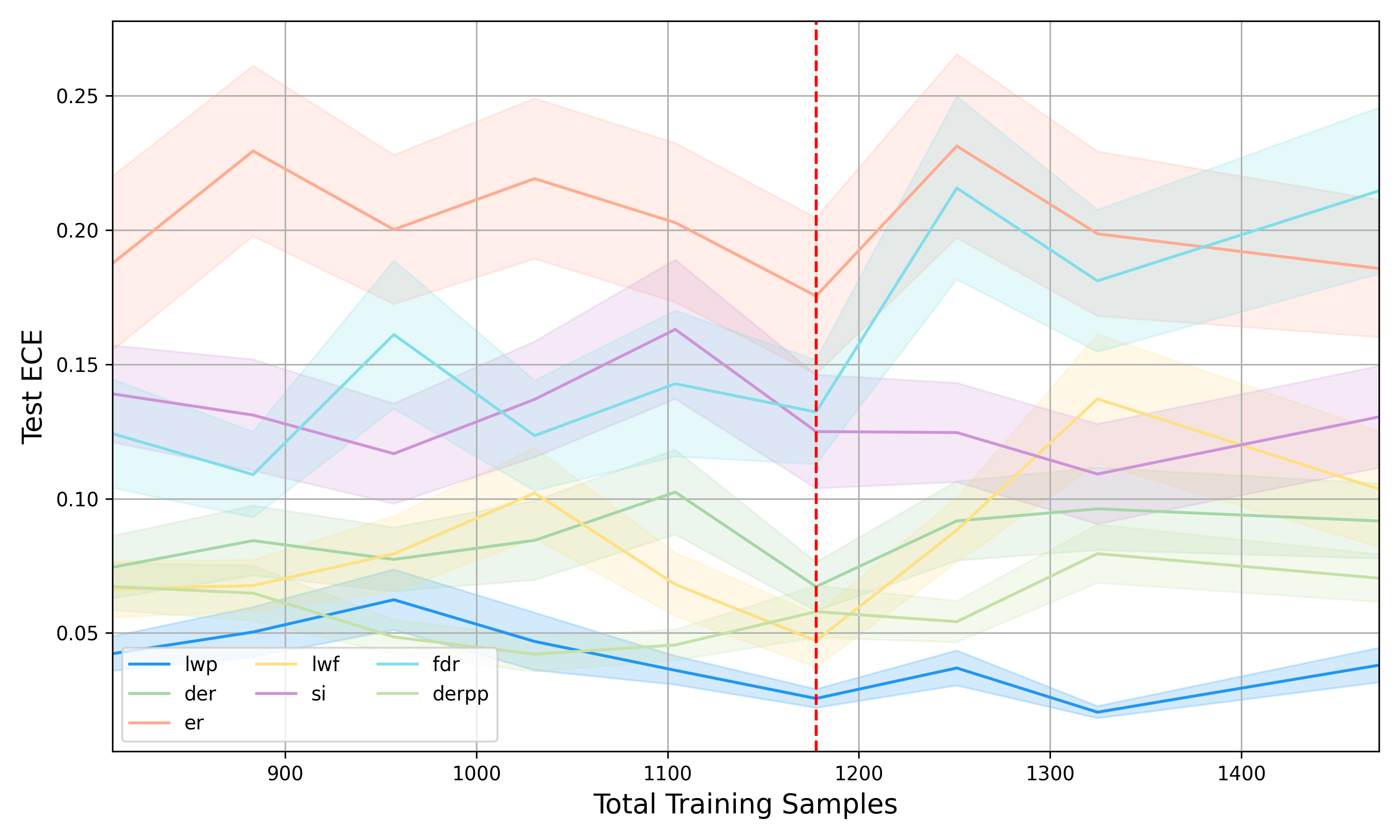}
        \caption{Training Sample ECE}
        \label{fig:subfig2_ece}
    \end{subfigure}
    \caption{Training Sample Accuracy and ECE. Noted, the red line represents the buffer size is greater or equal than the batch size, since the buffer size of replay buffer methods is determined by a percentage of the batch size.}
    \label{fig:main_figure}
\end{figure*}

We include experiment results on the influence of number of training samples to the performance, as shown in Fig. \ref{fig:subfig1_acc}. It shows that our approach outperforms others from 1000 labels and onward, when trained and tested on the PhysiQ dataset.

Fig. \ref{fig:subfig2_ece} illustrates the Expected Calibration Error (ECE) \cite{Nixon_2019_ECE} for each model in relation to the number of training samples. The ECE quantifies how much confidence a model deviates from the actual output distribution, with a lower ECE indicating more accurate confidence assignments for a given classification target. This metric is particularly crucial in safety-critical settings, where the model must provide reliable confidence output. As neural network models frequently demonstrate overconfidence \cite{wei2022mitigatingneuralnetworkoverconfidence}, monitoring ECE becomes essential. The figure reveals that LwP not only maintains the lowest variance across different seeds but also achieves the lowest ECE value when the training sample size exceeds roughly 1000.

\begin{figure*}[!t]
    \centering
    \begin{subfigure}[b]{0.24\textwidth}
        \centering
        \includegraphics[width=\textwidth]{Figures/celeba_backward_transfer_diagram.png}
        \caption{CelebA}
        \label{Fig:celeba_a}
    \end{subfigure}
    \begin{subfigure}[b]{0.24\textwidth}
        \centering
        \includegraphics[width=\textwidth]{Figures/physiq_backward_transfer_diagram.png}
        \caption{PhysiQ}
        \label{fig:physiq_a}
    \end{subfigure} 
    \begin{subfigure}[b]{0.24\textwidth}
        \centering
        \includegraphics[width=\textwidth]{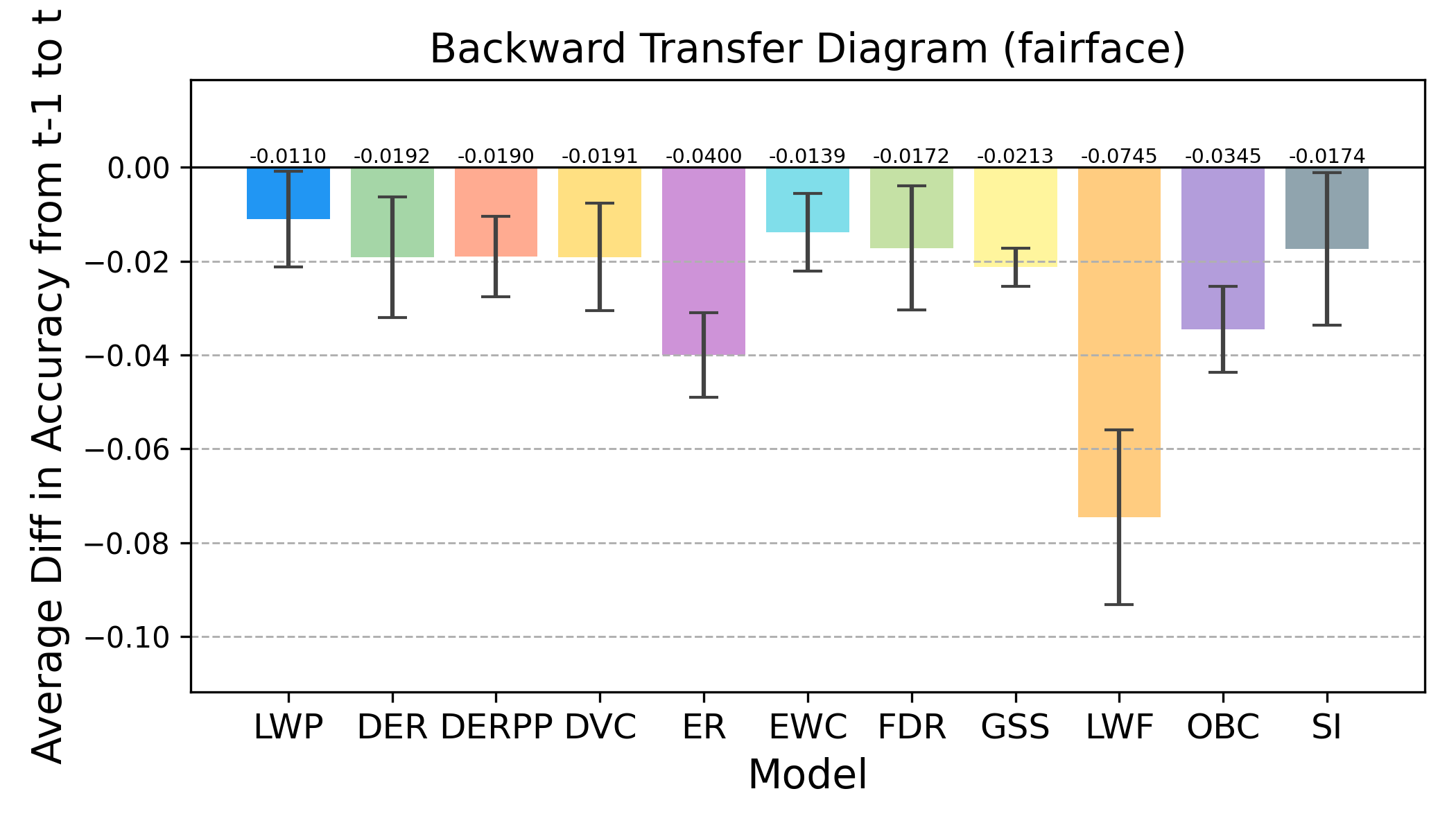}
        \caption{FairFace}
        \label{fig:fairface_a}
    \end{subfigure}
    \begin{subfigure}[b]{0.24\textwidth}
        \centering
        \includegraphics[width=\textwidth]{Figures/bdd100k_backward_transfer_diagram.png}
        \caption{BDD100k}
        \label{fig:bdd100k_a}
    \end{subfigure}
    
    \caption{The Backward Transfer Diagrams for All Four Benchmark Datasets }
    \label{fig:forward_info_loss_diagrams_appendix}
\end{figure*}




\subsection{Performance Improvement per Iteration}\label{appendix:acc_per_iter}

\begin{figure}[htbp]
    \centering
    \begin{subfigure}[b]{0.4\textwidth}
        \centering
        \includegraphics[width=\textwidth]{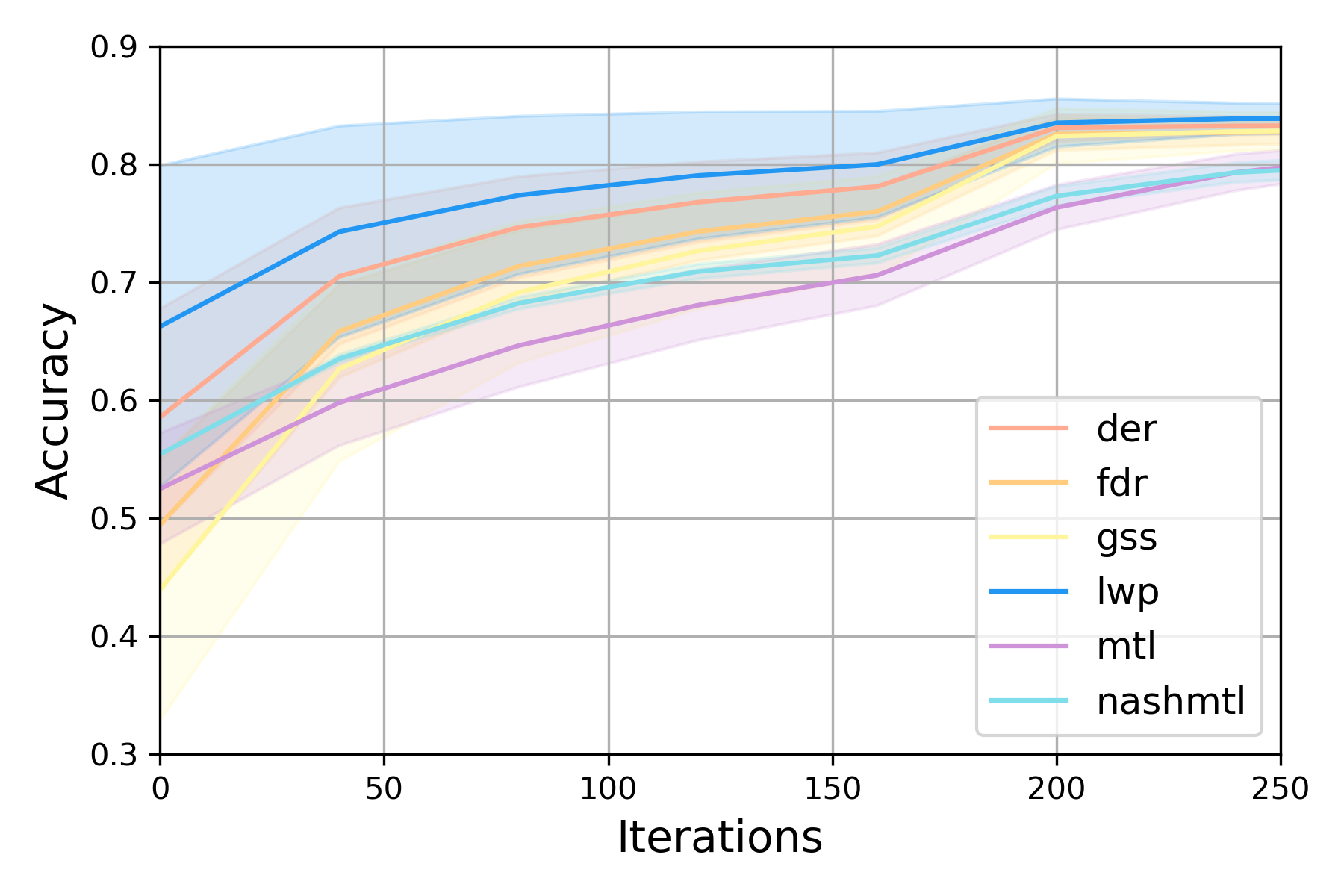}
        \caption{First task iteration}
        \label{Fig:celeba_first}
    \end{subfigure}
    \begin{subfigure}[b]{0.4\textwidth}
        \centering
        \includegraphics[width=\textwidth]{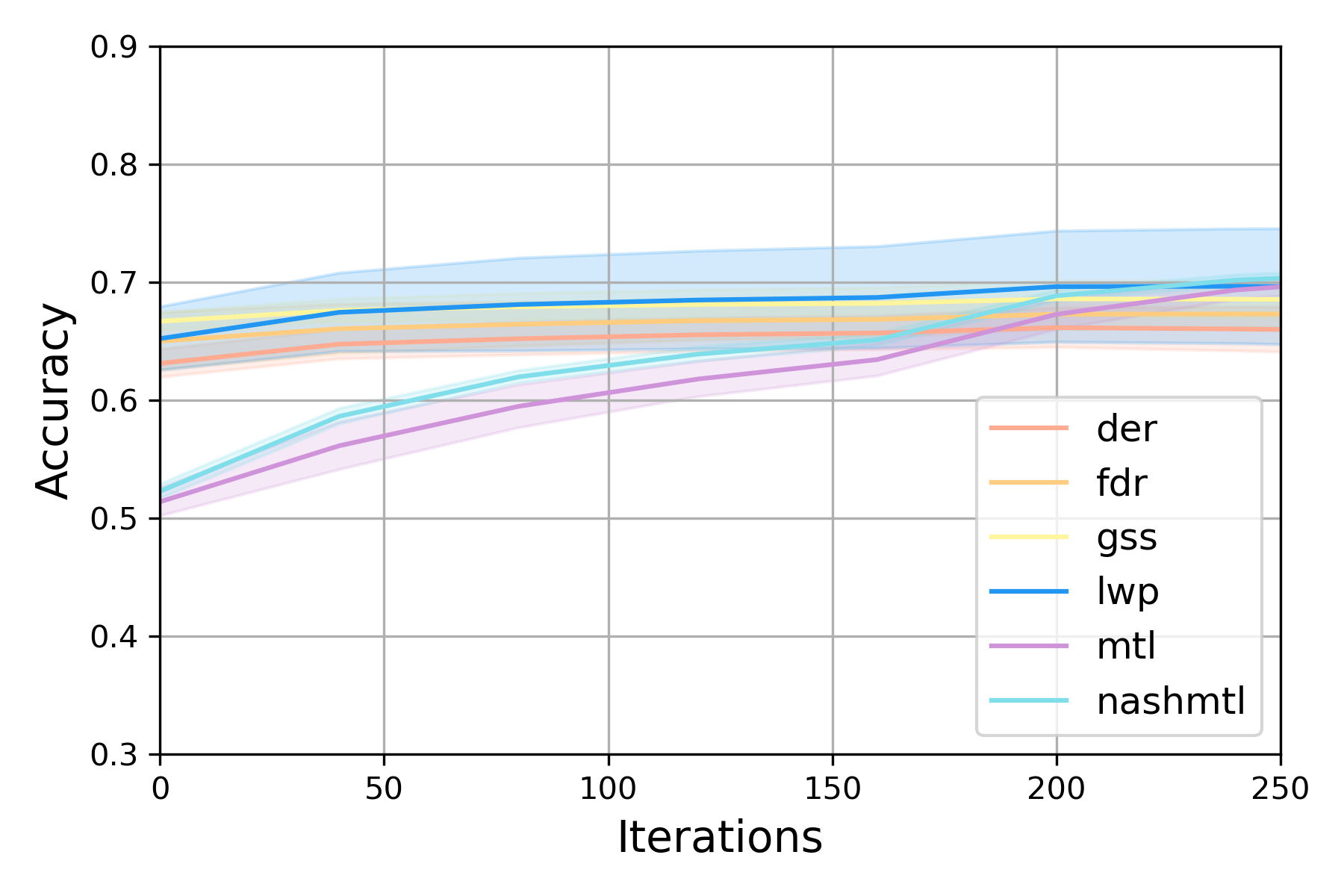}
        \caption{Last (fifth) task iteration}
        \label{fig:celeba_last}
    \end{subfigure} 

    \caption{Average accuracy progression per iteration for CelebA with 5 tasks}
    \label{fig:improving_speed_plots}
\end{figure}

We show that LwP demonstrates a faster improvement per iteration given the same batch size as other CL and MTL models. Here, we plot the evolution of average accuracy across all tasks seen on the test set over training iteration for the top performing CL and MTL baselines along with LwP. We used the CelebA dataset with 5 task splits. To make a fair comparison with CL models, MTL models were trained on the amount of train data that CL models saw in each iteration with access to all 5 tasks. The accuracies of MTL models are calculated up to what CL models have learned so far.

Fig. \ref{Fig:celeba_first} shows how quickly each CL and MTL model learns the first task. This can be understood as the speed at which the models acquire knowledge when they have no prior information to ``recall''. As shown, LwP learns consistently faster per iteration compared to CL and MTL baselines. As MTL models simultaneously learn multiple labels, their convergence per iteration is generally slower compared to CL models in this setting.

Conversely, Fig. \ref{fig:celeba_last} illustrates a case where MTL models are trained from the beginning with labels available for all $t$ tasks, whereas CL models, having been pretrained on $t-1$ tasks, must now incrementally learn the $t^{th}$ task while maintaining performance on old tasks. This configuration is crucial in real-world scenarios where the cost of labeling data typically exceeds that of data collection, prompting the decision to gather more partially labeled data rather than re-labeling existing data. Analogous to the prior scenario, the progression of test accuracy over iterations demonstrates that LwP consistently exceeds other CL models and exhibits performance that is competitive with MTL models, which are considered the upper bound for continual learning. This highlights the comparative benefit of LwP when users face the choice between relabeling existing data and obtaining new data with different labels.

\subsection{Training from MTL to CMTL}\label{appendix:mtltocl}

We initially train the model on the first five tasks using a MTL setting, employing ResNet18 as the encoder with input images of size $64 \times 64 \times 3$. After completing the MTL phase, we extract the encoder and freeze its weights. This frozen encoder is then used to train classifiers for the first five tasks in a CMTL setting with various models. Subsequently, we train the entire model for the last five tasks under the same CMTL setting, utilizing the same frozen encoder on the remaining tasks. This approach enables us to evaluate the effectiveness of our method in transitioning from MTL to CMTL setting while maintaining consistent performance across all tasks.

Our method significantly outperforms other models in terms of accuracy on the CelebA dataset. Specifically, as shown in Table \ref{tab:celeba_mtltocl_resnet18}, LwP achieves an average accuracy of 83.652\%, surpassing all other CL methods tested. The closest competitors, oEWC and SI, achieve accuracies of 82.250\% and 82.194\%, respectively. This demonstrates the effectiveness of our approach in leveraging an MTL pre-trained encoder for subsequent CMTL tasks.

The superior performance of LwP suggests that initializing the encoder with MTL on the first five tasks provides a robust foundation for learning new tasks. Our method effectively mitigates catastrophic forgetting by preserving essential features discovered during the MTL phase while adapting to new tasks, given that the continual tasks are shorter now. This balance between stability and plasticity still allows LwP to maintain high accuracy in the continual learning tasks.
\begin{figure}
    \centering
    \includegraphics[width=0.8\linewidth]{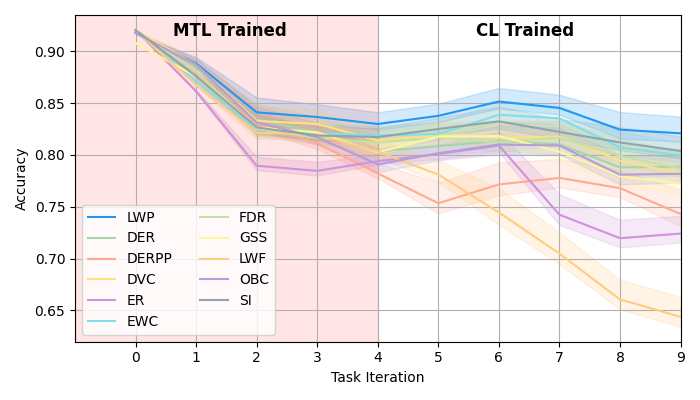}
    \caption{Train the first five tasks in the MTL method (meaning all tasks are trained simultaneously using the basic MTL model, then apply the encoder to CL method models. We used the encoder to further train in the CMTL setting with different models with ResNet 18, with an image size of 64.}
    \label{fig:mtltocl}
\end{figure}

\begin{table}[!ht]
\centering
\small
\caption{Accuracy Percentage Comparison Across Models on CelebA Dataset, Trained on MTL on the first five tasks, then CL on the last five tasks}
\label{tab:celeba_mtltocl_resnet18}
\begin{tabular}{@{}l l c@{}}
\toprule
\textbf{Method Type} & \textbf{Model} & \textbf{ResNet18 ($64\times64$)} \\
\midrule
\multirow{10}{*}{CL} 
& LwF & 74.057 $\pm$ 11.364 \\ 
& oEWC & 82.250 $\pm$ 6.362 \\
& ER & 77.245 $\pm$ 8.434 \\
& SI & 82.194 $\pm$ 6.460 \\
& GSS & 80.563 $\pm$ 8.239 \\
& FDR & 81.271 $\pm$ 7.738 \\
& DER & 81.010 $\pm$ 8.674 \\
& DERPP & 78.177 $\pm$ 9.532 \\
& DVC & 81.387 $\pm$ 7.821 \\
& OBC & 80.516 $\pm$ 8.446 \\
\midrule
\multirow{1}{*}{CMTL} 
& \textbf{LwP} & \textbf{83.652 $\pm$ 7.069} \\
\bottomrule
\end{tabular}
\end{table}


Moreover, the lower standard deviation in LwP's performance indicates consistent results across different runs, highlighting the reliability of our approach. The results confirm that combining MTL pre-training with our proposed CL strategy enhances the model's ability to generalize and adapt to new tasks without compromising performance on previously learned tasks.

Similarly, in Figure \ref{fig:mtltocl}, we averaged the results across task iterations to evaluate performance over time. Our method, LwP, demonstrates minimal accuracy loss when training on new tasks, highlighting its performance against forgetting. The standard deviation---represented as 20\% of the total for visualization purposes---remains low, indicating consistent performance. Although there is a slight increase in standard deviation during later tasks, suggesting a potential drop in accuracy due to forgetting, LwP still preserves knowledge at a superior level compared to other baselines. Even with the first five tasks trained in a multitask setting, our method maintains the best overall accuracy, outperforming other models in preserving learned information.

\begin{table}[htbp]
\centering
\footnotesize
\caption{Hyperparameter sensitivity analysis on BDD100k (weather shift scenario). The best-performing setting for each method is shown in bold.}
\renewcommand{\arraystretch}{1.1} 
\begin{tabular}{lll}
\toprule
\textbf{Method} & \textbf{Parameter Tested} & \textbf{Accuracy} \\
\midrule
\multirow{7}{*}{LwP (Ours)} & $\lambda_{dwdp}=0.01$ (fixed $\lambda_{old}=1$) & 77.71$\pm$4.30 \\
& $\lambda_{dwdp}=0.05$ (fixed $\lambda_{old}=1$) & 77.73$\pm$4.20 \\
& $\lambda_{dwdp}=0.1$ \phantom{0}(fixed $\lambda_{old}=1$) & 77.83$\pm$4.09 \\
& $\lambda_{dwdp}=1$ \phantom{0.0}(fixed $\lambda_{old}=1$) & \textbf{77.93$\pm$4.04} \\
\cmidrule(l){2-3} 
& $\lambda_{old}=0.1$ \phantom{0}(fixed $\lambda_{dwdp}=0.05$) & 77.72$\pm$4.23 \\
& $\lambda_{old}=0.5$ \phantom{0}(fixed $\lambda_{dwdp}=0.05$) & 77.75$\pm$4.23 \\
& $\lambda_{old}=1$ \phantom{0.0}(fixed $\lambda_{dwdp}=0.05$) & 77.53$\pm$4.48 \\
\midrule
\multirow{3}{*}{LwF} & $\lambda=0.01$ & 76.42$\pm$6.08 \\
& $\lambda=0.1$ & 76.91$\pm$5.32 \\
& $\lambda=1$ & 76.79$\pm$5.55 \\
\midrule
\multirow{3}{*}{DERPP} & $\beta=0.01$ ($\alpha=0.1$) & 76.04$\pm$6.84 \\
& $\beta=0.1$ \phantom{0}($\alpha=0.1$) & 76.58$\pm$6.07 \\
& $\beta=1$ \phantom{0.0}($\alpha=0.1$) & 75.36$\pm$9.67 \\
\midrule
\multirow{3}{*}{DVC} & $\lambda=0.01$ & 71.97$\pm$5.23 \\
 & $\lambda=0.1$\phantom{0} & 72.07$\pm$7.59 \\
 & $\lambda=1$\phantom{0.0} & 74.70$\pm$5.26 \\
\midrule
\multirow{3}{*}{OBC} & $\lambda=0.01$ & 75.94$\pm$7.11 \\
 & $\lambda=0.1$\phantom{0} & 72.78$\pm$13.23 \\
 & $\lambda=1$\phantom{0.0} & 77.28$\pm$4.78 \\
\bottomrule
\end{tabular}
\label{tab:hyperparam_combined}
\end{table}
\subsection{Hyperparameter testing}\label{appendix:hyperparameter}

We test the sensitivity of LwP with respect to the hyperparameter values. The experiments are conducted using BDD100k under the weather shift scenario. The results show that different hyperparameter settings do not significantly affect the performance of LwP. Additionally, we provide the hyperparameter tuning results for the two most comparable baselines.

\end{document}